\documentclass[journal]{IEEEtran}
\usepackage{graphicx}
\usepackage{bm}
\usepackage{amsmath}
\usepackage{algorithm}
\usepackage{graphicx}
\usepackage{color}
\usepackage{amsfonts}
\usepackage{amsmath}
\usepackage{multirow}
\usepackage{threeparttable} 
\usepackage{chngpage}
\usepackage{epstopdf}
\usepackage{epsfig}
\usepackage[table]{xcolor}
\usepackage{fixltx2e}
\usepackage{algorithmic}
\usepackage{colortbl}
\usepackage{multirow}
\usepackage{etoolbox}
\usepackage{arydshln}
\usepackage[table]{xcolor}
\usepackage{fixltx2e}
\usepackage{bbm}
\usepackage{scrextend}
\makeatletter
\patchcmd{\@makecaption}
{\scshape}
{}
{}
{}
\makeatother
\newcommand{\argmin}{\operatornamewithlimits{argmin}}

\newcommand{\eq}[1]{Eq. \eqref{#1}}

\begin{document}
\title{Clustering-aware Graph Construction: A Joint Learning Perspective }

\author{Yuheng~Jia,
	Hui~Liu,
	~Junhui~Hou,~\IEEEmembership{Member,~IEEE,}~and
	Sam~Kwong,~\IEEEmembership{Fellow,~IEEE} % <-this % stops a space
	\thanks{Y. Jia and H. Liu are with the Department
		of Computer Science, City University of Hong Kong, Kowloon, Hong Kong,
		(e-mail: yuheng.jia@my.cityu.edu.hk; hliu99-c@my.cityu.edu.hk).} \thanks{J. Hou and S. Kwong  are with the Department of Computer Science,
		City University of Hong Kong, Kowloon, Hong Kong and also with the City University of Hong Kong Shenzhen Research Institute, Shenzhen, 51800, China, (e-mail:  jh.hou@cityu.edu.hk; cssamk@cityu.edu.hk).}
	\thanks{This work was supported in part by the Natural Science Foundation of China
		under Grants 61772344, 61672443, 61873142 and in part by Hong Kong RGC General Research Funds
		9042489 (CityU 11206317), 9042322 (CityU 11200116), and Early Career Scheme Funds
		9048123 (CityU 21211518).}}
\maketitle

\begin{abstract}
Graph-based clustering methods have demonstrated the effectiveness in various applications. 
%The traditional graph-based clustering methods usually first construct a Laplacian graph and then partition the graph to generate the clustering results.%  sequential manner. 
%Graph-based clustering is due to its easy implemtned 
Generally, existing graph-based clustering methods  first construct a graph   to represent the input data  and then partition it to generate the clustering result. %  sequential manner. 
%Graph-based clustering methods due to their excellent performance. 
%A question naturally arises: does the constructed graph fit the requirements for the subsequent partition task? The answer is no!
%A question naturally arises: does the constructed graph fit the requirements for the subsequent partition task? The answer is no!
However, such a stepwise manner may make the constructed graph  not fit the requirements for the subsequent decomposition, leading to compromised clustering accuracy. 
%To this end, we propose to learn the affinity graph and generate the clustering result simultaneously to produce the affinity graph compatible with the subsequent partition task.
%To this end, we propose to learn an affinity matrix that is able to  compatible with the subsequent partition task
%by learning the graph and generating the clustering result simultaneously.
To this end, we propose a joint learning framework, which  is able to  learn the graph and  the clustering result simultaneously, such that the resulting graph is tailored to the clustering task.
The proposed method is formulated as a well-defined nonnegative and off-diagonal constrained optimization problem, which is further  efficiently solved with convergence theoretically guaranteed.  
	The advantage of the proposed model is demonstrated by comparing with 19 state-of-the-art clustering methods  on 10 datasets with 4 clustering metrics. 
\end{abstract}

\IEEEpeerreviewmaketitle
\section{Introduction}

%\red{What is SymNMF and its advantages and applications.}

% As a fundamental task in machine learning, 
 Clustering aims to partition  the input data into different groups, where the samples in the same group are more similar to each other than to those in other groups. 
Many real-world applications can be formulated as a clustering problem, e.g., image segmentation \cite{wu1993optimal,chuang2006fuzzy}, image classification \cite{wu2018pairwise}, community detection \cite{wu2018nonnegative}, recommender system \cite{das2014clustering}, tumor discovery \cite{yu2014double,yu2012sc3,yu2007graph}, and data visualization \cite{gorban2008principal}. 
%The most well known clustering methods is K-means, which iteratively finds the  center of each group of samples and partitions samples to the nearest center, until convergence. 
%However, K-means is only suitable for solving a clustering problem when data are spherically distributed \cite{von2007tutorial,ng2002spectral}. 
%K-means is good at for solving a clustering problem when data are spherically distributed \cite{von2007tutorial,ng2002spectral}. 
%However, K-means is only able to effectively solve a clustering problem when data are spherically distributed \cite{von2007tutorial,ng2002spectral}. 
Over the past several decades, many clustering methods were proposed like K-means, Gaussian mixture models (GMM) \cite{jian2011robust}, mean shift \cite{comaniciu2002mean,cheng1995mean}, and various graph-based clustering methods \cite{ng2002spectral,yang2011nonnegative,kuang2015symnmf,kuang2012symmetric,zelnik2005self}.  
%However, K-means is only able to effectively partition spherically distributed data \cite{von2007tutorial,ng2002spectral}. 
%To partition un-spherical data, many non-K-means models were proposed, like Gaussian mixture models (GMM) \cite{jian2011robust} and mean shift \cite{comaniciu2002mean,cheng1995mean}. 
%Among them,  graph clustering \cite{ng2002spectral,yang2011nonnegative,kuang2015symnmf,kuang2012symmetric,zelnik2005self} is one of the most effective methods.
%Among them, one of the most well known  methods is graph clustering. 
%
%Different from K-means based methods that perform clustering on the raw input, graph based methods first convert the raw data into an affinity graph, where vertexes denote the samples and edges represent the similarities between samples, and then performs clustering on the affinity matrix. 
Particularly,  graph-based clustering  methods have achieved impressive performance in various applications, which  represent input data with a graph,   and then partition the graph into subgraphs. 
The representative graph-based clustering methods are spectral clustering (SC) \cite{ng2002spectral,yang2011nonnegative,zelnik2005self} and symmetric nonnegative matrix factorization (SymNMF) \cite{kuang2015symnmf,kuang2012symmetric}.
How to build a reasonable  graph plays a critical role in graph-based  clustering, since the quality of the graph usually determines the final clustering performance seriously.
The most-well known graph construction method is $p$-nearest-neighbor algorithm that connects the sample with its top $p$ nearest samples with nonnegative weights to measure their similarities, and assigns $0$ to the non-connect samples. This method may not perform well as  it is not robust to various types of noise \cite{5357420}. 
To solve this problem, many advanced graph construction methods were proposed \cite{5357420,liu2010robust,elhamifar2009sparse,peng2017constructing,nie2014clustering,dong2016learning,dong2016learning,wu2018pairwise}. See the detailed review in Section II-B. 
%Although they have been applied to many applications, it is still not clear whether the constructed graph fits the task of clustering or not. 
%Generally, a typical graph clustering  as a three-step process, i.e., first construct a graph, second embed the graph into a reduced dimension space (like spectral embedding), and third partition the embedding through post-processing like K-means. 
Generally, given the constructed  graph, graph-based  clustering  needs two extra steps to complete the clustering task, i.e., $i)$ embed the graph into a low-dimensional  space (like spectral embedding), and $ii)$ divide the embeddings into different clusters through post-processing like K-means. 
\emph{The question then arises: does the constructed graph always fit the requirements for the subsequent partition task? The answer is no!}
In this paper, we study a joint learning  model that can simultaneously construct the  graph and divide the data into different clusters. 
When optimizing the proposed joint model, the tasks of the graph construction and data partition can well communicate with each other to
achieve mutual refinement. Therefore, the resulting graph is tailored to  the clustering task. 
It is also worth pointing out that using a joint optimization framework to deal with two correlated tasks   has  proven to be effective in many works \cite{wu2018pairwise,li2017structured,li2015learning,fang2016robust,lai2018generalized,zhang2019joint,nie2019semi,zhang2019joint}.
% with a novel self multi-view manner. 
%Specifically, we argue the learned graph should explore information from the following three views: $i)$ since the predefined graph is usually practical, the learned similarity graph should approximate the initial similarity graph; $ii)$, borrowing  from the idea that data lying in a union linear subspaces could be expressed as a linear combination of all other data points \cite{peng2017constructing,liu2010robust,fang2016robust,liu2013robust}, we adopt the self-expressiveness property to build the similarity graph, which connect the graph with raw features; and $iii)$, the clustering result is usually more discriminative than the input graph, if we feed back the clustering result properly, it may help to construct a better similarity graph.  
%Specifically, we argue graph learning should explore information from the following three views: 
Specifically, the constructed graph by our method  explores information from the following three aspects: 
%$i)$ since the predefined graph is usually practical, the learned similarity graph should approximate the initial similarity graph; 
$i)$ the initial similarity graph, which is  practical in many applications; 
$ii)$, the input data, which contain rich information; 
%borrowing  from the idea that data lying in a union linear subspaces could be expressed as a linear combination of all other data points \cite{peng2017constructing,liu2010robust,fang2016robust,liu2013robust}, we adopt the self-expressiveness property to build the similarity graph, which connect the graph with raw features; 
%and $iii)$, the clustering result is usually more discriminative than the input graph, if we feed back the clustering result properly, it may help to construct a better similarity graph. 
and $iii)$, the clustering result, which is more discriminative than the input data.  
The proposed model is finally formulated as a nonnegative and off-diagonal\footnote{The learned similarity matrix should be an  off-diagonal matrix to avoid the trivial solution.} constrained optimization problem, which can be solved efficiently in an iterative manner with convergence guaranteed.
%
%{Moreover, we extend the proposed model  to a kernel version to exploit the non-linear relation of the input data.}  
%\blue{Moreover, we constrain  the clustering membership matrix as an orthogonal matrix to increase the feedback from the output.} 
%Both the proposed model and its kernel version are solved efficiently with convergence theoretically guaranteed.  
%Both the proposed model and its kernel version are solved efficiently with convergence theoretically guaranteed.  
By comparing the proposed model with 19 state-of-the-art clustering methods on 10 widely used datasets with 4  clustering metrics, the  advantage  of the proposed model is validated. In addition, the improvement of the proposed model  is confirmed by the Wilcoxon rank sum test with a significance level of $0.05$.

%We use the self-expression property 

The main contributions of this paper are summarized as follows.
\begin{enumerate}
%	\item[1.] In contrast to the existing graph clustering methods that adopt a predefined similarity matrix, we propose to learn the similarity matrix by a self multi-view manner that explores the complementary information from the predefined similarity graph (view 1), the raw features (view 2) and the discriminative clustering result (view 3). 
%	
%	a graph clustering model that can simultaneously learn a similarity matrix and a clustering membership matrix, which connects the clustering performance with the input data directly. 
%	Moreover, 
%	The proposed model is formulated as non-negative constrained optimization problem.
	\item[1.] 
%	Different from the existing graph clustering methods that first build a similarity matrix and then perform clustering on the similarity matrix, the proposed method simultaneously learns a clustering membership matrix and a similarity matrix, which can exploit the mutual enhancement between the two separate steps and lead to a more global solution.  
	The proposed method simultaneously learns a cluster membership matrix and an affinity graph, which can exploit the mutual enhancement relation between the two separate steps and lead to a more global solution.  
%	\item[2.] The adopted graph is constructed   by a self multi-view manner that explores the complementary information from the predefined similarity graph (view 1), the raw features (view 2) and the discriminative clustering result (view 3).
%	\item[2.] The adopted graph is constructed  by exploring the complementary information from the predefined similarity graph (view 1), the raw features (view 2) and the discriminative clustering result (view 3).
%	Since the propose model simultaneously learns a clustering membership matrix and a similarity matrix, it can exploit the mutual enhancement between
%	that provides the discriminative information,
%	 
%	 the proposed model can exploit the mutual enhancement between similarity and clustering
%	\item[2.] The proposed optimization method has the following theoretical guarantees: $i)$, the constraints in the proposed model can be naturally satisfied\footnote{i.e., non-negativity for all  the variables and off-diagonal  for the similarity matrix.}  in the optimization process; 
%	%proposed model guarantees the non-negativity for each variable, 
%	$ii)$, each optimization step can decrease the value of the objective function; and  $iii)$, the converged limit point is a stationary point that satisfies the Karush-Kuhn-Tucker (KKT) conditions.
	\item[2.] \textcolor{black}{The proposed optimization algorithm has the following theoretical guarantees: $i)$ the constraints in the proposed model can be naturally satisfied\footnote{i.e., non-negativity for all  the variables and off-diagonal  for the similarity matrix.}  in the optimization process; 
	%proposed model guarantees the non-negativity for each variable, 
	and $ii)$ each iteration can decrease the objective function to converge.}
%	\item[4.] Since the optimization solution is only related to the inner product of the raw features, we extend the proposed model to a kernel version to exploit the non-linear relationship of the input data, which makes  the proposed model more flexible. 
%	\item[3.] To increase the feedback from clustering results, we constrain the output matrix as an orthogonal matrix. 
%	\item[3.] Since the proposed solution is only related to the inner product of the input data, the proposed model  can be  extended to a kernel version to exploit  the non-linear ship in the input data. 
	 
\end{enumerate}

The rest of this  paper is organized as follows. In Section \uppercase\expandafter{\romannumeral2}, we  discuss the related works. Section \uppercase\expandafter{\romannumeral3} presents the proposed model, the optimization algorithm, and its  computational complexity analysis and theoretical guarantees. 
%Section \uppercase\expandafter{\romannumeral4} provides the relations between the proposed model with several related works.
Experimental  comparisons and analyses are shown in  Section \uppercase\expandafter{\romannumeral4}, and finally Section \uppercase\expandafter{\romannumeral5} concludes this paper.

\section{Related Work}
\subsection{Notation}
Throughout this paper, matrices are denoted by bold uppercase letters, e.g., $\mathbf{A}$, and  the element at the $i$th row and $j$th column of $\mathbf{A}$ is denoted as $\mathbf{A}_{ij}$ or $a_{ij}$. Vectors are represented by bold lowercase letters, e.g., $\mathbf{a}$ and scalars are represented by italic lowercase letters, e.g., $a$. Moreover, $^\mathsf{T}$ stands for the transpose of a matrix, $\|\mathbf{A}\|_F=\sqrt{\sum_{i}\sum_{j}\mathbf{A}_{ij}^2}$ is the Frobenius norm of  matrix $\mathbf{A}$, $\|\mathbf{A}\|_\infty=\max_{ij}|\mathbf{A}_{ij}|$ returns the maximum absolute value of  matrix $\mathbf{A}$, ${\rm diag}(\cdot)$ returns the diagonal elements of a matrix as a vector, $\odot$ returns the Hadamard product of two matrices, i.e., the element-wise multiplication of two matrices,  ${\rm exp}(\cdot)$ returns the exponential value,  $\langle \cdot,\cdot\rangle$ calculates the inner product of two matrices, $\mathbf{I}_k$ denotes an identity matrix of size $k\times k$, and $\mathbf{A}\geq 0$ means  each element of $\mathbf{A}$ is not less than  $0$,  i,e., $\mathbf{A}_{ij}\geq0,\forall i,j$.
$\mathbf{X}=\{\mathbf{x}_1,\mathbf{x}_2,\dots,\mathbf{x}_n\}\in\mathbb{R}^{d\times n}$ denotes
 the input data,
$\mathbf{x}_i\in\mathbb{R}^{d\times 1}$ is the $i$th sample; $n$, $d$, and $c$  represent the number of samples, the dimension of features, and the number of classes, respectively,

\subsection{Graph-based Clustering}
%Although K-means can be solved efficiently, it is usually not able to correctly partition the non-spherical distributed samples. To solve this problem, many graph clustering methods were proposed \cite{kuang2015symnmf,ng2002spectral}. 
Different from traditional clustering methods (e.g., K-means) that  partition  the raw features  $\mathbf{X}$ straightforwardly,  graph-based clustering \cite{kuang2015symnmf,ng2002spectral} transforms data clustering as a graph partition problem. Specifically, a typical graph-based clustering method is composed of the following steps:
%graph clustering first builds a graph to represent the initial data, then performs data partition on the constructed graph, which usually outperforms the traditional clustering methods such as k-means. 
\begin{enumerate}
	\item[1.] Given  $\mathbf{X}$, generate an affinity   matrix $\mathbf{G}\in\mathbb{R}^{n\times n}$ to represent $\mathbf{X}$, where the entries in $\mathbf{G}$ denote the similarities between the  corresponding samples. 
%	\item[2.] Normalize the  affinity matrix $\mathbf{G}$ to obtain the Laplacian  matrix $\mathbf{W}\in\mathbb{R}^{n\times n}$. 
	\item[2.] \textcolor{black}{Decompose the normalized affinity matrix $\mathbf{W}\in\mathbb{R}^{n\times n}$  to generate the low-dimensional embeddings\footnote{\textcolor{black}{It is known that a normalized affinity matrix $\mathbf{W}\in\mathbb{R}^{n\times n}$ usually can achieve better performance than $\mathbf{G}$ \cite{ng2002spectral,kuang2015symnmf,von2007tutorial}. See the analysis about
				how to normalize $\mathbf{G}$ at \cite{von2007tutorial}.}}.}
	\item[3.] Obtain the  cluster indicator matrix according to  the   embeddings.
\end{enumerate}
In the following, we will briefly discuss the widely used approaches for each step. 
\subsubsection{Graph construction}
The most widely used graph is $p$-nearest-neighbor ($p$NN) graph \cite{von2007tutorial} that only connects a specified sample with its top $p$ nearest samples under some distance metrics. Specifically,  
\begin{equation}
\mathbf{G}_{ij}=\begin{cases}
\mathcal{G}_{ij}&~\mathbf{x}_j\in \mathcal{P}(\mathbf{x}_i)\\
0&~{\rm otherwise},
\end{cases}
\end{equation}
where $\mathcal{P}(\mathbf{x}_i)$ indicates the top $p$ nearest samples of $\mathbf{x}_i$, and  $\mathcal{G}_{ij}$ is the weight between  $\mathbf{x}_i$ and $\mathbf{x}_j$. 
The binary weighting strategy simply sets  $\mathcal{G}_{ij}=1$ for the connected samples. 
Another weighting strategy is to use the  radial basis function  kernel (RBF), i.e.,
\begin{equation}
\mathcal{G}_{ij}={\rm exp}\left(\frac{\|\mathbf{x}_i-\mathbf{x}_j\|^2}{\sigma^2}\right),
\end{equation} 
to measure the similarities between the samples, where $\sigma^2$ is the bandwidth of the RBF kernel.
Another widely used graph construction method is $\epsilon$-neighborhood graph that connects a certain sample with other samples within a ball of radius  $\epsilon$.

Both $p$NN and $\epsilon$-neighborhood graphs are sensitive to outliers and noise. To overcome this drawback,  many advanced learning methods were proposed to construct the weight matrix of the graph recently. For example, 
 Cheng \emph{et al.} \cite{5357420} proposed to learn an $\ell_1$ graph based on the sparsity property of the $\ell_1$ norm. 
  Nie \emph{et al.} \cite{nie2014clustering} proposed to learn the neighbors  adaptively. \textcolor{black}{Dong \emph{et al.}
 \cite{dong2016learning,dong2019learning} constructed the graph from the perspective of graph signal processing \cite{pavez2016generalized}.} \textcolor{black}{The smoothness prior is also investigated in graph construction \cite{kalofolias2016learn,chepuri2017learning}.}
 Wu \emph{et al.}  \cite{wu2018pairwise} constructed a discriminative graph with the guidance of the supervisory information. Moreover, 
 many models build the affinity graph  by the self-representative property of the input data  \cite{liu2010robust,elhamifar2009sparse,peng2017constructing}.
 
% \textcolor{blue}{In graph clustering methods like SC and SymNMF, the normalized graph $\mathbf{W}$ \cite{ng2002spectral,kuang2015symnmf,von2007tutorial} usually can achieve
% better clustering performance than the original graph $\mathbf{G}$. See the analysis about
% how to normalize $\mathbf{G}$ at \cite{von2007tutorial}.}

%\red{graph learning methods, semi-supervised learning}
%\red{discus k-nn graph method, learning method including the self expressive method, LLE, }

%\subsubsection{Normalization method}、
%Many graph clustering methods distinguish each other by different normalization methods. For example, Shi and Malik \cite{868688} proposed a spectral clustering called N-Cut with a unnormalized Laplacian, i.e., $\mathbf{W}=\mathbf{H}-\mathbf{G}$, where $\mathbf{H}$ is a diagonal matrix with its $i$th diagonal element $\mathbf{H}_{ii}=\sum_{j}\mathbf{G}_{ij}$; and Ng \emph{et al.} \cite{ng2002spectral} proposed  a symmetric unnormalized Lalpacian matrix, i.e., $\mathbf{W}=\mathbf{D}^{-1/2}\mathbf{G}\mathbf{D}^{-1/2}$. 

\subsubsection{Low-dimensional  embedding}
Given $\mathbf{W}$, graph-based clustering decomposes  $\mathbf{W}$ to generate a lower-dimensional embedding. For example, 
SC \cite{ng2002spectral} is formulated as
%SC \cite{ng2002spectral} formulate graph clustering as a spectral decomposition problem, 
\begin{equation}
\min_\mathbf{V}\|\mathbf{W}-\mathbf{VV}^\mathsf{T}\|_F^2,~{\rm s.t.,}\mathbf{V}^\mathsf{T}\mathbf{V}=\mathbf{I}_k,
\end{equation} 
where $\mathbf{V}\in\mathbb{R}^{n\times k}$ is the dimension-reduced embedding and can be calculated by the spectral decomposition of $\mathbf{W}$. 
As an alternative, SymNMF \cite{kuang2012symmetric,kuang2015symnmf} decomposes the affinity matrix to be the product of a nonnegative matrix and its transpose, 
\begin{equation}
\min_\mathbf{V}\|\mathbf{W}-\mathbf{VV}^\mathsf{T}\|_F^2,~{\rm s.t.,}\mathbf{V}\geq0,
\label{loss-SymNMF}
\end{equation}
to produce the lower-dimensional embedding.  Other advanced methods like sparse SC (SSC) \cite{lu2016convex} seeks a block diagonal appearance of  $\mathbf{VV}^\mathsf{T}$, and  nonnegative spectral clustering \cite{yang2011nonnegative} generates an orthogonal nonnegative embedding. 

\subsubsection{Generation of the clustering indicator matrix}
Since the lower-dimensional embedding in graph-based clustering usually cannot indicate the cluster membership, traditionally post-processing like K-means should be carried out to obtain the final clustering result. 
As a special case, SymNMF generates a nonnegative embedding and the position  of the largest value in the $i$th row indicates the cluster membership of $\mathbf{x}_i$. 
%
%
%\subsubsection{Symmetric Non-negative Matrix Factorization for Graph Clustering (SymNMF)}
%To improve the interpretability of SC, symmetric non-negative matrix factorization  (SymNMF) \cite{kuang2012symmetric,kuang2015symnmf} was proposed. Different from SC, which needs a post-processing step to generate the finial clustering results, SymNMF can directly produce the clustering results.
%
%Different from above mentioned clustering methods, graph clustering fi
%\begin{equation}
%\min_\mathbf{V}\|\mathbf{W}-\mathbf{VV}^\mathsf{T}\|_F^2,~{\rm s.t.,}\mathbf{V}\geq0,
%\label{loss-SymNMF}
%\end{equation}
%where $\mathbf{V}\in\mathbb{R}^{n\times c}$ is a clustering
%\subsubsection{Non-negative Spectral Clustering}
%Non-negative spectral clustering (NSC)  \cite{yang2011nonnegative}
%
%\begin{equation}
%\min_\mathbf{V}\|\mathbf{W}-\mathbf{VV}^\mathsf{T}\|_F^2+\eta\|\mathbf{V}^\mathbf{T}\mathbf{V}-\mathbf{I}\|,~{\rm s.t.,}\mathbf{V}\geq0,
%\end{equation}

%\subsection{Motivation}

\section{Proposed Model}
%As aforementioned, the existing graph clustering methods suffer from the following limitations:
%\begin{itemize}
%	\item their clustering performance highly rely on the quality of the  initial similarity graph, however, it is quite challenging to judge whether the initial graph is good enough or not. 
%	%	\item moreover, once the initial graph is determined, graph clustering is no longer related to the input data, which violates the common sense; 
%	\item moreover, once the initial graph is determined, graph clustering is no longer related to the input data, which neglects rich information from the raw features.
%	%	\item they establishes a data independent behavior, 
%	%	\item rely on a pre-defined graph  
%	\item the clustering result may discover the discriminative information of input data, however, graph clustering is forward, i.e., the discriminative clustering result cannot in turn affect the clustering process.
%\end{itemize}
\subsection{Model Formulation}
%As aforementioned, the quality of graph determines the performance of a graph clustering method. However, it is quite challenging to judge whether the initial graph is good enough or not. To solve this problem, we propose to learn a graph  in the clustering process adaptively. 
As aforementioned, the quality of the graph determines the clustering  performance of a graph-based  clustering method seriously. 
However, the existing graph-based clustering methods usually first construct a graph, and then partition it to generate the clustering result with extra processes. It is not clear  whether the constructed  graph fits the partitioning or not? 
In  other words, the  graph construction methods of existing graph-based clustering methods are not specifically  designed with a  clustering-purpose, which may lead to compromised clustering performance. 
%However, it is quite challenging to judge whether the initial graph is good enough or not. 

To this end, we propose a clustering-aware graph construction model, which can   learn an adaptive graph and  generate the clustering result simultaneously. 
Along with optimizing the proposed joint model, the constructed graph is able to fit the clustering task.
%learn a graph  in the clustering process adaptively. 
%And thus, the clustering result can feedback valuable discriminative information to guide the construction of the graph. 
Specifically, the proposed model is mathematically formulated as:
%\begin{equation}
%\begin{split}
%&\min_{\mathbf{V},\mathbf{S}}\|\mathbf{S}-\mathbf{VV}^\mathsf{T}\|_F^2+\alpha\|\mathbf{X}-\mathbf{XS}\|_F^2+\beta\|\mathbf{S}-\mathbf{W}\|_F^2\\
%&{\rm s.t.,}\mathbf{V}\geq0,\mathbf{S}\geq0,{\rm diag}(\mathbf{S})=0, 
%\end{split}
%\label{obj}
%\end{equation}
\begin{equation}
\begin{split}
&\min_{\mathbf{V},\mathbf{S}}\alpha\|\mathbf{X}-\mathbf{XS}\|_F^2+\|\mathbf{S}-\mathbf{VV}^\mathsf{T}\|_F^2+\beta\|\mathbf{S}-\mathbf{W}\|_F^2\\
&{\rm s.t.,}\mathbf{V}\geq0,\mathbf{S}\geq0,{\rm diag}(\mathbf{S})=0, 
\end{split}
\label{obj}
\end{equation}
where $\mathbf{S}\in\mathbb{R}^{n\times n}$ is the adaptive similarity matrix, $\mathbf{V}\in\mathbb{R}^{n\times c}$ is the  clustering indicator matrix with $c$ being the number of classes, $\mathbf{W}$ is the initial normalized affinity matrix via a typical graph reconstruction algorithm, \textcolor{black}{and $\alpha\geq0,\beta\geq0$ are two hyper-parameters to balance the contributions of different terms.}
In what follows, we will explain the proposed model in detail. 
%We will introduce each component of \eq{obj} in detail. 

The first term $\|\mathbf{X}-\mathbf{XS}\|_F^2$ as well as the constraints $\mathbf{S}\geq 0$, and ${\rm diag}(\mathbf{S})=0$  explores the relationship between   samples with a self-expressive manner, ${\rm diag} (\mathbf{S})=0$ removes the trivial solution, i.e.,  $\mathbf{S}=\mathbf{I}_n$, and $\mathbf{S}\geq 0$ guarantees that the learned weight matrix $\mathbf{S}$ is a valid similarity  matrix.
%First, the proposed model adopts a self-expressive model \cite{peng2017constructing} to explore  the information from raw features, i.e., 
%\begin{equation}
%\min_{\mathbf{S}}\|\mathbf{X}-\mathbf{XS}\|_F^2+\|\mathbf{S}\|_F^2, {\rm s.t., }\mathbf{S}\geq 0, {\rm diag} (\mathbf{S})=0,
%\label{X-component}
%\end{equation}
%where ${\rm diag} (\mathbf{S})=0$ removes the trivial solution that $\mathbf{S}=\mathbf{I}_n$. 
In addition, it has been theoretically  proven  that a self-expressive  model  has the property of intra-subspace projection dominance (IPD) \cite{peng2017constructing}, i.e., the coefficients over intra subspaces data points are larger than those over inter subspace data points.  Based on IPD, it is expected that $\mathbf{S}_{ij}$ for samples from the same subspace will have larger values.
%Note  that for the traditional graph clustering methods, once the initial graph is determined, graph clustering  is no longer related to the input data, which violates the common sense. Our model solves this problem by exploiting the information of raw features in the clustering process. 

From the forward perspective, the  second term $\|\mathbf{S}-\mathbf{VV}^\mathsf{T}\|_F^2$ with the nonnegative constraint $\mathbf{V}\geq 0$ is responsible for generating the clustering result. 
%The second term $\|\mathbf{S}-\mathbf{VV}^\mathsf{T}\|_F^2$ uses the clustering result 
%to refine the learned $\mathbf{S}$. 
That is, assuming $\mathbf{S}$ is available,   
the position of the largest value in $i$th row of the decomposed  $\mathbf{V}$ indicates the cluster membership of $\mathbf{x}_i,i\in\{i,\dots,n\}$  like SymNMF. 
%Therefore  $\mathbf{V}$ is able to play as a clustering indicator matrix in our model. 
From the backward perspective, the clustering result $\mathbf{V}$  will be beneficial to the learning of the unknown similarity matrix $\mathbf{S}$. 
\textcolor{black}{That is, the nonnegative constraint on $\mathbf{V}$ makes the rows of the resulting $\mathbf{V}$, which are the low-dimensional representations of input samples,  to be more discriminative, and  the inner product $\mathbf{VV}^\mathsf{T}$ can indicate the similarity between  samples precisely, which is further propagated to $\mathbf{S}$ by minimizing the second term. }
Moreover, for an ideal similarity matrix, we have 
\begin{equation}
\mathbf{S}_{ij}=\begin{cases}
1,&{\rm if~}l(\mathbf{x}_i)=l(\mathbf{x}_j)\\
0,&{\rm if~}l(\mathbf{x}_i)\neq l(\mathbf{x}_j),\\
\end{cases}
\end{equation}
where $l(\mathbf{x}_i)$ returns the ground-truth label of $\mathbf{x}_i$. It is clear that the ideal similarity matrix is  block diagonal and  low rank. Minimizing the second term will also seek the low-rankness of $\mathbf{S}$ to pursue the ideal appearance, since the rank of $\mathbf{VV}^\mathsf{T}$ is no greater  than $c$.
%We argue that  the clustering result may discover the discriminative information of the input data (for example, an imperfect graph may produce a perfect cluster membership), which may be helpful in the graph construction.  
Such a bi-directional strategy   is  different from the traditional forward graph-based clustering methods, which overlook the information from clustering result in graph construction. 
%Note that both the cluster indicator matrix $\mathbf{V}$ and similarity matrix $\mathbf{S}$ are the to be optimized variables in  \eq{V-component}, which is different from the objective function of SymNMF in \eq{loss-SymNMF} that only learns the clustering indicator matrix. 
%
%Second, different from the traditional forward graph clustering methods, we argue that 
%the clustering result may discover the discriminative information of input data, which maybe helpful in the construction of graph. Accordingly, 
% the similarity matrix $\mathbf{S}$ can also gain information from the cluster membership matrix, i.e., 
%\begin{equation}
%\begin{split}
%&\min_{\mathbf{V},\mathbf{S}}\|\mathbf{S}-\mathbf{VV}^\mathsf{T}\|_F^2\\
%&{\rm s.t.,}\mathbf{V}\geq0,
%\end{split}
%\label{V-component}
%\end{equation}
%where $\mathbf{V}$ indicates the clustering membership  like SymNMF, 
%Note that both the cluster indicator matrix $\mathbf{V}$ and similarity matrix $\mathbf{S}$ are the to be optimized variables in  \eq{V-component}, which is different from the objective function of SymNMF in \eq{loss-SymNMF} that only learns the clustering indicator matrix. 

The third term  has two functions:  $i)$ it approximates the
initial graph which is usually useful in
practical applications, and  $ii)$ it serves as an $\ell_2$ regularization  prior,  which encourages the self-expressive  term to produce more connections between data samples \cite{lu2012robust}. 
When optimizing \eq{obj} iteratively,  the clustering result can generate  valuable discriminative information and give feedback  to guide the construction of the similarity graph. 
Therefore, the learned graph is tailored to  the clustering task. Moreover, since $\mathbf{V}$ is nonnegative, our model is able to generate the clustering indicator without extra post-processing like K-means.

\subsection{Optimization Method}
To solve \eq{obj}, we first introduce the Lagrangian function as
\begin{equation}
\mathcal{L}(\mathbf{S},\mathbf{V},\mathbf{\Phi},\mathbf{\Psi})=\mathcal{O}(\mathbf{V},\mathbf{S})-\langle \mathbf{\Phi},\mathbf{S} \rangle-\langle \mathbf{\Psi},\mathbf{V} \rangle,
\label{Lagrangian}
\end{equation}
where $\mathbf{\Phi}\in\mathbb{R}^{n\times n}\geq 0$  and $\mathbf{\Psi}\in\mathbb{R}^{c\times n}\geq 0$ are  the Lagrangian multiplier matrices and $\mathcal{O}(\mathbf{V},\mathbf{S})$ denotes the objective function in \eq{obj}. 
Following the Karush-Kuhn-Tucker (KKT) conditions, the optimal solution of \eq{obj} also makes the derivatives of $\mathcal{L}(\mathbf{S},\mathbf{V},
\mathbf{\Phi},\mathbf{\Psi})$ with respect to (w.r.t.) $\mathbf{S}$ and $\mathbf{V}$ to be $0$, i.e.,
\begin{small}
\begin{equation}
\frac{\partial\mathcal{L}}{\partial\mathbf{S}}=2\left(\mathbf{S}-\mathbf{VV}^\mathsf{T}\right)+2\alpha\mathbf{X}^\mathsf{T}\left(\mathbf{XS}-\mathbf{X}\right)+2\beta\left(\mathbf{S}-\mathbf{W}\right)-\mathbf{\Phi}=\mathbf{0},
\end{equation}
\end{small}and 
\begin{equation}
\frac{\partial\mathcal{L}}{\partial\mathbf{V}}=-2(\mathbf{SV}+\mathbf{S}^\mathsf{T}\mathbf{V})+4\mathbf{VV}^\mathsf{T}\mathbf{V}-\mathbf{\Psi}=\mathbf{0},
\end{equation}
where $\mathbf{0}$ is a zero matrix with proper size.
%\begin{equation}
%\frac{\partial\mathcal{L}}{den}=0
%\end{equation}
From the KKT complementary slackness conditions  $\mathbf{\Phi}_{ij}\mathbf{S}_{ij}^2=0$ and $\mathbf{\Psi}_{ij}\mathbf{V}_{ij}^4=0,\forall i,j$, we obtain the following updating equations for $\mathbf{S}$ and $\mathbf{V}$, respectively, i.e.,
\textcolor{black}{
\begin{small}
\begin{equation}
\mathbf{S}_{ij}^{t+1}=\mathbf{S}_{ij}^{t} \left(\frac{\left(\mathbf{V}^t\mathbf{V}^{t^\mathsf{T}}+\alpha\left(\mathbf{X}^\mathsf{T}\mathbf{X}\right)^{+}+\alpha\left(\mathbf{X}^\mathsf{T}\mathbf{X}\right)^{-}\mathbf{S}^t+\beta\mathbf{W}\right)_{ij}}{\left(\mathbf{S}^t+\alpha\left(\mathbf{X}^\mathsf{T}\mathbf{X}\right)^{+}\mathbf{S}^t+\alpha\left(\mathbf{X}^\mathsf{T}\mathbf{X}\right)^{-}+\beta\mathbf{S}^t\right)_{ij}}\right)^{\frac{1}{2}},
\label{update-S}
\end{equation}
\end{small}
and
\begin{equation}
\mathbf{V}_{ij}^{t+1}=\mathbf{V}_{ij}^t\left(\frac{\left(\mathbf{S}^{t+1}\mathbf{V}^t+\mathbf{S}^{{t+1}^\mathsf{T}}\mathbf{V}^t\right)_{ij}}{\left(2\mathbf{V}^t\mathbf{V}^{t^\mathsf{T}}\mathbf{V}^t\right)_{ij}}\right)^{\frac{1}{4}}.
\label{update-V}
\end{equation}}where $\left(\mathbf{X}^\mathsf{T}\mathbf{X}\right)^{+}$ and $\left(\mathbf{X}^\mathsf{T}\mathbf{X}\right)^{-}$ separate the positive and negative elements of $\mathbf{X}^\mathsf{T}\mathbf{X}$\footnote{As will be shown in Section III-D, this separation guarantees the non-negativity of $\mathbf{S}$ and $\mathbf{V}$ in the optimization procedure.}, i.e.,
%\begin{equation}
%\left(\mathbf{X}^\mathsf{T}\mathbf{XS}\right)^{+}=\frac{|\mathbf{X}^\mathsf{T}\mathbf{XS}|+\mathbf{X}^\mathsf{T}\mathbf{XS}}{2}
%\end{equation}
%and 
%\begin{equation}
%\left(\mathbf{X}^\mathsf{T}\mathbf{XS}\right)^{-}=\frac{|\mathbf{X}^\mathsf{T}\mathbf{XS}|-\mathbf{X}^\mathsf{T}\mathbf{XS}}{2}
%\end{equation}
%
%\begin{equation}
%\left(\mathbf{X}^\mathsf{T}\mathbf{X}\right)^{+}=\frac{|\mathbf{X}^\mathsf{T}\mathbf{X}|+\mathbf{X}^\mathsf{T}\mathbf{X}}{2},
%\end{equation}
%and 
%\begin{equation}
%\left(\mathbf{X}^\mathsf{T}\mathbf{X}\right)^{-}=\frac{|\mathbf{X}^\mathsf{T}\mathbf{X}|-\mathbf{X}^\mathsf{T}\mathbf{X}}{2}.
%\end{equation}
\begin{equation}
\left(\mathbf{X}^\mathsf{T}\mathbf{X}\right)^{+}=\frac{|\mathbf{X}^\mathsf{T}\mathbf{X}|+\mathbf{X}^\mathsf{T}\mathbf{X}}{2}, {\rm and }\left(\mathbf{X}^\mathsf{T}\mathbf{X}\right)^{-}=\frac{|\mathbf{X}^\mathsf{T}\mathbf{X}|-\mathbf{X}^\mathsf{T}\mathbf{X}}{2},
\end{equation}
%and 
%\begin{equation}
%\left(\mathbf{X}^\mathsf{T}\mathbf{X}\right)^{-}=\frac{|\mathbf{X}^\mathsf{T}\mathbf{X}|-\mathbf{X}^\mathsf{T}\mathbf{X}}{2}.
%\end{equation}
\textcolor{black}{and $\mathbf{S}^{t}$ and $\mathbf{V}^{t}$ denote the  values of $\mathbf{S}$ and $\mathbf{V}$ at the $t$-th iteration, respectively.}
The  optimization method  is summarized in Algorithm 1.    The convergence criteria for Algorithm 1 is $\|\mathbf{V}^{t+1}-\mathbf{V}^{t}\|_\infty<10^{-4}~\&~ \|\mathbf{S}^{t+1}-\mathbf{S}^{t}\|_\infty<10^{-4}$, where $\&$ is the AND operator.
Note that the constraints (i.e., $\mathbf{S}\geq 0$, $\mathbf{V}\geq 0$ and ${\rm diag}(\mathbf{S})=0$) can be naturally satisfied by the above updating rules. See Section-III-D for the detailed analysis. 

\begin{algorithm}
	\caption{Optimization algorithm for solving \eq{obj} }
	\begin{algorithmic}[1]
		\renewcommand{\algorithmicrequire}{\textbf{Input:}}
		\renewcommand{\algorithmicensure}{\textbf{Initialization:}}
		\REQUIRE   A predefined weight matrix $\mathbf{W}$, the data matrix $\mathbf{X}$, and hyper-parameters $\alpha$ and $\beta$;
		\ENSURE \textcolor{black}{Assign $\mathbf{V}$ and the off-diagonal elements of $\mathbf{S}$ with  positive random values, ${\rm diag}(\mathbf{S})=0$}; 
		%       \WHILE not converged
		%        Randomly
		%       \ENDWHILE
		%		\STATE $\mathbf{L}=\mathbf{A}-\mathbf{W}$;
		\WHILE{not converged}
		\STATE Update $\mathbf{S}$ with fixed $\mathbf{V}$  by Eq. (\ref{update-S});%  
		\STATE Update $\mathbf{V}$ with fixed $\mathbf{S}$ by Eq. (\ref{update-V});% 
		
		\ENDWHILE
		\STATE \textbf{Return} $\mathbf{V}$ and $\mathbf{S}$.
	\end{algorithmic}
	\label{alg: proposed}
\end{algorithm}

\subsection{Computational Complexity}
The sizes of $\mathbf{X}$, $\mathbf{V}$, $\mathbf{S}$, $\mathbf{X}^\mathsf{T}\mathbf{X}$\footnote{Since $\mathbf{X}^\mathsf{T}\mathbf{X}$ can be computed in advance, it can be regarded as a single matrix in computing the computational complexity of each iteration. } are $d\times n$, $n\times c$, $n\times n$, $n\times n$, respectively. 
The computational complexities for $\mathbf{VV}^\mathsf{T}$, $\left(\mathbf{X}^\mathsf{T}\mathbf{X}\right)^-\mathbf{S}$,
$\left(\mathbf{X}^\mathsf{T}\mathbf{X}\right)^+\mathbf{S}$, $\mathbf{SV}$, $\mathbf{S}^\mathsf{T}\mathbf{V}$, $\mathbf{VV}^\mathsf{T}\mathbf{V}$ are $\mathsf{O}(n^2c)$, $\mathsf{O}(n^3)$, $\mathsf{O}(n^3)$, $\mathsf{O}(n^2c)$,  $\mathsf{O}(n^2c)$, $\mathsf{O}(2c^2n)$, respectively.
Therefore, the computational complexities for step-2 and step-3 of Algorithm 1  are $\mathsf{O}(n^2c+2n^3)$ and $\mathsf{O}(2n^2c+2c^2n)$, respectively.
And the overall computational complexity of each iteration of Algorithm 1 is $\mathsf{O}(2n^3+3n^2c+2c^2n)$. 
%As a conclusion, the computational complexity for each iteration of Algorithm 1 is $\mathsf{O}$
%In addition, since Algorithm 2 has the similar computational steps as Algorithm 1, it has the same complexity as Algorithm 1.
%Since Algorithm 2 has the similar computational steps as Algorithm 1, it also have a computational complexity of $\mathsf{O}(2n^3+3n^2c+2c^2n)$ for each iteration. 

\subsection{Convergence Analysis of Algorithm 1}
\noindent\textcolor{black}{
\textbf{Theorem 1}: \emph{Algorithm \ref{alg: proposed}  has the following properties:}
%\begin{enumerate}
%	%	\item \emph{
%	%		when both $\mathbf{V}$ and $\mathbf{S}$ are initialized with non-negative matrices,  the non-negativity for them are guaranteed at each iteration, i.e., $\mathbf{V}^t\geq0,\mathbf{S}^t\geq0,\forall t$.
%	%		Moreover, 
%	%		}.
%	\item \emph{the objective value decreases (i.e., non-increases) at each iteration, and Algorithm 1 is locally convergent};
%	\item \emph{when  it is convergent, the limit point satisfies the KKT conditions, which indicates the correctness of it}. 
%	\item \emph{when both $\mathbf{V}$ and $\mathbf{S}$ are initialized with nonnegative matrices,  the non-negativity for them is guaranteed at each iteration, i.e., $\mathbf{V}^t\geq0,\mathbf{S}^t\geq0,\forall t$, where $\mathbf{V}^t$ and $\mathbf{S}^t$ denote the values of $\mathbf{V}$ and $\mathbf{S}$ at the $t$-th iteration.
%		Moreover, when $\mathbf{S}$ is initialized with a nonnegative matrix  whose diagonal elements  equal to zero, at each iteration, ${\rm diag}(\mathbf{S}^t)=0,\forall t$ is guaranteed}.
%\end{enumerate}
\begin{enumerate}
	%	\item \emph{
	%		when both $\mathbf{V}$ and $\mathbf{S}$ are initialized with non-negative matrices,  the non-negativity for them are guaranteed at each iteration, i.e., $\mathbf{V}^t\geq0,\mathbf{S}^t\geq0,\forall t$.
	%		Moreover, 
	%		}.
	\item \emph{the objective function decreases (i.e., non-increases) at each iteration, and is lower-bounded, which guarantee the convergence of the objective function};
%	\item \emph{when  it is convergent, the limit point satisfies the KKT conditions, which indicates the correctness of it}. 
	\item \emph{when $\mathbf{V}$ and the off-diagonal elements of $\mathbf{S}$ (i.e.,$\{\mathbf{S}_{ij}|\forall i,j, i\neq j\}$) are initialized with strictly positive values, i.e., $\mathbf{S}_{ij}^0>0,\forall i,j, i\neq j$, and $\mathbf{V}_{ij}^0>0,\forall i,j$, and the diagonal elements of $\mathbf{S}$ are initialized with $0$, i.e., $\mathbf{S}_{ii}^0=0,\forall i$, we have
		\begin{small}
			\begin{equation}
			\mathbf{V}_{ij}^t>0,\forall i,j,t, {\rm and }~\begin{cases}
			\mathbf{S}_{ij}^t>0,\forall i,j,t,{\rm and }~i\neq j\\
			\mathbf{S}_{ii}^t=0,\forall i,t.
			\end{cases}
			\end{equation}
		\end{small}}
\end{enumerate}}
The detailed proof of \textbf{Theorem 1} can be found in the Appendix A. 

\section{Experimental Analysis}
In this section, we conducted extensive experiments to validate the effectiveness of the proposed model. Specifically, we compared the proposed model with 19 state-of-the-art methods on 10 commonly used datasets with 4 clustering metrics. 
Moreover, we adopted the Wilcoxon rank sum test \cite{jia2019sparse} to evaluate the performance of the proposed model with a significance level of $0.05$.
%For reproducing the experiments, we provided our code at \underline{https://github.com/jyh-learning}.

%In this section, we compared the proposed models with 13 state-of-the-art models on 10 commonly used benchmark datasets with 4 clustering metrics. Specifically, the adopted data in this paper are summarized at Table \ref{tab:Data}.
%For the purpose of  reproducibility, we provide our code at :

\begin{table}[!t]
	\renewcommand{\arraystretch}{1.3}
	\begin{center}
		\caption{Descriptions of Employed Datasets} \label{tab:Data}
		\scalebox{1}{
		\begin{tabular}{cccc}
				\hline\hline
				Dataset & $\small{\#}$ Samples ($n$) & $\#$ Classes ($c$)  & $\#$ Dimensions ($d$)  \\
				\hline
				SPYBEAN	 & 683 & 19 & 35 \\
				ECOIL & 336 & 7 & 8\\
				LIBRAS & 360 & 15 & 90\\
				YEAST & 1484 & 10 & 8\\
				IONSPHERE & 351 & 2 & 34\\
				BINALPHA & 1404 & 36 & 320 \\
				IRIS & 150 & 4& 3\\
				WINE & 178 & 3 & 13\\
				ISOLET & 1560 & 26 & 617\\
				MSRA & 1799 & 12 & 256\\
				\hline\hline
		\end{tabular}}
	\end{center}
\end{table}
\subsection{Experiment Settings}
%\subsubsection{Compared Methods}
The 19 methods under comparison  are summarized as follows.
\begin{enumerate}
	\item[1,] SymNMF \cite{kuang2012symmetric,kuang2015symnmf} is a symmetric low rank decomposition of  graph, which can directly produce the clustering result.
	\item[2,] SC is a graph-based clustering method based on spectral decomposition. In this paper, we adopted the SC method presented in \cite{ng2002spectral}. %We performed k-means on 
%	\item[3,] SSC \cite{lu2016convex} is a sparse SC method that converts the non-convex SC problem into a convex problem, and uses a sparse regularizer   term to seek a ideal representation of the traditional SC.
	\item[3,] SSC \cite{lu2016convex} is a convex formulation of SC with a  sparse regularizer.
	\item[4,] PCA is a linear dimensionality reduction method.  
	\item[5,] RPCA \cite{wright2009robust,candes2011robust} solves a convex optimization problem that is  more robust to noise and outliers than the traditional PCA.
	\item[6,] GLPCA \cite{jiang2013graph} is a kind of PCA with a nonlinear graph regularization. 
	\item[7,] NMF \cite{lee2001algorithms} is a linear dimensionality reduction method that decomposes a nonnegative matrix into two nonnegative matrices with smaller sizes. 
	\item[8,] GNMF \cite{cai2011graph} is an  NMF model with graph regularization. 
	\item[9,] GMF \cite{zhang2013low} is a  graph regularized low rank matrix approximation method. 
%	\item[9,] GMF \cite{zhang2013low} is a  graph regularized low rank matrix approximation model. 
	\item[10,] GRPCA \cite{shahid2015robust} is a  graph regularized robust PCA method. 
	\item[11,] K-means is a basic clustering method.
%	\item[12,] Sparse spectral clustering, is a kind of spectral clustering method with sparsity regularization.
	\item[12,] LRR \cite{liu2010robust,liu2013robust} is subspace clustering method with a low rank constraint on the coefficient matrix. 
	\item[13,] L2-Graph \cite{peng2017constructing} is  subspace clustering method with a Frobenius norm on the coefficient matrix.
	\item[14,] CAN \cite{nie2014clustering} is a SC method with a learned graph according to the raw features. 
	\item[15,] RSS \cite{guo2015robust} simultaneously learns an  affinity matrix  and a subspace coefficient matrix. RSSA uses the affinity matrix to build the  graph. 
	\item[16,] RSSR \cite{guo2015robust} uses the coefficient matrix to build the  graph.
	\item[17,] RSSAR \cite{guo2015robust} adopts both the affinity matrix and coefficient matrix to construct the graph. 
	\item[18,] 
	\textcolor{black}{CGL \cite{egilmez2017graph} learns a graph under the connectivity, sparsity and Laplacian constraints. }
	\item[19,] To evaluate the effectiveness of the joint manner of graph construction and clustering, we made up a model termed L2-SymNMF  that first builds an L2-Graph, then applies SymNMF on that graph to produce the clustering result. 
%	\item[20,] ``Proposed'' denotes the proposed model.
%	\item[21,] ``Kernel'' denotes the proposed model with a RBF kernel. 
	
\end{enumerate}
\textcolor{black}{For all the methods involving a graph structure, we adopted the same $p$NN graph with the RBF kernel, where  $p$ was  set to ${\rm LR}({{\rm log}_2n}+1)$ \cite{von2007tutorial}, $\sigma$ equals to the mean distance between the sample and its $p$-nearest-neighbors, and ${\rm LR}({x})$ rounds $x$ to the next smaller integer.} For all the methods, the hyper-parameters were determined via exhaustive searching from $\{0.01,0.1,1,10,100,1000\}$ for fair comparisons.
For all the graph learning methods like LLR and L2-Graph, a standard SC \cite{ng2002spectral} was adopted to produce the clustering result. 
For all the data representation methods and SC-based  methods, K-means was performed on the low-dimensional  embeddings to generate the final clustering result. To exclude the influence of the randomness on K-means and initialization, we repeated each  methods 20 times and reported the mean values with standard deviations.
%\subsubsection{Evaluation Metrics}

Clustering results were evaluated by four commonly used   metrics: clustering accuracy (ACC) \cite{cai2005document}, normalized  mutual information (NMI) \cite{cai2005document}, purity (PUR) and adjust  rand index (ARI). 
%\begin{itemize}
%	\item Clustering accuracy (ACC)  first permutes the clustering result according to the ground-truth labels through a mapping function \cite{cai2005document}, then calculates the accuracy of the permuted labels. %ACC lays in the range of $[0,1]$, and the larger the better. 
%	\item NMI is the normalized version of mutual information (MI), where MI measures correlation of two distributions. In clustering, the ground-truth labels and the clustering labels act as two distributions, where NMI computes how similar  they are.% NMI lays in the range of $[0,1]$, and the larger the better. 
%	\item Purity (PUR) first assigns each cluster to a  class which is the  most  frequent in that cluster, then calculates the average accuracy of each cluster.   %PUR lays in the range of $[0,1]$, and the larger the better. 
%	\item Adjust  rand index (ARI) computes how similar the clustering result and the ground though labels are in a pairwise manner. %, and lays in the range of  $[-1,1]$, the larger the better.
%\end{itemize}
%
ACC, NMI and PUR all lay in the range of $[0,1]$, while ARI lays in the range of $[-1,1]$. A larger value indicates better clustering performance for all the metrics. 

%\textbf{NMI}
%
%\textbf{Purity}
%
%\textbf{Recall}
%
%\textbf{Precision}

%\subsubsection{Datasets Description}
We selected 10 datasets to evaluate the performance of different methods. 
The number of samples varies from hundreds to thousands and the number of classes varies from $2$ to $36$.
See the detailed information about those datasets from Table \ref{tab:Data}.

%\subsection{Toy Illustration}
%We evaluated the clustering performance 5 commonly-used clustering metrics.

%\subsection{Experimental Setting}

\subsection{Clustering Performance Analysis}

\begin{table}[!t]
	\renewcommand{\arraystretch}{1.3}
	\begin{center}
		\caption{Clustering Performance on ECOIL} \label{tab:ECOIL}
		\scalebox{0.7}{
		\begin{tabular}{ccccc}
				\hline
				\hline
				Methods & ACC & NMI  &PUR & ARI\\
				\hline\hline	
				CAN&
				$\mathbf{0.693}\downarrow$&
				$\mathbf{0.612}\bullet$&
				$\mathbf{0.824}\downarrow$&
				$\mathbf{0.560}\downarrow$\\
				
					GLPCA & 
					$0.554\pm0.067\downarrow$ &
					$\underline{0.529\pm0.035}\downarrow$ &
					$0.800\pm0.021\downarrow$ &
					$\underline{0.422\pm0.069}\downarrow$ \\
				
					PCA & 
					$0.567\pm0.061\downarrow$ &
					$0.402\pm0.028\downarrow$ &
					$0.728\pm0.016\downarrow$ &
					$0.346\pm0.054\downarrow$\\ 
				GMF &
				$0.533\pm0.055\downarrow$ &
				$0.513\pm0.025\downarrow$ &
				$0.796\pm0.027\downarrow$ &
				$0.389\pm0.057\downarrow$\\
				
				GNMF &
				$\underline{0.581\pm0.056}\downarrow$ &
				$0.473\pm0.039\downarrow$ &
				$0.760\pm0.030\downarrow$ &
				$\underline{0.458\pm0.093}\downarrow$\\
				
				GRPCA &
				$\mathbf{0.651\pm0.071}\downarrow$ &
				$\mathbf{0.611\pm0.043}\bullet$ &
				$\underline{0.812\pm0.022}\downarrow$ &
				$\mathbf{0.554\pm0.106}\bullet$ \\
				
				K-means &
				$0.553\pm0.067\downarrow$ &
				$\underline{0.532\pm0.032}\downarrow$ &
				$0.804\pm0.024\downarrow$ &
				$0.420\pm 0.024\downarrow$ \\
				
				L2-Graph &
				$0.465\pm0.032\downarrow$ &
				$0.334\pm0.017\downarrow$ &
				$0.668\pm0.019\downarrow$ &
				$0.220\pm0.034\downarrow$\\
				
				L2-SymNMF &$0.502\pm0.032\downarrow$
				&$0.350\pm0.015\downarrow$
				&$0.665\pm0.019\downarrow$
				& $0.249\pm0.0317\downarrow$\\
				
				LRR & $0.544\pm0.071\downarrow$
				& $0.524\pm0.032\downarrow$
				& $0.800\pm0.022\downarrow$
				& $0.414\pm0.085\downarrow$\\
				
				NMF & $0.558\pm0.054\downarrow$ 
				& $0.446\pm0.034\downarrow$
				& $0.750\pm0.026\downarrow$
				& $0.417\pm0.026\downarrow$\\
				
				RPCA & $0.547\pm0.067\downarrow$ &$0.519\pm0.024\downarrow$ & $\underline{0.809\pm0.019}\downarrow$ & $0.405\pm0.073\downarrow$\\
				
				RSSAR & $0.507\pm0.034\downarrow$ & $0.431\pm0.021\downarrow$ & $0.761\pm0.019\downarrow$ & $0.315\pm0.040\downarrow$\\
				
				RSSR & $0.486\pm0.027\downarrow$ & $0.355\pm0.015\downarrow$ &
				$0.708\pm0.015\downarrow$ & $0.270\pm0.021\downarrow$\\
				
				RSSA & $0.515\pm0.025\downarrow$ & $0.429\pm0.020\downarrow$ &  $0.767\pm0.018\downarrow$ & $0.331\pm0.044\downarrow$\\
				
				SC & $0.544\pm0.044\downarrow$ & $0.508\pm0.029\downarrow$ & $\mathbf{0.818\pm0.024}\downarrow$ & $0.381\pm0.024\downarrow$ \\
				
				SSC & $\underline{0.601\pm0.036}\downarrow$ & $0.492\pm 0.034\downarrow$ &
				$0.797\pm0.037\downarrow$ & $0.357\pm0.054\downarrow$\\
				
				\textcolor{black}{CGL} & $0.498\pm0.005\downarrow$ & $0.468\pm0.08\downarrow$ & $0.759\pm0.005\downarrow$ & $0.301\pm0.005\downarrow$\\
				
				SymNMF & $0.571\pm0.070\downarrow$ & $0.506\pm0.055\downarrow$ & $0.761\pm0.056\downarrow$ & $0.401\pm0.056\downarrow$\\
				
				\hline
				\textbf{Proposed} & $\cellcolor[gray]{0.8}0.735\pm0.094$ & $\cellcolor[gray]{0.8}0.626\pm0.046$ & $0.836\pm0.019\cellcolor[gray]{0.8}$ & $0.632\pm0.130\cellcolor[gray]{0.8}$\\
%				\cdashline{1-5}
%				Kernel (Gau) & 
%				$0.736\pm	0.063$ &
%				$0.615\pm	0.041$ &
%				$0.835\pm	0.021$ &
%				$0.619\pm	0.101$ \\
				
%				Kernel (POL) & $\mathbf{0.95\pm0.032}$ & $\mathbf{0.95\pm0.032}$ & $\mathbf{0.95\pm0.032}$ & $\mathbf{0.95\pm0.032}$\\
				\hline\hline
		\end{tabular}}
	\end{center}
	\begin{tablenotes}
		\small
		\item \textcolor{black}{The highest value is highlighted by \colorbox{lightgray}{gray},  the second and third highest values are marked by \textbf{bold},  and the fourth and the fifth highest values are  \underline{underlined}. 
		%		$\downarrow$ means the proposed method is significantly better than the corresponding method, $\bullet$ indicates the compared method has no significance difference with the proposed one, and $\diamondsuit$ shows that the compared method is significantly better than the proposed one.
		$\downarrow$ and $\diamondsuit$ indicate the proposed method is significantly better/worse, respectively, than the compared methods according to the  Wilcoxon rank sum test. 
		Moreover, $\bullet$ means there is no significant difference between the proposed model and the compared methods.}
	\end{tablenotes}
\end{table}
\begin{table}[!t]
	\renewcommand{\arraystretch}{1.3}
	\begin{center}
		\caption{Clustering Performance on YEAST} \label{tab:yeast}
		\scalebox{0.7}{
			\begin{tabular}{ccccc}
				\hline
				\hline
				Methods & ACC & NMI  &PUR & ARI\\
				\hline\hline		
				CAN &
				$\underline{0.4218}\downarrow$&
				$0.1451\downarrow$&
				$0.4299\downarrow$&
				$0.0848\downarrow$\\
				
				GLPCA & 
				$0.378\pm0.024\downarrow$&
				$0.248\pm0.012\downarrow$&
				$\underline{0.532\pm0.011}\diamondsuit$&
				$0.148\pm0.010\downarrow$\\
				
				PCA &
				$0.359\pm0.019\downarrow$&
				$0.233\pm0.013\downarrow$&
				$0.498\pm0.023\downarrow$&
				$0.134\pm0.013\downarrow$\\
				
				GMF &
				$0.366\pm0.025\downarrow$&
				$0.239\pm0.008\downarrow$&
				$0.520\pm0.006\diamondsuit$&
				$0.138\pm0.008\downarrow$\\
				
				GNMF &
				$0.323\pm0.0322\downarrow$&
				$0.190\pm0.0294\downarrow$&
				$0.466\pm0.0211\downarrow$&
				$0.101\pm0.0266\downarrow$\\
				
				GRPCA &
				$\mathbf{0.451\pm0.035}\bullet$&
				$\mathbf{0.260\pm0.031}\bullet$&
				$\mathbf{0.535\pm0.009}\diamondsuit$&
				$\mathbf{0.171\pm0.031}\bullet$\\
				
				K-means &
				$0.378\pm0.024\downarrow$&
				$\underline{0.249\pm0.015}\downarrow$&
				$0.531\pm0.013\diamondsuit$&
				$0.148\pm0.013\downarrow$\\
				
				L2-Graph  &
				$0.358\pm0.015\downarrow$&
				$0.208\pm0.007\downarrow$&
				$0.502\pm0.004\downarrow$&
				$0.118\pm0.008\downarrow$\\
				
				L2-SymNMF &
				$0.361\pm0.021\downarrow$&
				$0.222\pm0.009\downarrow$&
				$0.490\pm0.013\downarrow$&
				$0.123\pm0.013\downarrow$\\
				
				LRR &
				$0.381\pm0.022\downarrow$&
				$\underline{0.251\pm0.006}\downarrow$&
				$\underline{0.532\pm0.010}\diamondsuit$&
				${0.150\pm0.007}\downarrow$\\
				
				NMF &
				$0.334\pm0.027\downarrow$&
				$0.203\pm0.024\downarrow$&
				$0.484\pm0.024\downarrow$&
				$0.113\pm0.024\downarrow$\\
				
				RPCA &
				$0.384\pm0.021\downarrow$&
				$\underline{0.249\pm0.009}\downarrow$&
				$\underline{0.532\pm0.013}\diamondsuit$&
				${0.150\pm0.009}\downarrow$\\
				
				RSSAR &
				$0.331\pm0.010\downarrow$&
				$0.211\pm0.007\downarrow$&
				$0.510\pm0.004\downarrow$&
				$0.122\pm0.006\downarrow$\\
				
				RSSR &
				$0.333\pm0.016\downarrow$&
				$0.208\pm0.008\downarrow$&
				$0.511\pm0.004\downarrow$&
				$0.118\pm0.007\downarrow$\\
				
				RSSA &
				$0.327\pm0.014\downarrow$&
				$0.210\pm0.004\downarrow$&
				$0.512\pm0.002\downarrow$&
				$0.116\pm0.004\downarrow$\\
				
				SC &
				$0.360\pm0.011\downarrow$&
				$0.245\pm0.007\downarrow$&
				$\mathbf{0.535\pm0.008}\diamondsuit$&
				$\underline{0.151\pm0.008}\downarrow$\\
				
				SSC &
				$0.383\pm0.021\downarrow$&
				$\underline{0.249\pm0.012}\downarrow$&
				$0.527\pm0.010\bullet$&
				$0.149\pm0.010\downarrow$\\
				
				\textcolor{black}{CGL} & 
				$\mathbf{0.457\pm0.022}\bullet$&
				$0.387\pm0.013\diamondsuit\cellcolor[gray]{0.8}$&
				$0.662\pm0.014\diamondsuit\cellcolor[gray]{0.8}$&
				$0.266\pm0.019\diamondsuit\cellcolor[gray]{0.8}$\\
				
				SymNMF & 
				$\underline{0.404\pm0.045}\downarrow$&
				$0.208\pm0.040\downarrow$&
				$0.426\pm0.037\downarrow$&
				$\underline{0.152\pm0.037}\downarrow$\\

				\hline
				\textbf{Proposed} 
				&$0.467\pm0.048\cellcolor[gray]{0.8}$
				&$\mathbf{0.273\pm0.021}$
				&$0.518\pm0.019$
				&$\mathbf{0.186\pm0.048}$\\

%				\hline
%				\textbf{Proposed} 
%				&$0.467\pm0.048\cellcolor[gray]{0.8}$
%				&$0.273\pm0.021\cellcolor[gray]{0.8}$
%				&$0.518\pm0.019$
%				&$0.186\pm0.048\cellcolor[gray]{0.8}$\\
				
%				\cdashline{1-5}
%				Kernel (Gau) & 
%				$0.462\pm0.038$&
%				$0.274\pm0.017$&
%				$0.529\pm0.015$&
%				$0.191\pm0.056$\\
				
%				Kernel (POL) & $\mathbf{0.95\pm0.032}$ & $\mathbf{0.95\pm0.032}$ & $\mathbf{0.95\pm0.032}$ & $\mathbf{0.95\pm0.032}$\\
				\hline\hline
			\end{tabular}}
		\end{center}
\end{table}
\begin{table}[!t]
\renewcommand{\arraystretch}{1.3}
\begin{center}
\caption{Clustering Performance on IONSPHERE} \label{tab:Ionsphere}
\scalebox{0.7}{
\begin{tabular}{ccccc}
					\hline
					\hline
					Methods & ACC & NMI  &PUR & ARI\\
					\hline\hline	
					CAN &
					$0.547\downarrow$&
					$0.054\downarrow$&
					$0.641\downarrow$&
					$-0.0455\downarrow$\\
					
					GLPCA &
					$0.658\pm0.001\downarrow$&
					$0.139\pm0.074\downarrow$&
					$0.675\pm0.026\downarrow$&
					$0.095\pm0.001\downarrow$\\
					
					PCA &
					$0.514\pm0.0035\downarrow$&
					$0.021\pm0.0051\downarrow$&
					$0.641\pm0\downarrow$&
					$-0.026\pm0.0018\downarrow$\\
					
					GMF &
					$0.704\pm0.024\downarrow$&
					$0.104\pm0.007\downarrow$&
					$0.704\pm0.024\downarrow$&
					$0.135\pm0\downarrow$\\
					
						GNMF& --- & --- & --- & ---\\
					
					GRPCA &
					$\mathbf{0.727\pm0.032}\downarrow$&
					$\underline{0.149\pm0.056}\downarrow$&
					$\mathbf{0.727\pm0.032}\downarrow$&
					$\mathbf{0.177\pm0}\downarrow$\\
					
					K-means &
					$0.708\pm0.015\downarrow$&
					$0.124\pm0.028\downarrow$&
					$0.708\pm0.015\downarrow$&
					$0.167\pm0.015\downarrow$\\
					
					L2-Graph & 
					$0.555\pm0\downarrow$&
					$\underline{0.153\pm0}\downarrow$&
					$0.641\pm0\downarrow$&
					$-0.017\pm0.002\downarrow$\\
					
					L2-SymNMF &
					$0.671\pm0.106\downarrow$&
					$\mathbf{0.199\pm0.109}\bullet$&
					$0.706\pm0.061\downarrow$&
					$0.145\pm0.146\downarrow$\\
					
					LRR &
					$\underline{0.711\pm0.001}\downarrow$&
					$0.130\pm0.001\downarrow$&
					$\underline{0.711\pm0.001}\downarrow$&
					$\underline{0.176\pm0.002}\downarrow$\\
					
					NMF& --- & --- & --- & ---\\
					
					RPCA &
					$\underline{0.711\pm0.001}\downarrow$&
					$0.130\pm0.001\downarrow$&
					$\underline{0.711\pm0.001}\downarrow$&
					$\underline{0.176\pm0.002}\downarrow$\\
					
					RSSRA &
					$0.529\pm0\downarrow$&
					$0.131\pm0\downarrow$&
					$0.641\pm0\downarrow$&
					$-0.0336\pm0\downarrow$\\
					
					RSSR &
					$0.555\pm0\downarrow$&
					$0.117\pm0\downarrow$&
					$0.641\pm0\downarrow$&
					$-0.016\pm0\downarrow$\\
					
					RSSA&
					$0.529\pm0\downarrow$&
					$0.131\pm0\downarrow$&
					$0.641\pm0\downarrow$&
					$-0.033\pm0\downarrow$\\
					
					SC &
					$0.643\pm0\downarrow$&
					$0.046\pm0\downarrow$&
					$0.643\pm0\downarrow$&
					$0.077\pm0\downarrow$\\
					
					SSC &
					$\mathbf{0.766\pm0}\downarrow$&
					$\mathbf{0.205\pm0}\downarrow$&
					$\mathbf{0.766\pm0}\downarrow$&
					$\mathbf{0.281\pm0}\downarrow$\\
					
					\textcolor{black}{
					CGL} & 
					$0.661\pm0.000\downarrow$&
					$0.066\pm0.000\downarrow$&
					$0.661\pm0.000\downarrow$&
					$0.100\pm0.000\downarrow$\\
					
					SymNMF & 
					$0.652\pm0.080\downarrow$&
					$0.111\pm0.071\downarrow$&
					$0.679\pm0.051\downarrow$&
					$0.110\pm0.051\downarrow$\\

%					\cdashline{1-5}
					\hline
					\textbf{Proposed}
					&$0.787\pm0.066\cellcolor[gray]{0.8}$
					&$0.256\pm0.065\cellcolor[gray]{0.8}$
					&$0.790\pm0.056\cellcolor[gray]{0.8}$
					&$0.339\pm0.123\cellcolor[gray]{0.8}$\\
					
%					\cdashline{1-5}
%					Kernel (Gau) & 
%					$0.823\pm0.060$&
%					$0.314\pm0.107$&
%					$0.823\pm	0.060$&
%					$0.418\pm	0.138$\\
					
					\hline\hline
\end{tabular}}
\end{center}
%\begin{tablenotes}
%	\small
%	\item $\blacktriangleright$
%	Since 
%	NMF and and GNMF can only be applied to  nonnegative matrix, and 
%	 IONSHPERE contains several negative features,  NMF and GNMF cannot be applied to  IONSHPERE.
%\end{tablenotes}
\end{table}
		
		\begin{table}[!t]
			\renewcommand{\arraystretch}{1.3}
			\begin{center}
				\caption{Clustering Performance on BINALPHA} \label{tab:BinAlpha}
				\scalebox{0.7}{
					\begin{tabular}{ccccc}
						\hline
						\hline
						Methods & ACC & NMI  &PUR & ARI\\
						\hline\hline	
						CAN &
						$0.332\downarrow$&
						$0.445\downarrow$&
						$0.363\downarrow$&
						$0.091\downarrow$\\
						
						GLPCA &
						$0.409\pm0.022\downarrow$&
						$0.570\pm0.010\downarrow$&
						$0.439\pm0.018\downarrow$&
						$0.268\pm0.015\downarrow$\\
						
						PCA &
						$0.352\pm0.022\downarrow$&
						$0.512\pm0.015\downarrow$&
						$0.376\pm0.021\downarrow$&
						$0.210\pm0.017\downarrow$\\
						
						GMF &
						$0.449\pm0.016\downarrow$&
						$0.606\pm0.008\downarrow$&
						$0.487\pm0.015\downarrow$&
						$0.307\pm0.013\downarrow$\\
						
						GNMF &
						$0.366\pm0.018\downarrow$&
						$0.523\pm0.013\downarrow$&
						$0.392\pm0.017\downarrow$&
						$0.218\pm0.014\downarrow$\\
						
						GRPCA &
						${0.458\pm0.020}\downarrow$&
						$\mathbf{0.619\pm0.009}\bullet$&
						${0.490\pm0.017}\downarrow$&
						$\underline{0.328\pm0.012}\downarrow$\\
						
						K-means &
						$0.394\pm0.015\downarrow$&
						$0.564\pm0.010\downarrow$&
						$0.425\pm0.016\downarrow$&
						$0.259\pm0.016\downarrow$\\
						
						L2-Graph &
						$0.345\pm0.010\downarrow$&
						$0.481\pm0.007\downarrow$&
						$0.374\pm0.009\downarrow$&
						$0.192\pm0.008\downarrow$\\
						
						L2-SymNMF &
						$0.312\pm0.013\downarrow$&
						$0.452\pm0.008\downarrow$&
						$0.352\pm0.012\downarrow$&
						$0.172\pm0.009\downarrow$\\
						
						LLR &
						$0.414\pm0.025\downarrow$&
						$0.573\pm0.010\downarrow$&
						$0.443\pm0.022\downarrow$&
						$0.272\pm0.015\downarrow$\\
						
						NMF &
						$0.357\pm0.017\downarrow$&
						$0.518\pm0.013\downarrow$&
						$0.386\pm0.017\downarrow$&
						$0.211\pm0.017\downarrow$\\
						
						RPCA &
						$0.412\pm0.017\downarrow$&
						$0.572\pm0.008\downarrow$&
						$0.442\pm0.014\downarrow$&
						$0.273\pm0.016\downarrow$\\
						
						RSSAR & 
						$0.121\pm0.004\downarrow$&
						$0.185\pm0.006\downarrow$&
						$0.131\pm0.003\downarrow$&
						$0.023\pm0.004\downarrow$\\
						
						RSSR & 
						$0.203\pm0.011\downarrow$&
						$0.309\pm0.006\downarrow$&
						$0.216\pm0.010\downarrow$&
						$0.075\pm0.005\downarrow$\\
						
						RSSA & 
						$0.116\pm0.003\downarrow$&
						$0.185\pm0.007\downarrow$&
						$0.124\pm0.003\downarrow$&
						$0.014\pm0.002\downarrow$\\
						
						SC &
						$\mathbf{0.477\pm0.018}\bullet$&
						$\underline{0.615\pm0.008}\downarrow$&
						$\mathbf{0.506\pm0.014}\downarrow$&
						$\mathbf{0.329\pm0.014}\downarrow$\\
						
						SSC &
						$\mathbf{0.466\pm0.020}\downarrow$&
						${0.613\pm0.010}\downarrow$&
						$\underline{0.501\pm0.018}\downarrow$&
						$\underline{0.327\pm0.016}\downarrow$\\
						
						\textcolor{black}{CGL} & 
						$\underline{0.467\pm0.018}\downarrow$&
						$\underline{0.615\pm0.006}\downarrow$&
						$\underline{0.502\pm0.012}\downarrow$&
						${0.320\pm0.010}\downarrow$\\
						
						SymNMF & 
						$\underline{0.465\pm0.019}\downarrow$&
						$\mathbf{0.619\pm0.009}\bullet$&
						$\mathbf{0.504\pm0.016}\downarrow$&
						$\mathbf{0.335\pm0.016}\bullet$\\
%						\cdashline{1-5}
						\hline
						\textbf{Proposed} & 
						$0.484\pm0.018$\cellcolor[gray]{0.8}&
						\cellcolor[gray]{0.8}$0.622\pm0.010$&
						$0.516\pm0.017$\cellcolor[gray]{0.8}&
						$0.337\pm0.015$\cellcolor[gray]{0.8}\\
						
%						\cdashline{1-5}
%						Kernel (Gau) & 
%						$0.487\pm0.017$&
%						$0.625\pm0.008$&
%						$0.516\pm	0.013$&
%						$0.343\pm0.013$\\
						
						\hline\hline
					\end{tabular}}
				\end{center}
			\end{table}
\begin{table}[!t]
	\renewcommand{\arraystretch}{1.3}
	\begin{center}
		\caption{Clustering Performance on IRIS} \label{tab:IRIS}
		\scalebox{0.7}{
			\begin{tabular}{ccccc}
				\hline
				\hline
				Methods & ACC & NMI  &PUR & ARI\\
				\hline\hline
				CAN &
				$0.693\downarrow$&
				$0.596\downarrow$&
				$0.693\downarrow$&
				$0.560\downarrow$\\

				GLPCA &
				$0.520\pm0.083\downarrow$&
				$0.228\pm0.014\downarrow$&
				$0.556\pm0.042\downarrow$&
				$0.148\pm0.137\downarrow$\\

				PCA &
				$0.722\pm0.121\downarrow$&
				$0.597\pm0.011\downarrow$&
				$0.760\pm	0.062\downarrow$&
				$0.528\pm	0.053\downarrow$\\

				GMF &
				$0.821\pm0.141\downarrow$&
				$0.704\pm0.089\downarrow$&
				$0.845\pm0.091\downarrow$&
				$0.663\pm0.125\downarrow$\\
				
				GNMF &
				$0.776\pm0.067\downarrow$&
				$0.626\pm0.034\downarrow$&
				$0.780\pm0.057\downarrow$&
				$0.567\pm0.053\downarrow$\\
				
				GRPCA &
				$\mathbf{0.884\pm0.085}\bullet$&
				$\mathbf{0.781\pm0.047}\bullet$&
				$\mathbf{0.892\pm0.053}\bullet$&
				$\mathbf{0.737\pm0.070}\bullet$\\
				
				K-means &
				$0.757\pm0.183\downarrow$&
				$0.668\pm0.101\downarrow$&
				$0.811\pm0.108\downarrow$&
				$0.615\pm0.108\downarrow$\\
				
				L2 Graph &
				$0.830\pm0.019\downarrow$&
				$0.663\pm0.015\downarrow$&
				$0.830\pm0.019\downarrow$&
				$0.621\pm0.022\downarrow$\\
				
				L2-SymNMF &
				$0.807\pm0.006\downarrow$&
				$0.638\pm0.026\downarrow$&
				$0.807\pm0.006\downarrow$&
				$0.585\pm0.009\downarrow$\\
				
				LRR &
				$\underline{0.852\pm0.115}\downarrow$&
				$\underline{0.722\pm0.065}\downarrow$&
				$\underline{0.867\pm0.068}\downarrow$&
				$\underline{0.693\pm0.093}\downarrow$\\
				
				NMF & 
				$0.719\pm0.112\downarrow$&
				$0.614\pm0.098\downarrow$&
				$0.738\pm0.087\downarrow$&
				$0.545\pm0.087\downarrow$\\
				
				RPCA & 
				${0.838\pm0.125}\downarrow$&
				${0.711\pm0.077}\downarrow$&
				${0.856\pm0.081}\downarrow$&
				${0.677\pm0.109}\downarrow$\\
				
				RSSAR & 
				$0.510\pm0.026\downarrow$&
				$0.419\pm0.069\downarrow$&
				$0.615\pm0.022\downarrow$&
				$0.312\pm0.065\downarrow$\\
				
				RSSR &
				$0.826\pm0\downarrow$&
				$0.652\pm0.001\downarrow$&
				$0.826\pm0\downarrow$&
				$0.610\pm0.001\downarrow$\\
				
				RSSA & 
				$0.519\pm0.029\downarrow$&
				$0.579\pm0\downarrow$&
				$0.666\pm0\downarrow$&
				$0.442\pm0.009\downarrow$\\
				
				SC & 
				$0.461\pm0.002\downarrow$&
				$0.298\pm0.004\downarrow$&
				$0.561\pm0.002\downarrow$&
				$0.187\pm0.002\downarrow$\\
				
				SSC &
				$\underline{0.854\pm0.115}\downarrow$&
				$\underline{0.725\pm0.066}\downarrow$&
				$\underline{0.869\pm0.069}\downarrow$&
				$\underline{0.695\pm0.094}\downarrow$\\
				
				\textcolor{black}{CGL} & 
				$0.920\pm0.000\bullet$\cellcolor[gray]{0.8}&
				$0.796\pm0.000\bullet$\cellcolor[gray]{0.8}&
				$0.920\pm0.000\bullet$\cellcolor[gray]{0.8}&
				$0.785\pm0.000\diamondsuit$\cellcolor[gray]{0.8}\\
				
				SymNMF & 
				$0.665\pm0.126\downarrow$&
				$0.417\pm0.170\downarrow$&
				$0.681\pm0.116\downarrow$&
				$0.384\pm0.116\downarrow$\\
				
%				\cdashline{1-5}
%				\hline
%				\textbf{Proposed} & 
%				$0.903\pm0.005$\cellcolor[gray]{0.8}&
%				$0.786\pm0.012$\cellcolor[gray]{0.8}&
%				$0.903\pm0.005$\cellcolor[gray]{0.8}&
%				$0.751\pm0.011$\cellcolor[gray]{0.8}\\
				\hline
				\textbf{Proposed} & 
				$\mathbf{0.903\pm0.005}$&
				$\mathbf{0.786\pm0.012}$&
				$\mathbf{0.903\pm0.005}$&
				$\mathbf{0.751\pm0.011}$\\
				
%				\cdashline{1-5}
%				Kernel (Gau) & 
%				$0.914\pm0.020$&
%				$0.796\pm0.031$&
%				$0.914\pm0.020$&
%				$0.776\pm0.047$\\
				
				\hline\hline
			\end{tabular}}
		\end{center}
	\end{table}
	
	\begin{table}[!t]
		\renewcommand{\arraystretch}{1.3}
		\begin{center}
			\caption{Clustering Performance on WINE} \label{tab:WINE}
			\scalebox{0.7}{
				\begin{tabular}{ccccc}
					\hline
					\hline
					Methods & ACC & NMI  &PUR & ARI\\
					\hline\hline
					CAN &
					$0.668\downarrow$ &
					$0.521\downarrow$ &
					$0.707\downarrow$&
					$0.490\downarrow$\\
					
					GLPCA &
					$0.923\pm0.005\downarrow$&
					$0.767\pm0.012\downarrow$&
					$0.923\pm0.005\downarrow$&
					$0.773\pm0.015\downarrow$\\
					
					PCA &
					$\underline{0.929\pm0.002}\downarrow$&
					$0.764\pm0.006\downarrow$&
					$0.929\pm0.002\downarrow$&
					$0.789\pm0.007\downarrow$\\

					GMF &
					$0.921\pm0\downarrow$&
					$0.754\pm0\downarrow$&
					$0.921\pm0\downarrow$&
					$0.768\pm0\downarrow$\\
					
					GNMF &
					$0.918\pm0.009\downarrow$&
					$0.741\pm0.025\downarrow$&
					$0.918\pm0.009\downarrow$&
					$0.761\pm0.025\downarrow$\\
					
					GRPCA &
					$0.923\pm0.005\downarrow$&
					$0.776\pm0.004\downarrow$&
					$0.923\pm0.005\downarrow$&
					$0.774\pm0.014\downarrow$\\
					
					K-means &
					$0.923\pm0.005\downarrow$&
					$0.766\pm0.011\downarrow$&
					$0.923\pm0.005\downarrow$&
					$0.773\pm0.005\downarrow$\\
					
					L2-Graph & 
					$\mathbf{0.946\pm0.002}\downarrow$&
					$\mathbf{0.808\pm0.009}\downarrow$&
					$\mathbf{0.946\pm0.002}\downarrow$&
					$\mathbf{0.837\pm0.008}\downarrow$\\
					
					L2-SymNMF &
					$0.874\pm0.134\downarrow$&
					$0.713\pm0.155\downarrow$&
					$0.881\pm0.117\downarrow$&
					$0.722\pm0.183\downarrow$\\
					
					LRR & 
					$0.924\pm0.004\downarrow$&
					$0.769\pm0.009\downarrow$&
					$0.924\pm0.004\downarrow$&
					$0.776\pm0.012\downarrow$\\
					
					NMF & 
					$0.854\pm0.107\downarrow$&
					$0.654\pm0.129\downarrow$&
					$0.857\pm0.099\downarrow$&
					$0.648\pm0.099\downarrow$\\
					
					RPCA &
					$0.926\pm0.008\downarrow$&
					$0.779\pm0.024\downarrow$&
					$0.926\pm0.008\downarrow$&
					$0.782\pm0.022\downarrow$\\
					
					RSSAR & 
					$0.927\pm0.088\downarrow$&
					$\underline{0.797\pm0.075}\downarrow$&
					$\underline{0.930\pm0.073}\downarrow$&
					$\underline{0.817\pm0.103}\downarrow$\\
					
					RSSR &
					$\underline{0.935\pm0.002}\downarrow$&
					$\underline{0.801\pm0.006}\downarrow$&
					$\underline{0.935\pm0.002}\downarrow$&
					$\underline{0.809\pm0.007}\downarrow$\\
					
					RSSA &
					$0.921\pm0\downarrow$&
					$0.741\pm0\downarrow$&
					$0.921\pm0\downarrow$&
					$0.770\pm0\downarrow$\\
					
					SC &
					$\mathbf{0.949\pm0}\downarrow$&
					$\mathbf{0.829\pm0}\downarrow$&
					$\mathbf{0.949\pm0}\downarrow$&
					$\mathbf{0.848\pm0}\downarrow$\\
					
					SSC & 
					$0.903\pm0.088\downarrow$&
					$0.766\pm0.086\downarrow$&
					$0.909\pm0.070\downarrow$&
					$0.765\pm0.105\downarrow$\\
					
					\textcolor{black}{CGL} & 
					$0.902\pm0.003\downarrow$&
					$0.705\pm0.006\downarrow$&
					$0.902\pm0.003\downarrow$&
					$0.719\pm0.007\downarrow$\\
					
					SymNMF & 
					$0.852\pm0.128\downarrow$&
					$0.667\pm0.147\downarrow$&
					$0.856\pm0.116\downarrow$&
					$0.667\pm0.116\downarrow$\\
					
%					\cdashline{1-5}
					\hline
					\textbf{Proposed} & 
					$0.959\pm0.004$\cellcolor[gray]{0.8}&
					$0.852\pm0.109$\cellcolor[gray]{0.8}&
					$0.959\pm0.004$\cellcolor[gray]{0.8}&
					$0.881\pm0.127$\cellcolor[gray]{0.8}\\
					
%					\cdashline{1-5}
%					Kernel (Gau) & 
%					$0.958\pm0.008$&
%					$0.846\pm0.021$&
%					$0.958\pm0.008$&
%					$0.869\pm0.025$\\
					
					\hline\hline
				\end{tabular}}
			\end{center}
		\end{table}
		\begin{table}[!t]
			\renewcommand{\arraystretch}{1.3}
			\begin{center}
				\caption{Clustering Performance on MSRA} \label{tab:MSRA}
				\scalebox{0.7}{
					\begin{tabular}{ccccc}
						\hline
						\hline
						Methods & ACC & NMI  &PUR & ARI\\
						\hline\hline
						CAN &
						$0.533\downarrow$&
						$0.602\downarrow$&
						$0.537\downarrow$&
						$0.313\downarrow$\\
						
						GLPCA & 
						$0.514\pm0.033\downarrow$&
						$0.583\pm0.032\downarrow$&
						$0.541\pm0.030\downarrow$&
						$0.349\pm0.045\downarrow$\\
						
						PCA &
						$0.525\pm0.030\downarrow$&
						$0.585\pm0.029\downarrow$&
						$0.551\pm0.027\downarrow$&
						${0.378\pm0.041}\downarrow$\\

						GMF &
						$0.495\pm0.037\downarrow$&
						$0.553\pm0.031\downarrow$&
						$0.522\pm0.029\downarrow$&
						$0.324\pm0.043\downarrow$\\
						
						GNMF &
						$0.497\pm0.034\downarrow$&
						$0.559\pm0.032\downarrow$&
						$0.525\pm0.029\downarrow$&
						$0.345\pm0.036\downarrow$\\
						
						GRPCA &
						${0.546\pm0.049}\bullet$&
						$\underline{0.667\pm0.034}\bullet$&
						$0.584\pm0.041\bullet$&
						$\underline{0.409\pm0.044}\bullet$\\
						
						K-means &
						$0.497\pm0.040\downarrow$&
						$0.573\pm0.033\downarrow$&
						$0.529\pm0.033\downarrow$&
						$0.342\pm0.033\downarrow$\\
						
						L2-Graph & 
						$0.539\pm0.037\bullet$&
						$\underline{0.646\pm0.048}\downarrow$&
						${0.587\pm0.033}\bullet$&
						$0.369\pm0.059\downarrow$\\
						
						L2-SymNMF & 
						$\mathbf{0.566\pm0.039}\bullet$&
						$\mathbf{0.688\pm0.026}\diamondsuit$&
						$\mathbf{0.610\pm0.031}\diamondsuit$&
						$\mathbf{0.463\pm0.043}\diamondsuit$\\
						
						LRR &
						$0.512\pm0.034\downarrow$&
						$0.594\pm0.031\downarrow$&
						$0.548\pm0.027\downarrow$&
						$0.364\pm0.041\downarrow$\\
						
						NMF & 
						$0.484\pm0.035\downarrow$&
						$0.539\pm0.031\downarrow$&
						$0.507\pm0.029\downarrow$&
						$0.321\pm0.029\downarrow$\\
						
						RPCA &
						$0.520\pm0.036\downarrow$&
						$0.589\pm0.029\downarrow$&
						$0.547\pm0.027\downarrow$&
						$0.359\pm0.032\downarrow$\\
						
						RSSAR & 
						$\underline{0.549\pm0.041}\bullet$&
						$0.565\pm0.020\downarrow$&
						$\underline{0.593\pm0.031}\bullet$&
						$0.335\pm0.042\downarrow$\\
						
						RSSR &
						$0.642\pm0.029\diamondsuit$\cellcolor[gray]{0.8}&
						$0.736\pm0.027\diamondsuit$\cellcolor[gray]{0.8}&
						$0.690\pm0.029\diamondsuit$\cellcolor[gray]{0.8}&
						$0.548\pm0.038\diamondsuit$\cellcolor[gray]{0.8}\\
						
						RSSA &
						$0.509\pm0.020\downarrow$&
						$0.485\pm0.013\downarrow$&
						$0.542\pm0.020\downarrow$&
						$0.275\pm0.021\downarrow$\\
						
						SC & 
						$0.447\pm0.027\downarrow$&
						$0.546\pm0.022\downarrow$&
						$0.484\pm0.025\downarrow$&
						$0.292\pm0.025\downarrow$\\
						
						SSC & 
						$0.495\pm0.046\downarrow$&
						$0.566\pm0.039\downarrow$&
						$0.532\pm0.042\downarrow$&
						$0.335\pm0.049\downarrow$\\
						
						\textcolor{black}{CGL} & 
						$\mathbf{0.559\pm0.022}\bullet$&
						$0.641\pm0.011\downarrow$&
						$\mathbf{0.595\pm0.019}\bullet$&
						$\underline{0.427\pm0.017}\bullet$\\
						
						SymNMF & 
						$0.457\pm0.027\downarrow$&
						$0.569\pm0.027\downarrow$&
						$0.502\pm0.029\downarrow$&
						$0.333\pm0.029\downarrow$\\
						
%						\cdashline{1-5}
						\hline
						\textbf{Proposed} & 
						$\underline{0.557\pm0.045}$&
						$\mathbf{0.673\pm0.032}$&
						$\underline{0.588\pm0.038}$&
						$\mathbf{0.432\pm0.041}$\\
%						
%						\cdashline{1-5}
%						Kernel (Gau) & 
%						$0.553\pm0.041$&
%						$0.665\pm0.031$&
%						$0.590\pm0.034$&
%						$0.435\pm0.048$\\
						
						\hline\hline
					\end{tabular}}
				\end{center}
			\end{table}
			\begin{table}[!t]
				\renewcommand{\arraystretch}{1.3}
				\begin{center}
					\caption{Clustering Performance on ISOLET} \label{tab:1NN}
					\scalebox{0.7}{
						\begin{tabular}{ccccc}
							\hline
							\hline
							Methods & ACC & NMI  &PUR & ARI\\
							\hline\hline
							CAN & 
							$0.553\downarrow$ &
							$0.733\downarrow$ &
							$0.590\downarrow$ &
							$0.416\downarrow$ \\
							
							GLPCA &
							$0.589\pm0.037\downarrow$ &
							$0.744\pm0.016\downarrow$ &
							$0.632\pm0.028\downarrow$ &
							$0.528\pm0.033\downarrow$ \\
							
							PCA &
							$0.635\pm0.047\bullet$\cellcolor[gray]{0.8} &
							$\underline{0.750\pm0.024}\downarrow$ &
							$\mathbf{0.675\pm0.040}\bullet$ &
							$\mathbf{0.563\pm0.035}\bullet$ \\
							
							GNMF& --- & --- & --- & ---\\
							GRPCA &
							$\underline{0.593\pm0.034}\downarrow$ &
							$\mathbf{0.752\pm0.011}\downarrow$ &
							$\underline{0.633\pm0.020}\downarrow$ &
							$\mathbf{0.540\pm0.018}\downarrow$ \\
							
							K-means &
							$0.588\pm0.034\downarrow$ &
							$0.742\pm0.019\downarrow$ &
							$0.631\pm0.031\downarrow$ &
							$0.524\pm0.031\downarrow$ \\
							
							L2-Graph & 
							$0.547\pm0.036\downarrow$ &
							$0.702\pm0.016\downarrow$ &
							$0.587\pm0.026\downarrow$ &
							$0.472\pm0.033\downarrow$ \\
							
							L2-SymNMF &
							$0.523\pm0.019\downarrow$ &
							$0.682\pm0.014\downarrow$ &
							$0.585\pm0.018\downarrow$ &
							$0.443\pm0.024\downarrow$ \\
							
							LRR &
							$\mathbf{0.601\pm0.024}\downarrow$ &
							$\underline{0.749\pm0.013}\downarrow$ &
							$\underline{0.639\pm0.023}\downarrow$ &
							$\underline{0.536\pm0.021}\downarrow$ \\
							
							NMF& --- & --- & --- & ---\\
							RPCA &
							$\underline{0.593\pm0.026}\downarrow$ &
							$0.745\pm0.012\downarrow$ &
							$0.632\pm0.024\downarrow$ &
							$0.529\pm0.027\downarrow$ \\
							
							RSSAR & 
							$0.217\pm0.006\downarrow$ &
							$0.266\pm0.004\downarrow$ &
							$0.245\pm0.004\downarrow$ &
							$0.074\pm0.003\downarrow$ \\
							
							RSSR &
							$0.366\pm0.018\downarrow$ &
							$0.476\pm0.010\downarrow$ &
							$0.397\pm0.015\downarrow$ &
							$0.243\pm0.010\downarrow$ \\
							
							RSSA &
							$0.221\pm0.009\downarrow$ &
							$0.224\pm0.004\downarrow$ &
							$0.254\pm0.008\downarrow$ &
							$0.072\pm0.003\downarrow$ \\
							
							SC &
							$0.583\pm0.037\downarrow$ &
							$0.743\pm0.014\downarrow$ &
							$0.620\pm0.029\downarrow$ &
							$0.512\pm0.029\downarrow$\\
							
							SSC & 
							$0.547\pm0.037\downarrow$ &
							$0.720\pm0.019\downarrow$ &
							$0.583\pm0.027\downarrow$ &
							$0.477\pm0.034\downarrow$ \\
							
							\textcolor{black}{CGL} & 
							$0.589\pm0.000\downarrow$ &
							$0.024\pm0.000\downarrow$ &
							$0.589\pm0.000\downarrow$ &
							$0.031\pm0.000\downarrow$ \\
							
							SymNMF & 
							$0.582\pm0.023\downarrow$ &
							$\mathbf{0.774\pm0.010}\downarrow$ &
							$\mathbf{0.659\pm0.017}\downarrow$ &
							$\underline{0.536\pm0.017}\downarrow$ \\
							
%							\cdashline{1-5}
							\hline
							\textbf{Proposed} & 
							$\mathbf{0.633\pm0.024}$ &
							$0.781\pm0.005$\cellcolor[gray]{0.8} &
							$0.685\pm0.014$\cellcolor[gray]{0.8} &
							$0.571\pm0.013$\cellcolor[gray]{0.8} \\
							
%							\cdashline{1-5}
%							Kernel (Gau) & 
%							$0.638\pm0.023$ &
%							$0.783\pm0.009$ &
%							$0.692\pm0.016$ &
%							$0.575\pm0.026$ \\
							
							\hline\hline
						\end{tabular}}
					\end{center}
%					\begin{tablenotes}
%						\small
%						\item $\blacktriangleright$
%						Since 
%						NMF and and GNMF can only be applied to  nonnegative matrix, and 
%						ISOLET contains several negative features,  NMF and GNMF cannot be applied to  ISOLET.
%					\end{tablenotes}
				\end{table}
\begin{table}[!t]
	\renewcommand{\arraystretch}{1.3}
	%	\begin{center}
	\caption{Clustering Performance on LIBRAS} \label{tab:libras}
	\scalebox{0.70}{
		\begin{tabular}{ccccc}
			\hline
			\hline
			Methods & ACC & NMI  &PUR & ARI\\
			\hline\hline	
			CAN & $\underline{0.483}\downarrow$ & $0.654\diamondsuit\cellcolor[gray]{0.8}$ & $\mathbf{0.525}\downarrow$ & $0.398\bullet\cellcolor[gray]{0.8}$\\
			
			GLPCA & $0.442\pm0.019\downarrow$ & $0.572\pm0.017\downarrow$ & $0.469\pm0.024\downarrow$ & $0.303\pm0.019\downarrow$\\
			
			PCA &
			$0.431\pm0.022\downarrow$&
			$0.506\pm0.018\downarrow$&
			$0.459\pm0.019\downarrow$&
			$0.237\pm0.020\downarrow$\\
			
			GMF & $0.445\pm0.030\downarrow$ & $0.577\pm0.020\downarrow$ & $0.479\pm0.029\downarrow$ & $0.303\pm0.022\downarrow$\\
			
			GNMF & $0.463\pm0.028\downarrow$ & $0.564\pm0.030\downarrow$ & $0.490\pm0.023\downarrow$ & $0.300\pm0.034\downarrow$\\
			
			GRPCA & $0.464\pm0.033\downarrow$ & $0.593\pm0.027\downarrow$ & $0.494\pm0.027\downarrow$ & $0.236\pm0.031\downarrow$\\
			
			K-means & $0.435\pm0.023\downarrow$ & $0.562\pm0.023\downarrow$ & $0.464\pm0.025\downarrow$ & $0.293\pm0.025\downarrow$\\
			
			L2-Graph & $\mathbf{0.497\pm0.028}\bullet$ & $\underline{0.622\pm0.026}\downarrow$ & $\underline{0.524\pm0.021}\bullet$ & $\underline{0.361\pm0.018}\downarrow$\\
			
			L2-SymNMF & $0.521\pm0.022\diamondsuit$ \cellcolor[gray]{0.8}& $\mathbf{0.639\pm0.018}\bullet$ & $0.544\pm0.016\diamondsuit$ \cellcolor[gray]{0.8}& $\mathbf{0.383\pm0.024}\downarrow$\\
			
			LRR & $0.440\pm0.023\downarrow$ & $0.569\pm0.015\downarrow$ & $0.469\pm0.021\downarrow$ & $0.299\pm0.018\downarrow$\\
			
			NMF & $0.460\pm0.032\downarrow$ & $0.558\pm0.024\downarrow$ & $0.493\pm0.025\downarrow$ & $0.296\pm0.025\downarrow$\\
			
			RPCA & $0.444\pm0.027\downarrow$ & $0.569\pm0.020\downarrow$ & $0.470\pm0.022\downarrow$ & $0.300\pm0.022\downarrow$\\
			
			RSSAR & $\underline{0.492\pm0.024}\bullet$ & $0.604\pm0.020\downarrow$ & $0.519\pm0.024\bullet$ & $0.340\pm0.022\downarrow$\\
			
			RSSR & $0.476\pm0.027\downarrow$ & $0.574\pm0.016\downarrow$ & $0.506\pm0.023\downarrow$ & $0.314\pm0.022\downarrow$\\
			
			RSSA & $0.472\pm0.024\downarrow$ & $0.594\pm0.015\downarrow$ & $0.496\pm0.016\downarrow$ & $0.331\pm0.020\downarrow$\\
			
			SC & $0.466\pm0.030\downarrow$ & $0.615\pm0.023\downarrow$ & $0.500\pm0.025\downarrow$ & $0.347\pm0.025\downarrow$\\
			
			SSC & 
			$0.457\pm0.032\downarrow$&
			$0.600\pm0.022\downarrow$&
			$0.495\pm0.023\downarrow$&
			$0.336\pm0.033\downarrow$
			\\
			
			\textcolor{black}{CGL} & 
			$0.470\pm0.027\downarrow$ & $0.588\pm0.017\downarrow$ & $0.500\pm0.021\downarrow$ & $0.333\pm0.022\downarrow$\\
			
			SymNMF & $0.482\pm0.023\downarrow$ & $\underline{0.619\pm0.020}\downarrow$ & $\underline{0.520\pm0.019}\downarrow$ & $\underline{0.356\pm0.020}\downarrow$\\
			%				\cdashline{1-5}
			\hline
			\textbf{Proposed} & $\mathbf{0.501\pm0.023}$ & $\mathbf{0.643\pm0.012}$ & $\mathbf{0.532\pm0.011}$ & $\mathbf{0.397\pm0.016}$\\
			
			%				Improvement Ratio & * & * & * & *\\
			%				\cdashline{1-5}
			%				Kernel (Gau) & 
			%				$0.502\pm0.027$ &
			%				$0.637\pm0.012$ &
			%				$0.534\pm0.021$ &
			%				$0.388\pm0.023$ 
			%				\\
			\hline\hline
		\end{tabular}}
		%	\end{center}
	\end{table}
	
\begin{table}[!t]
	\renewcommand{\arraystretch}{1.3}
	\begin{center}
		\caption{Clustering Performance on SOYBEAN} \label{tab:soybean}
		\scalebox{0.7}{
			\begin{tabular}{ccccc}
				\hline
				\hline
				Methods & ACC & NMI  &PUR & ARI\\
				\hline\hline	
				CAN &
				$0.540\downarrow$ &
				$0.697\downarrow$ &
				$0.638\downarrow$ &
				$0.341\downarrow$ \\
				
				GLPCA& 
				$0.573\pm0.029\downarrow$ &
				$0.700\pm0.021\downarrow$ &
				$0.675\pm0.030\downarrow$ &
				$\underline{0.445\pm0.044}\bullet$ \\
				
				PCA & 
				$0.591\pm0.035\downarrow$ &
				$0.688\pm0.022\downarrow$ &
				$0.690\pm0.029\downarrow$ &
				$\mathbf{0.446\pm0.048}\bullet$ \\
				
				GMF & $0.551\pm0.049\downarrow$ &
				$0.689\pm0.033\downarrow$ &
				$0.667\pm0.040\downarrow$ &
				$0.420\pm0.057\bullet$\\
				
				GNMF & $\mathbf{0.605\pm0.039}\downarrow$ & $\underline{0.715\pm0.021}\downarrow$ & $\underline{0.704\pm0.026}\bullet$ & $\cellcolor[gray]{0.8}0.475\pm0.045\diamondsuit$\\
				
				GRPCA &$0.576\pm0.043\downarrow$ & $0.710\pm0.027\downarrow$ & $0.676\pm0.027\downarrow$
				& $\underline{0.445\pm0.050}\bullet$\\
				
				K-means & $0.560\pm0.033\downarrow$ & $0.699\pm0.022\downarrow$ & $0.659\pm0.037\downarrow$ & $0.424\pm0.037\bullet$\\
				
				L2-Graph & $0.588\pm0.026\downarrow$ &
				$\mathbf{0.730\pm0.018}\downarrow$ &  $0.692\pm0.026\bullet$ & $0.424\pm0.025\bullet$\\
				
				L2-SymNMF &$\underline{0.593\pm0.042}\downarrow$& $\mathbf{0.728\pm0.027}\downarrow$ & $0.697\pm0.031\bullet$ & $0.422\pm0.045\bullet$\\	
				
				LRR & $0.566\pm0.039\downarrow$ & $0.698\pm0.027\downarrow$ & $0.666\pm0.032\downarrow$ & $0.439\pm0.057\bullet$\\
				
				NMF & $\mathbf{0.598\pm0.043}\downarrow$ & $\underline{0.718\pm0.021}\downarrow$& $\underline{0.709\pm0.031}\bullet$ & $\mathbf{0.461\pm0.031}\diamondsuit$\\
				
				RPCA & $0.570\pm0.030\downarrow$ & $0.700\pm0.015\downarrow$ & $0.668\pm0.024$&
				$0.435\pm0.041\bullet$\\
				
				RSSAR & $0.573\pm0.041\downarrow$ & $\underline{0.715\pm0.028}\downarrow$ & $	0.693\pm0.030\downarrow$ & $0.442\pm0.037\bullet$\\
				
				RSSR & $0.578\pm0.053\downarrow$& $0.703\pm0.014\downarrow$ &  $0.696\pm0.037\bullet$ & $0.442\pm0.034\bullet$\\
				
				RSSA & $0.558\pm0.043\downarrow$ & $0.700\pm0.024\downarrow$ & $0.671\pm0.038\downarrow$ & $0.414\pm0.034\bullet$\\
				
				SC & $0.495\pm0.038\downarrow$ & $0.641 \pm0.026\downarrow$
				& $0.604\pm0.040\downarrow$& $0.355\pm0.040\downarrow$\\
				
				SSC & $0.574\pm0.031\downarrow$ & $0.704\pm0.020\downarrow$ & $0.672\pm0.031\downarrow$ & $0.441\pm	0.038\bullet$\\
				
				\textcolor{black}{CGL} & $\underline{0.593\pm0.022}\downarrow$ & $0.712\pm0.019\downarrow$ & $0.680\pm0.018\downarrow$ & $0.441\pm0.016\bullet$\\

				SymNMF & $0.513\pm0.042\downarrow$ & $0.664\pm0.020\downarrow$ & $0.0620\pm0.029\downarrow$ & $0.326\pm0.029\downarrow$\\
				%				\cdashline{1-5}
				\hline
				\textbf{Proposed} & \cellcolor[gray]{0.8}$0.638\pm0.025$ &\cellcolor[gray]{0.8} $0.744\pm0.017$ &\cellcolor[gray]{0.8} $0.713\pm0.025$ & $0.428\pm0.050$\\
				%				\textbf{Proposed} & \cellcolor[gray]{0.5}$\gray{0.638}\pm\gray{0.025}$ & $\gray{0.744}\pm\gray{0.017}$ & $\gray{0.713}\pm\gray{0.025}$ & $0.428\pm0.050$\\
				%				\cdashline{1-5}
				%				\hline
				%				Kernel (Gau) & 
				%				$0.628\pm0.021$ &
				%				$0.740\pm	0.017$ &
				%				$0.716\pm0.021$ &
				%				$0.452\pm0.032$ \\
				
				\hline\hline
			\end{tabular}}
		\end{center}
		
	\end{table}

%\begin{table}[!t]
%	\renewcommand{\arraystretch}{1.}
%	\begin{center}
%		\caption{Rank Counting for the Proposed Method } \label{tab:res-summarization}
%		\scalebox{1}{
%			\begin{tabular}{ccccc}
%				\hline
%%				\hline
%				 & Rank 1 & Rank 2$\sim$3  & Rank 4$\sim$5  & Rank 6$\sim$20 \\
%				\hline
%%				\cdashline{1-5}
%				Quantity & $29/40$ & $8/40$ & $1/40$ & $2/40$\\
%				\cdashline{1-5}
%				
%				Ratio & $72.5\%$ & $20.0\%$ & $2.5\%$ & $5\%$\\
%			
%				\hline
%		\end{tabular}}
%	\end{center}
%\end{table}

\begin{table}[!t]
	\renewcommand{\arraystretch}{1.}
	\begin{center}
		\caption{\textcolor{black}{Rank Counting for the Proposed Method} } \label{tab:res-summarization}
		\scalebox{1}{
			\begin{tabular}{ccccc}
				\hline
				%				\hline
				& Rank 1 & Rank 2$\sim$3  & Rank 4$\sim$5  & Rank 6$\sim$20 \\
				\hline
				%				\cdashline{1-5}
				Quantity & $23/40$ & $13/40$ & $2/40$ & $2/40$\\
				\cdashline{1-5}
				
				Ratio & $57.5\%$ & $32.5\%$ & $5\%$ & $5\%$\\
				
				\hline
			\end{tabular}}
		\end{center}
	\end{table}

\begin{table}[!t]
	\renewcommand{\arraystretch}{1.5}
	\begin{center}
		\caption{\textcolor{black}{Statistics of the Comparison Between the Proposed Method and the Compared Methods in Terms of the Wilcoxon Rank Sum Test }} \label{tab:res-sig}
		\scalebox{0.9}{
			\begin{tabular}{cccc}
				\hline
%				\hline
			
%				\cdashline{1-4}
				
				 & Significantly better & No significant difference  & Significantly worse  \\
				\hline
				%				\cdashline{1-5}
				Quantity & $662/744$ & $60/744$& $22/744$ \\
				\cdashline{1-4}
				Ratio & $89.0\%$ & $8.1\%$&  $2.9\%$ \\
				\hline
		\end{tabular}}
	\end{center}
\end{table}

\begin{figure}	
	\centering
	\includegraphics[width=0.95\linewidth]{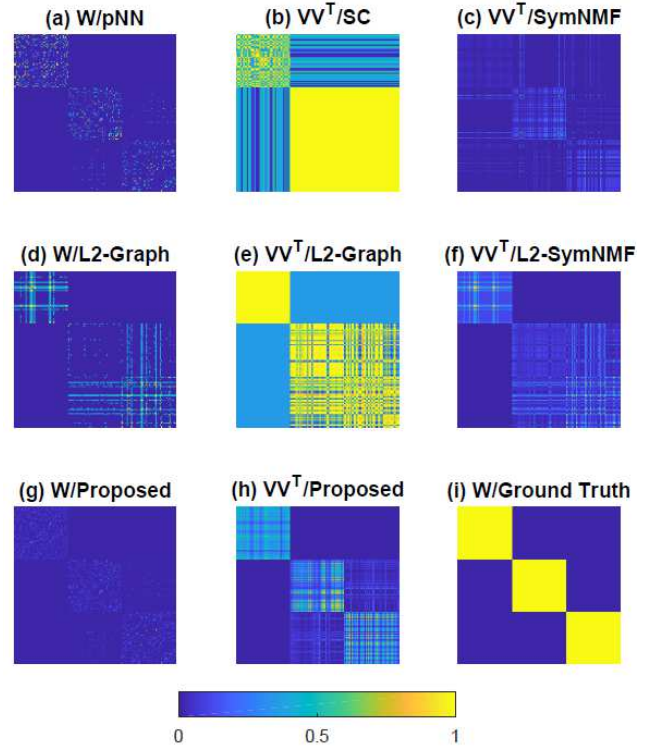}
%	\centerline{\epsfig{figure=visual_final-c.eps,width=9cm}}
%	\centerline{figure=visual.png,width=9cm}
	\caption{\textcolor{black}{Visual comparison of the affinity matrices constructed by different methods, and the corresponding $\mathbf{V}^\mathsf{T}\mathbf{V}$ by different methods on the IRIS dataset. (a), (d), (g) and (i) are the affinity matrices built by the $p$NN graph, L2-Graph, our method, and the ground truth cluster indicator, respectively. (b), (c), (e), (f) and (h) denote the inner product of the low-dimensional  embeddings by SC, SymNMF, L2-Graph, L2-SymNMF, and our method, respectively. All the matrices are normalized to $[0,1]$, and share the same color map.}}
	\label{fig:graph}
\end{figure}

%\subsection{Parameter Sensitiveness}
\begin{figure*}[!t]
	\begin{minipage}[b]{0.195\linewidth}
		\centering
		\centerline{\epsfig{figure=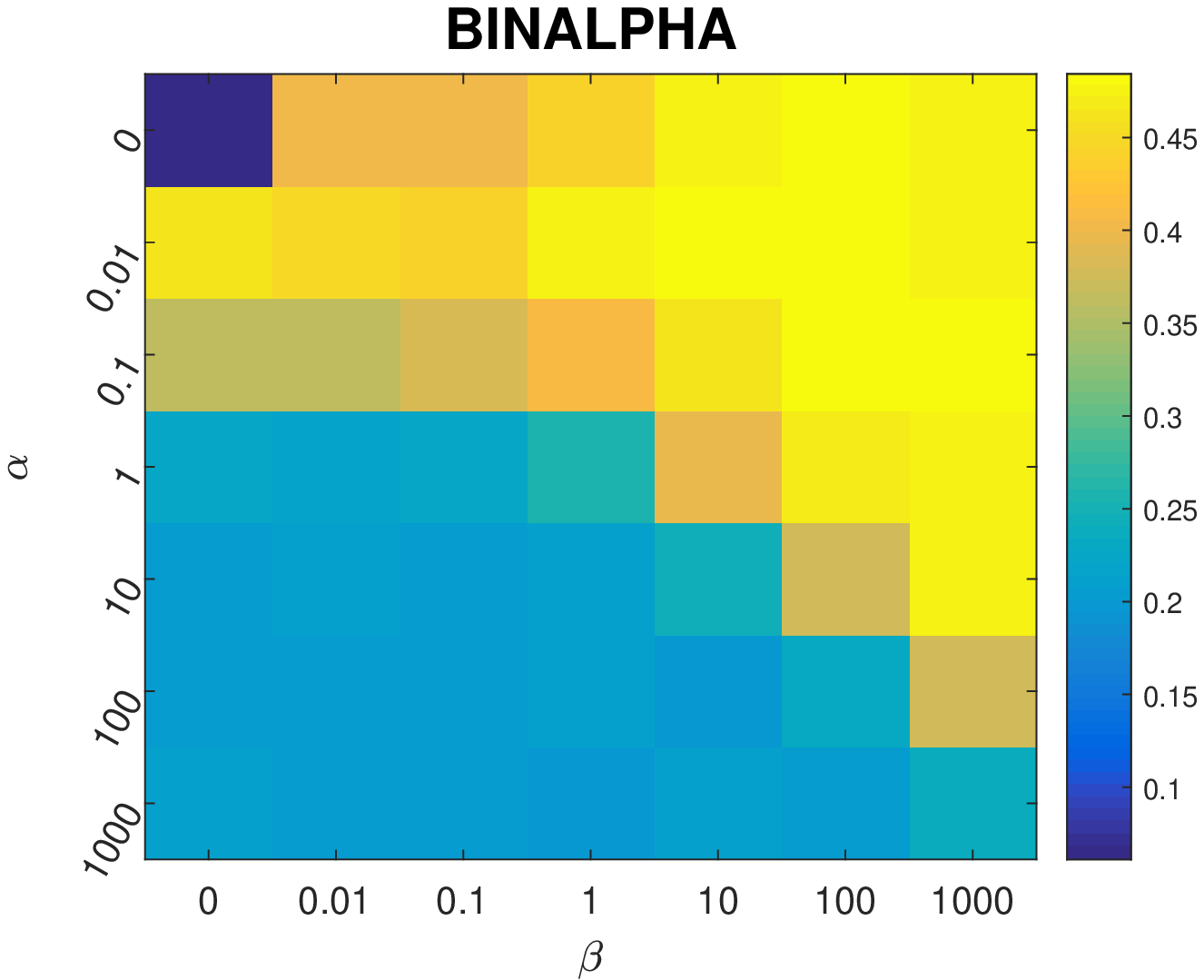,width=4cm}}
		%		\vspace{-0.1cm}
%		\centerline{(a) COIL20}\medskip
	\end{minipage}
	\begin{minipage}[b]{0.195\linewidth}
		\centering
		\centerline{\epsfig{figure=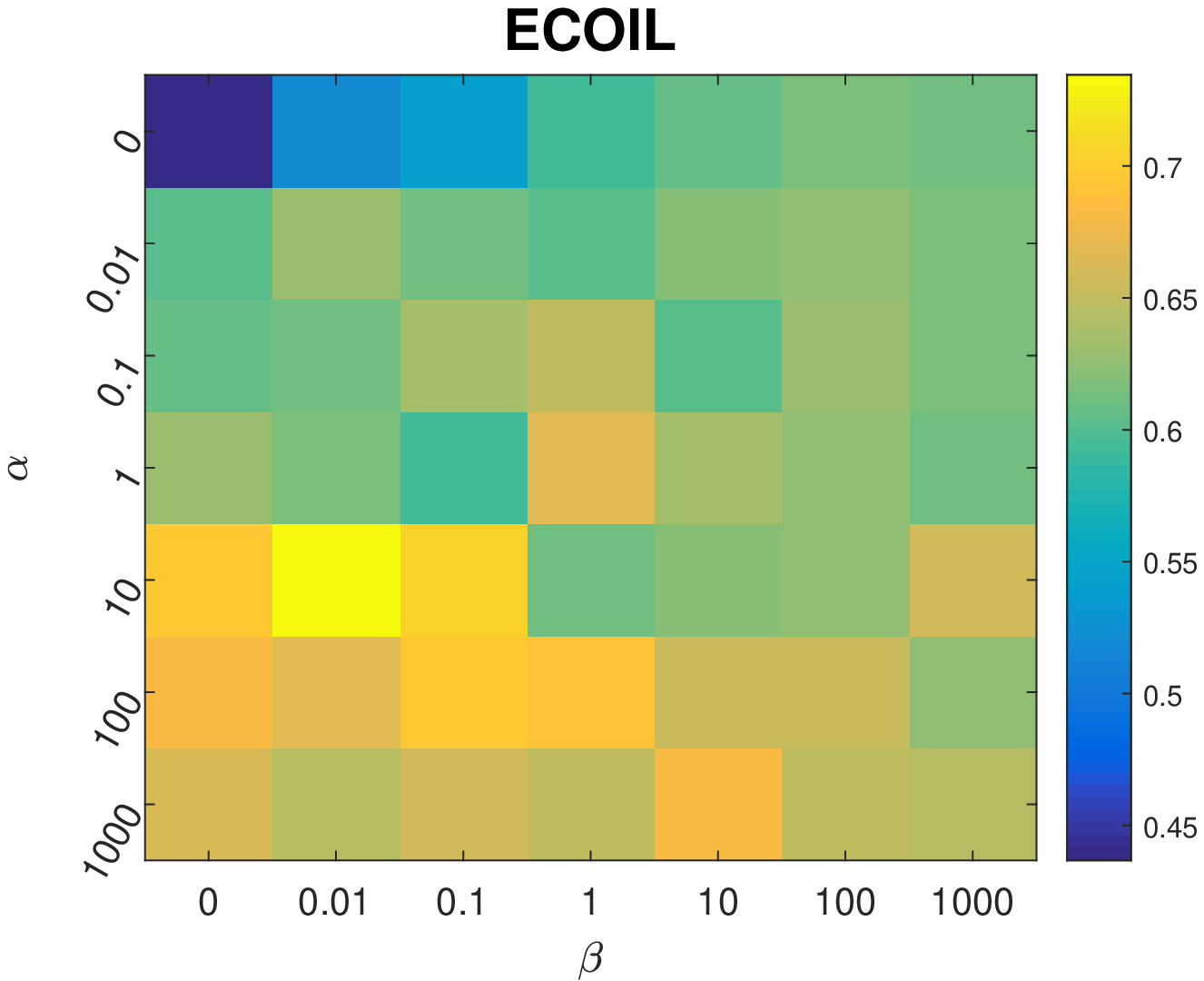,width=4cm}}
		%		\vspace{-0.1cm}
		%		\centerline{(a) COIL20}\medskip
	\end{minipage}
	\begin{minipage}[b]{0.195\linewidth}
		\centering
		\centerline{\epsfig{figure=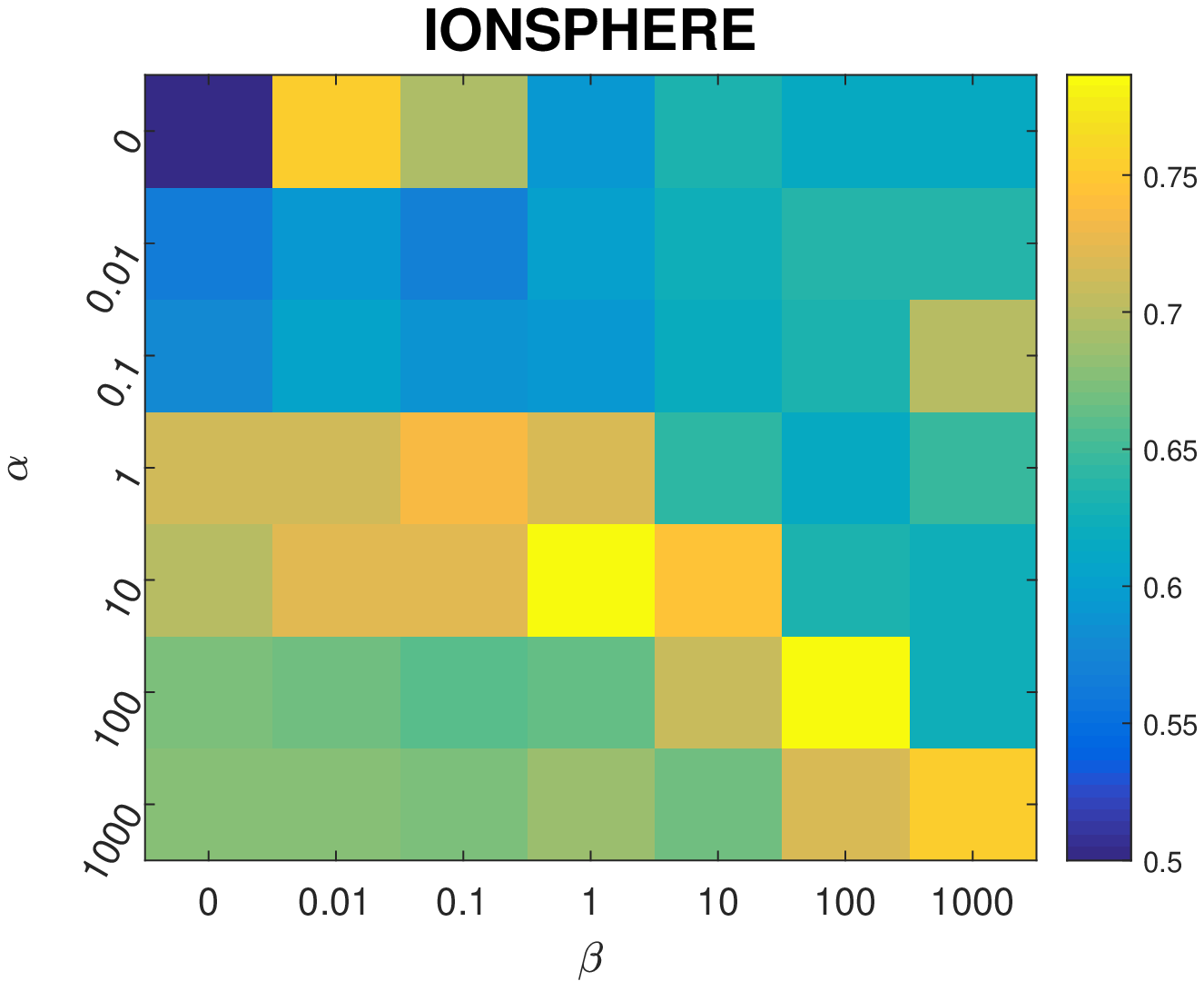,width=4cm}}
		%		\vspace{-0.1cm}
		%		\centerline{(a) COIL20}\medskip
	\end{minipage}
	\begin{minipage}[b]{0.195\linewidth}
		\centering
		\centerline{\epsfig{figure=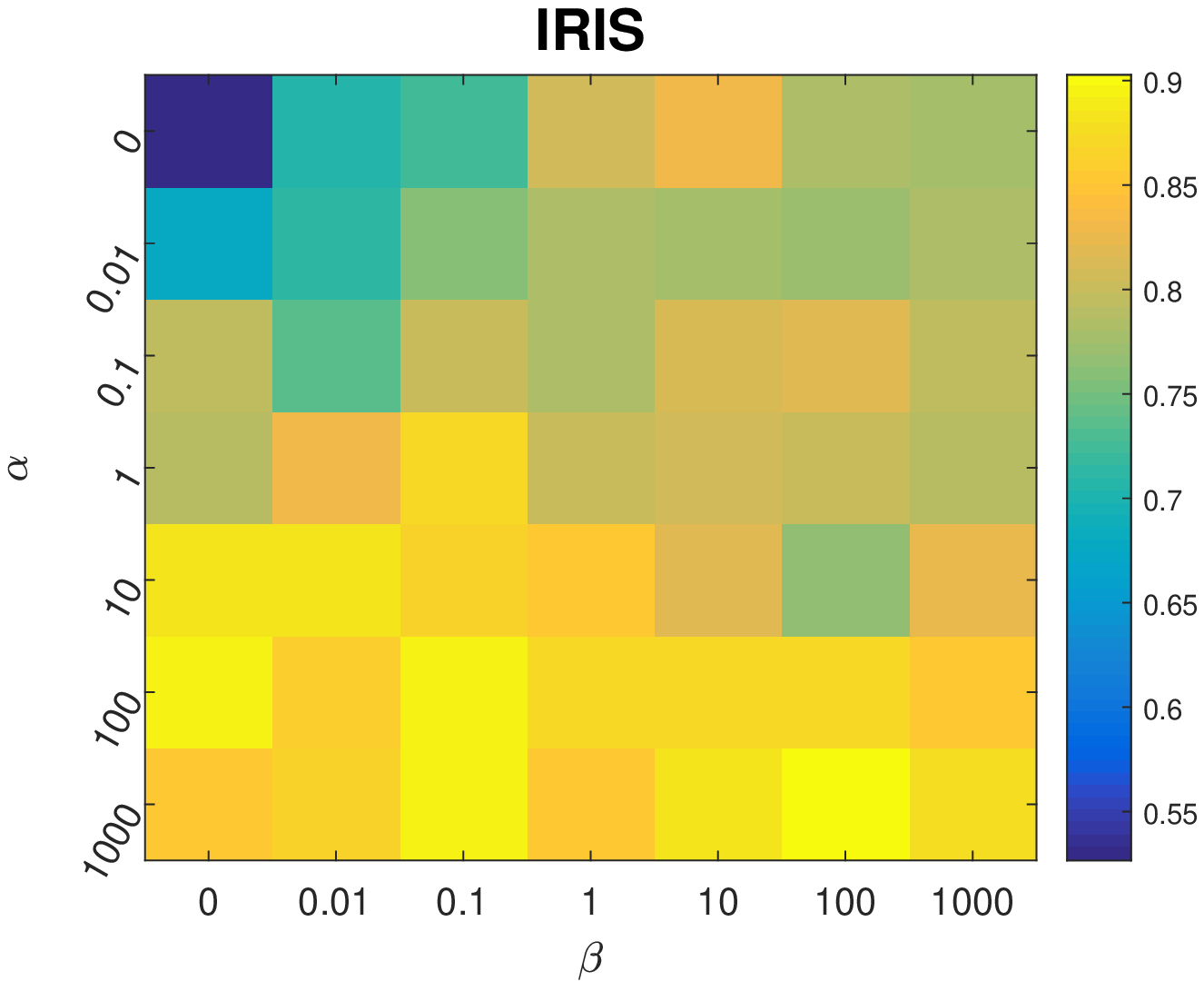,width=4cm}}
		%		\vspace{-0.1cm}
		%		\centerline{(a) COIL20}\medskip
	\end{minipage}
	\begin{minipage}[b]{0.195\linewidth}
			\centering
			\centerline{\epsfig{figure=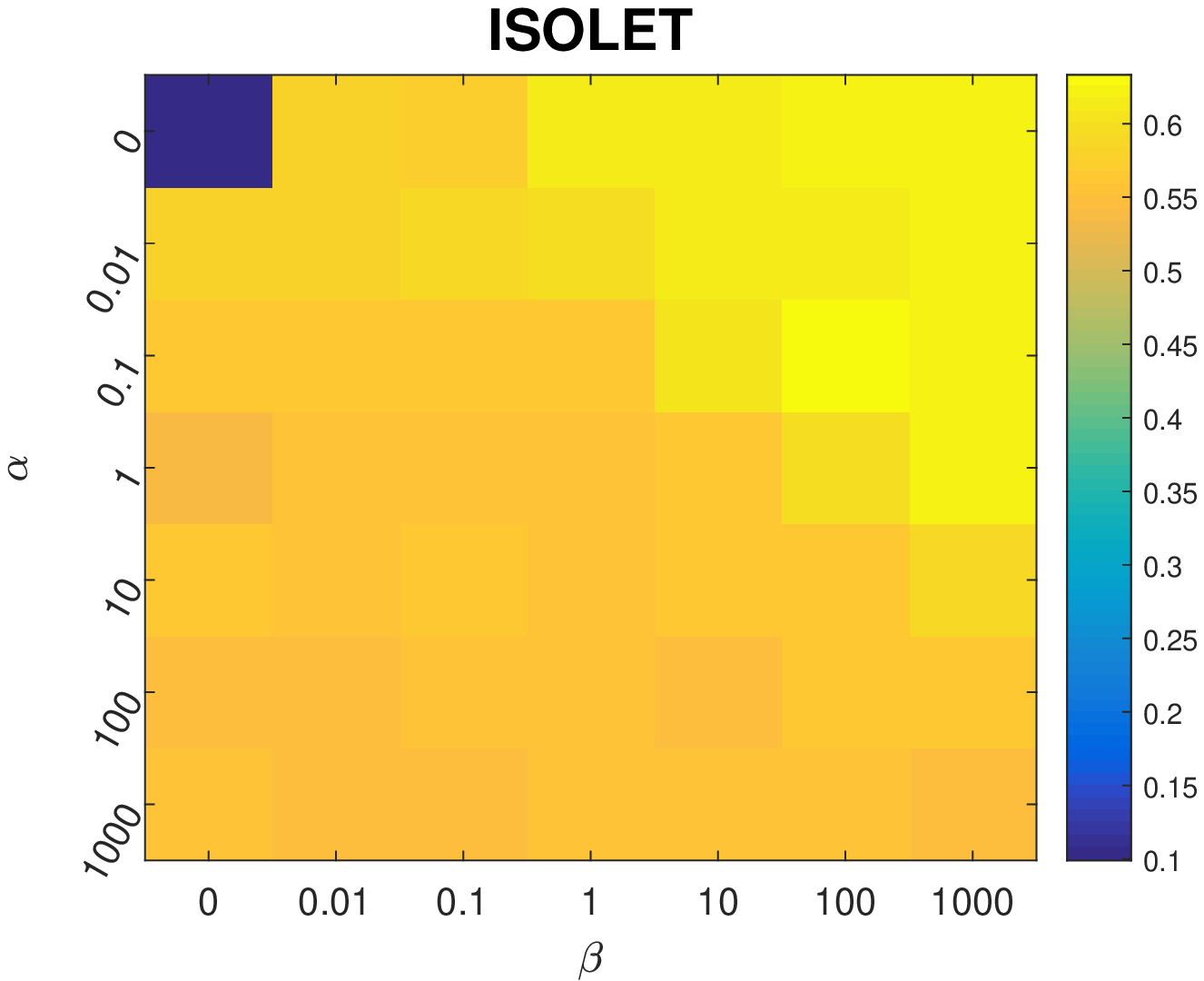,width=4cm}}
			%		\vspace{-0.1cm}
			%		\centerline{(a) COIL20}\medskip
	\end{minipage}
	\begin{minipage}[b]{0.195\linewidth}
		\centering
		\centerline{\epsfig{figure=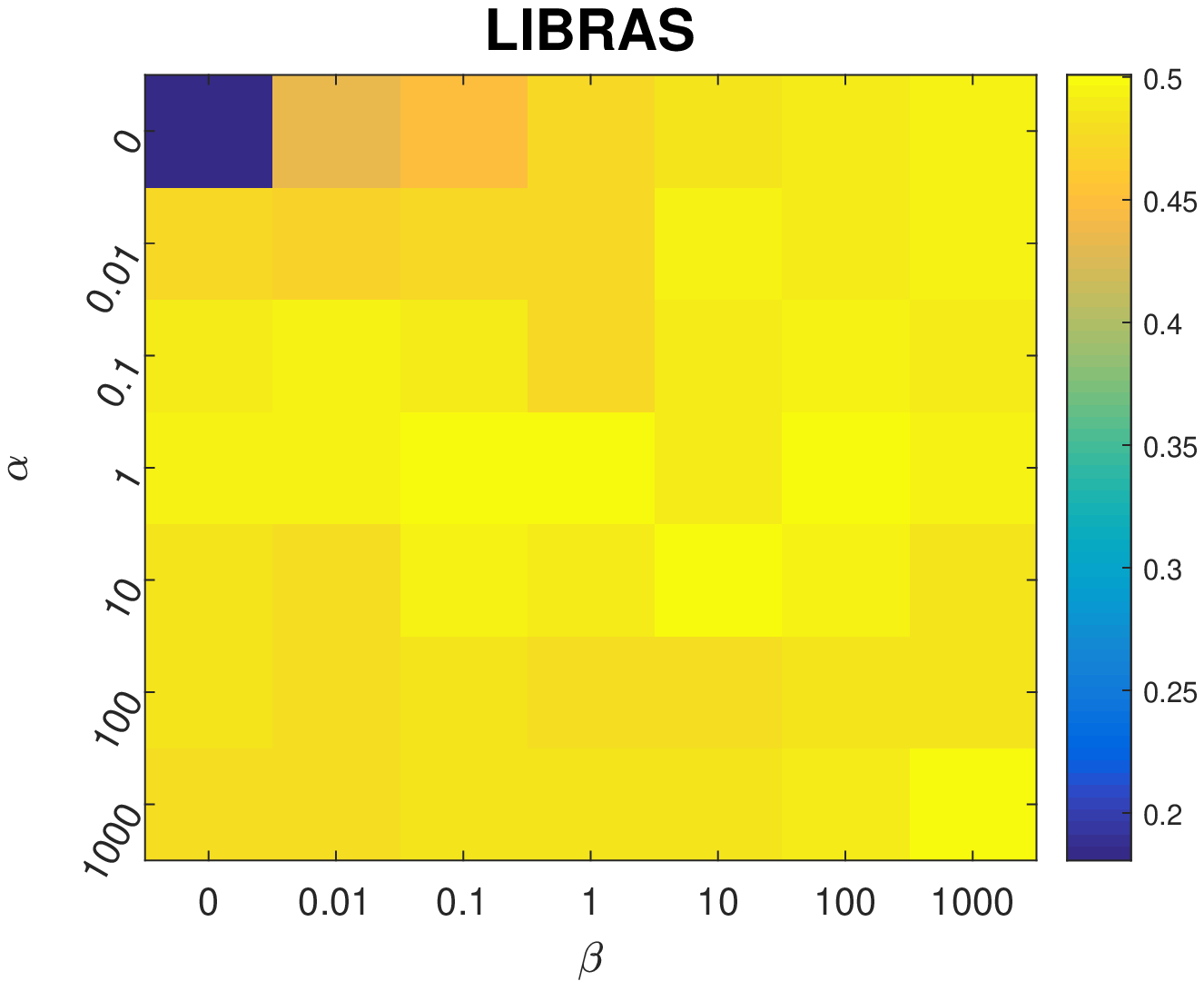,width=4cm}}
		%		\vspace{-0.1cm}
		%		\centerline{(a) COIL20}\medskip
	\end{minipage}
	\begin{minipage}[b]{0.195\linewidth}
		\centering
		\centerline{\epsfig{figure=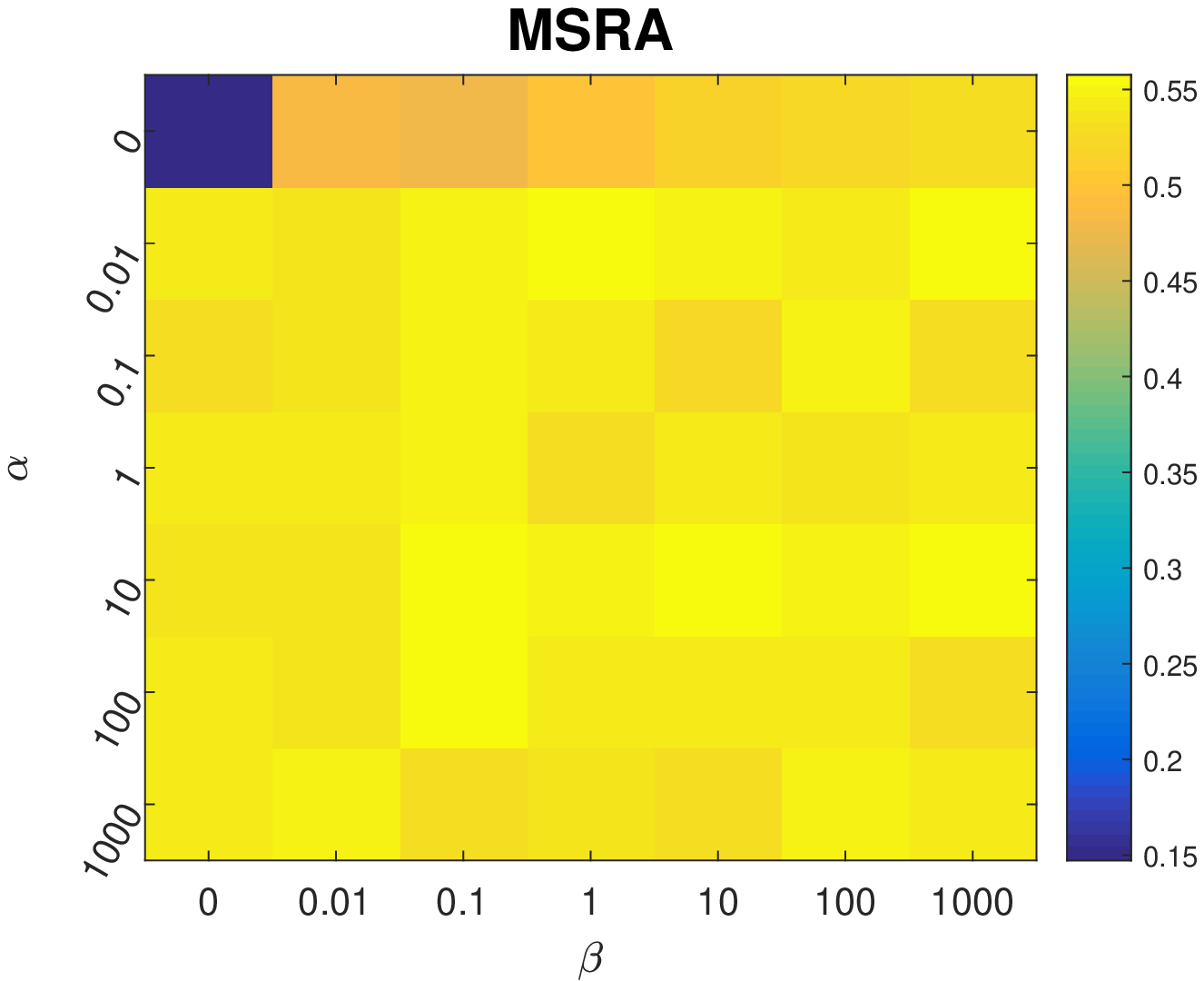,width=4cm}}
		%		\vspace{-0.1cm}
		%		\centerline{(a) COIL20}\medskip
	\end{minipage}
	\begin{minipage}[b]{0.195\linewidth}
			\centering
			\centerline{\epsfig{figure=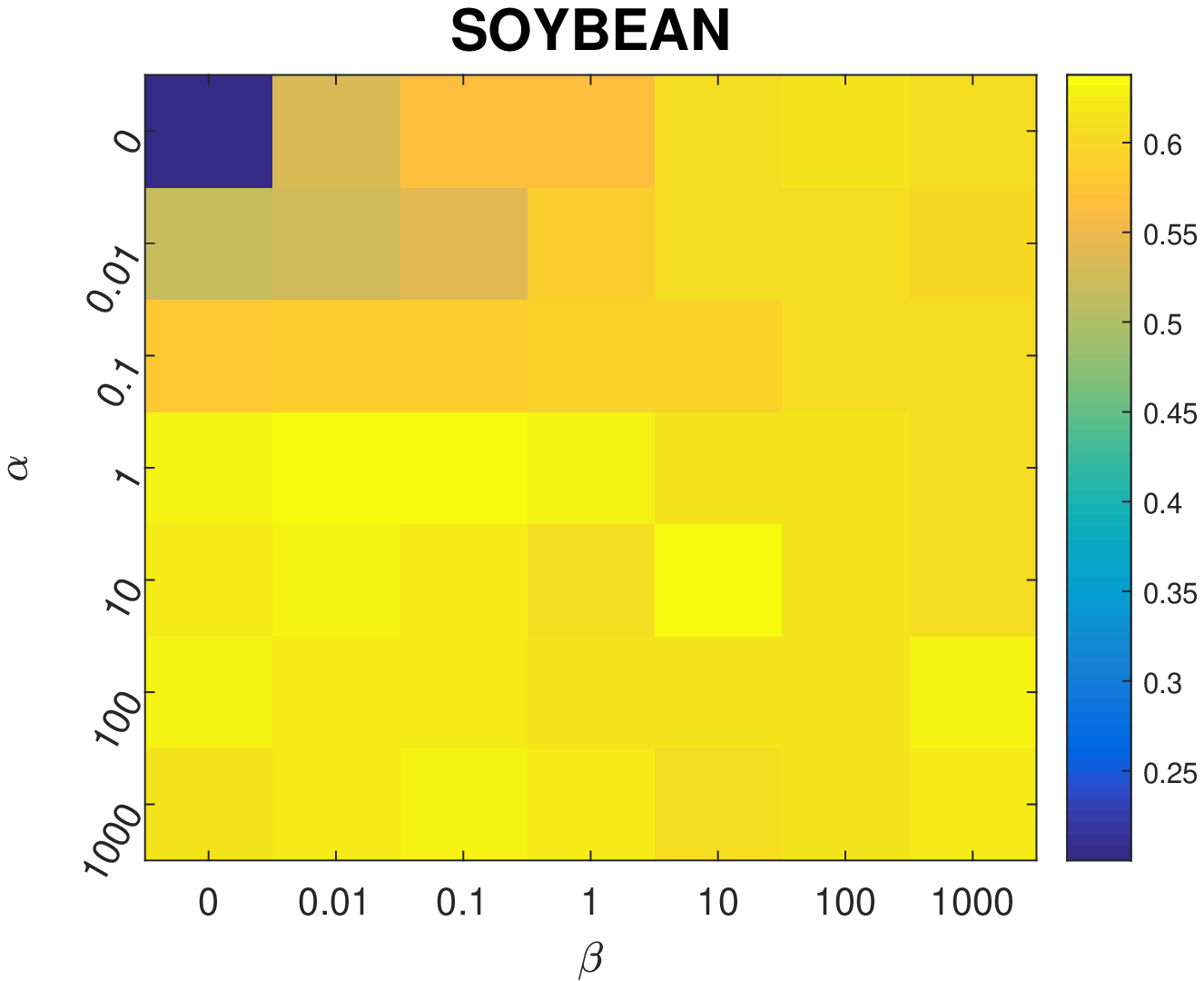,width=4cm}}
			%		\vspace{-0.1cm}
			%		\centerline{(a) COIL20}\medskip
	\end{minipage}
\begin{minipage}[b]{0.195\linewidth}
	\centering
	\centerline{\epsfig{figure=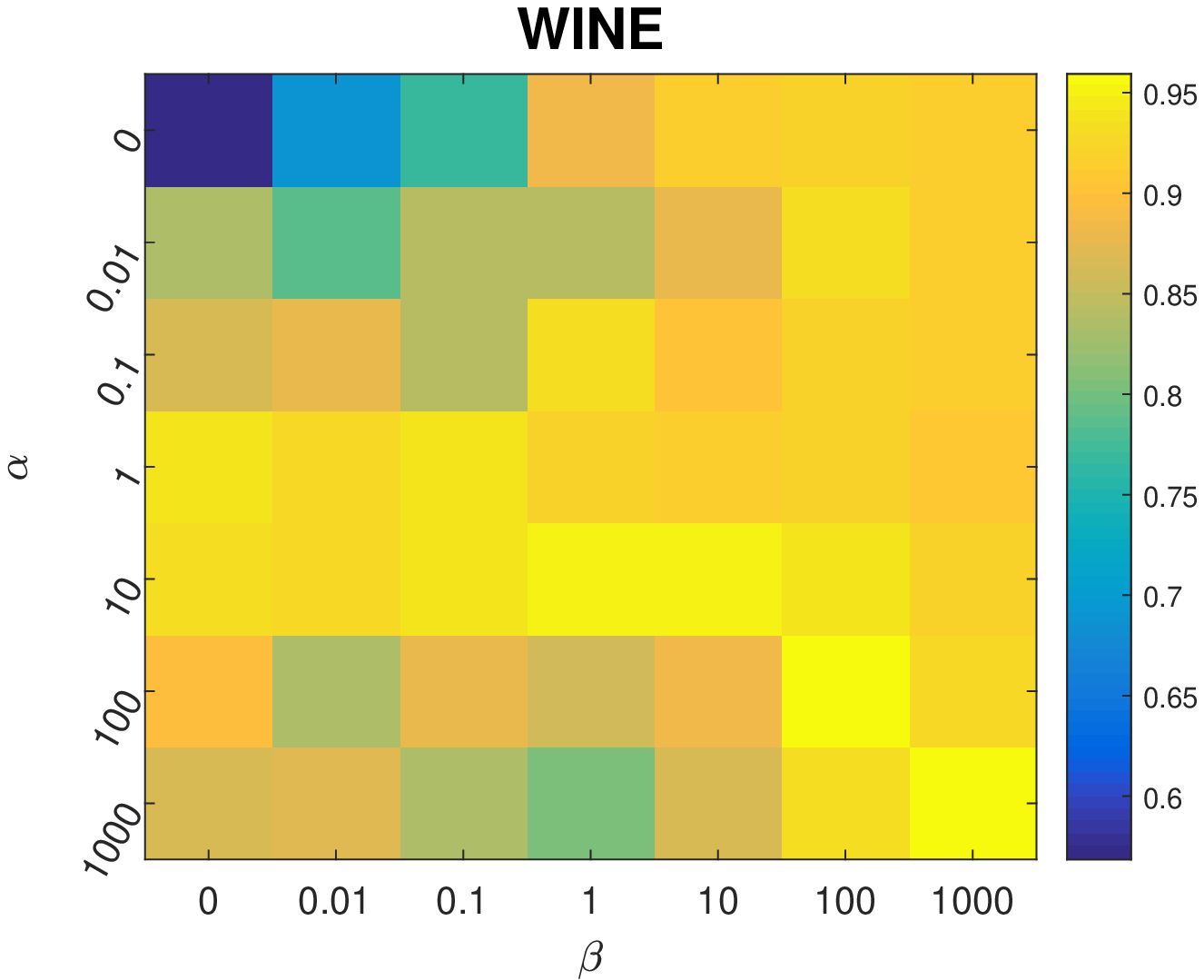,width=4cm}}
	%		\vspace{-0.1cm}
	%		\centerline{(a) COIL20}\medskip
\end{minipage}
\begin{minipage}[b]{0.195\linewidth}
	\centering
	\centerline{\epsfig{figure=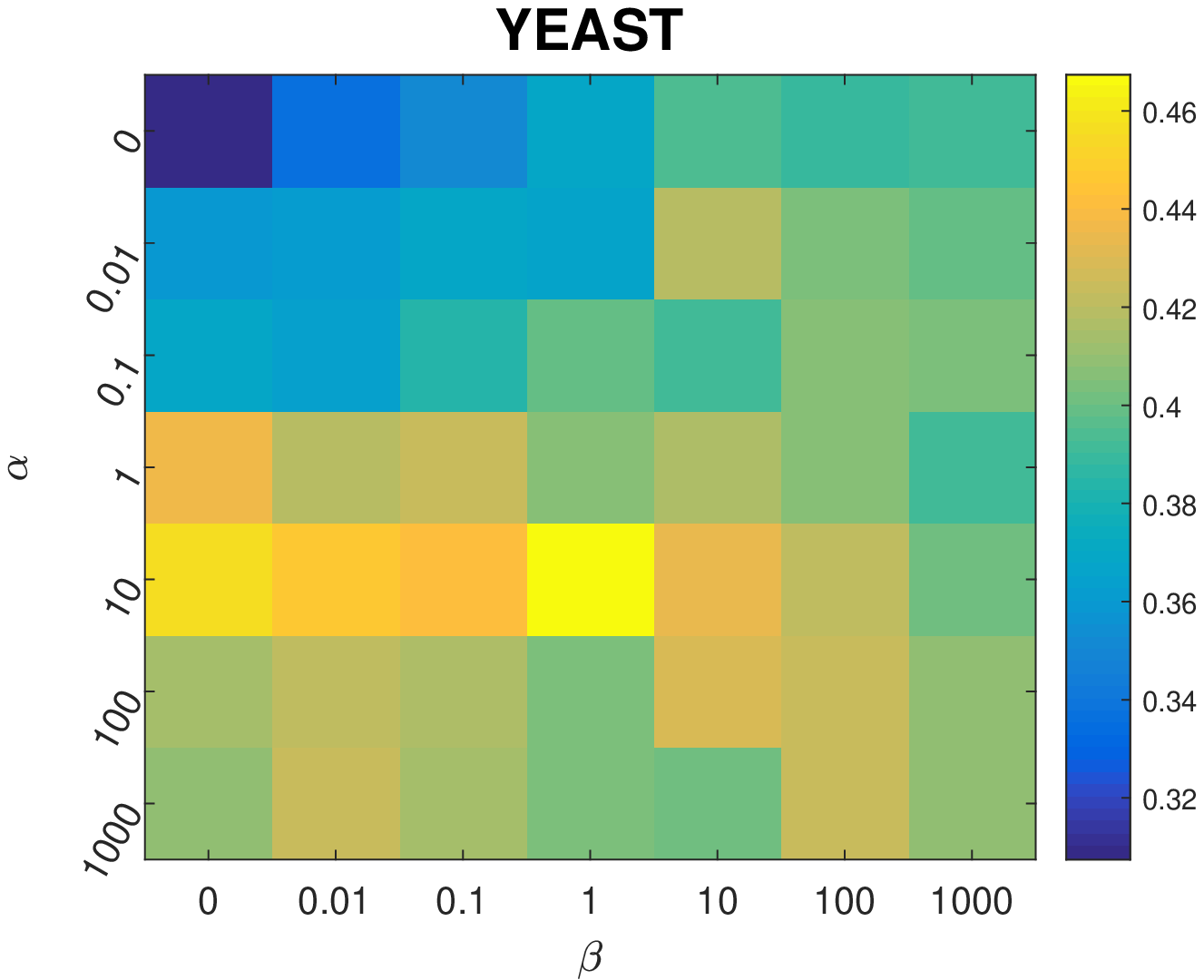,width=4cm}}
	%		\vspace{-0.1cm}
	%		\centerline{(a) COIL20}\medskip
\end{minipage}
	\caption{Clustering ACC of the proposed model versus  $\alpha$ and $\beta$ on 10 datasets.}
	\label{fig:Para}
\end{figure*}

%\begin{table}[!t]
%	\renewcommand{\arraystretch}{1.3}
%	\begin{center}
%		\caption{Classification Accuracy Comparison} \label{tab:complexity}
%		\scalebox{1}{
%			\begin{tabular}{cc}
%				\hline
%				\hline
%				Methods & Complexity \\
%				\hline
%				%				\cdashline{1-5}
%				\textbf{Proposed} & \\
%
%				\hline\hline
%			\end{tabular}}
%		\end{center}
%	\end{table}

\begin{figure*}[!t]
	\begin{minipage}[b]{0.195\linewidth}
		\centering
		\centerline{\epsfig{figure=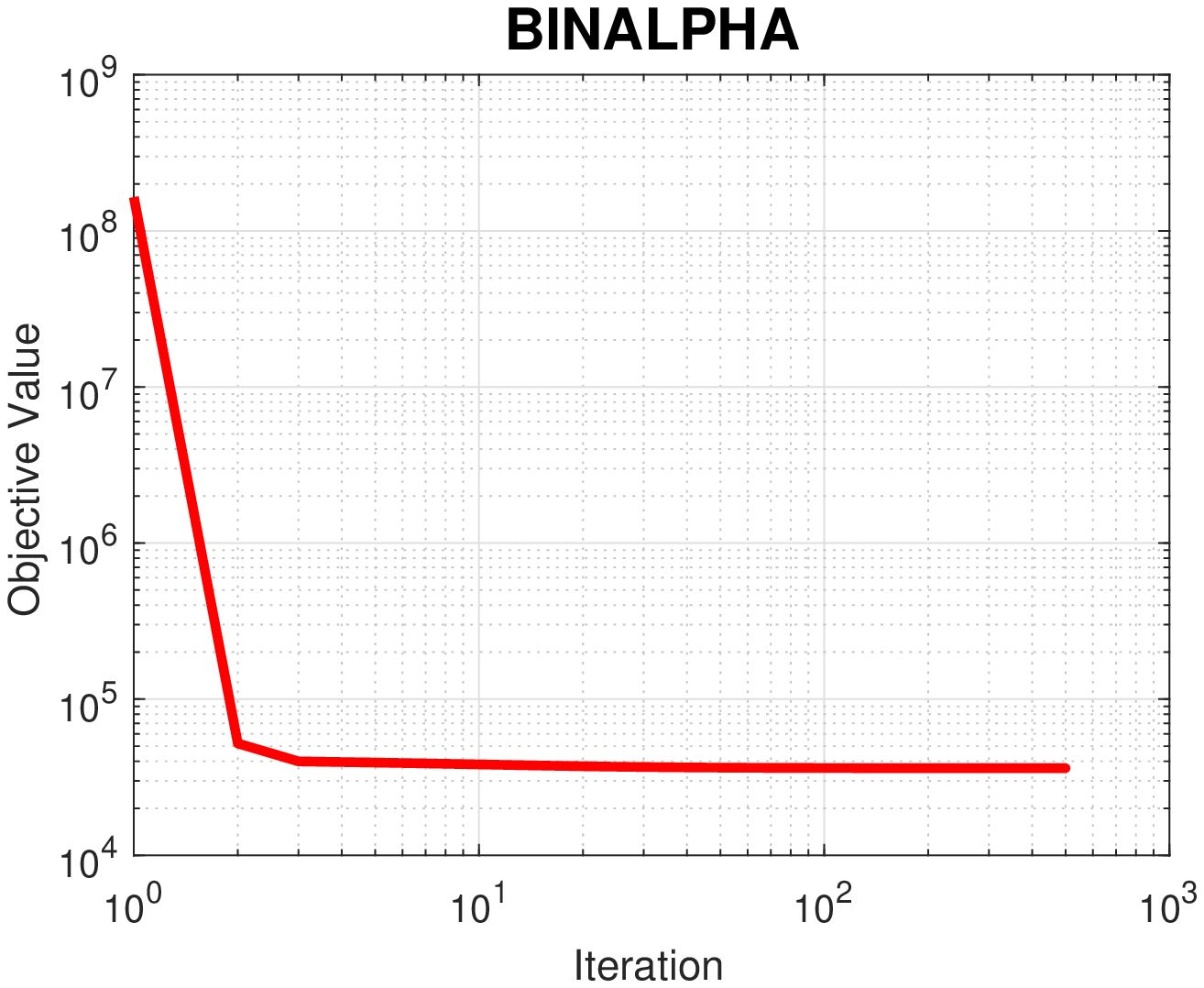,width=3.9cm}}
		%		\vspace{-0.1cm}
		%		\centerline{(a) COIL20}\medskip
	\end{minipage}
	\begin{minipage}[b]{0.195\linewidth}
		\centering
		\centerline{\epsfig{figure=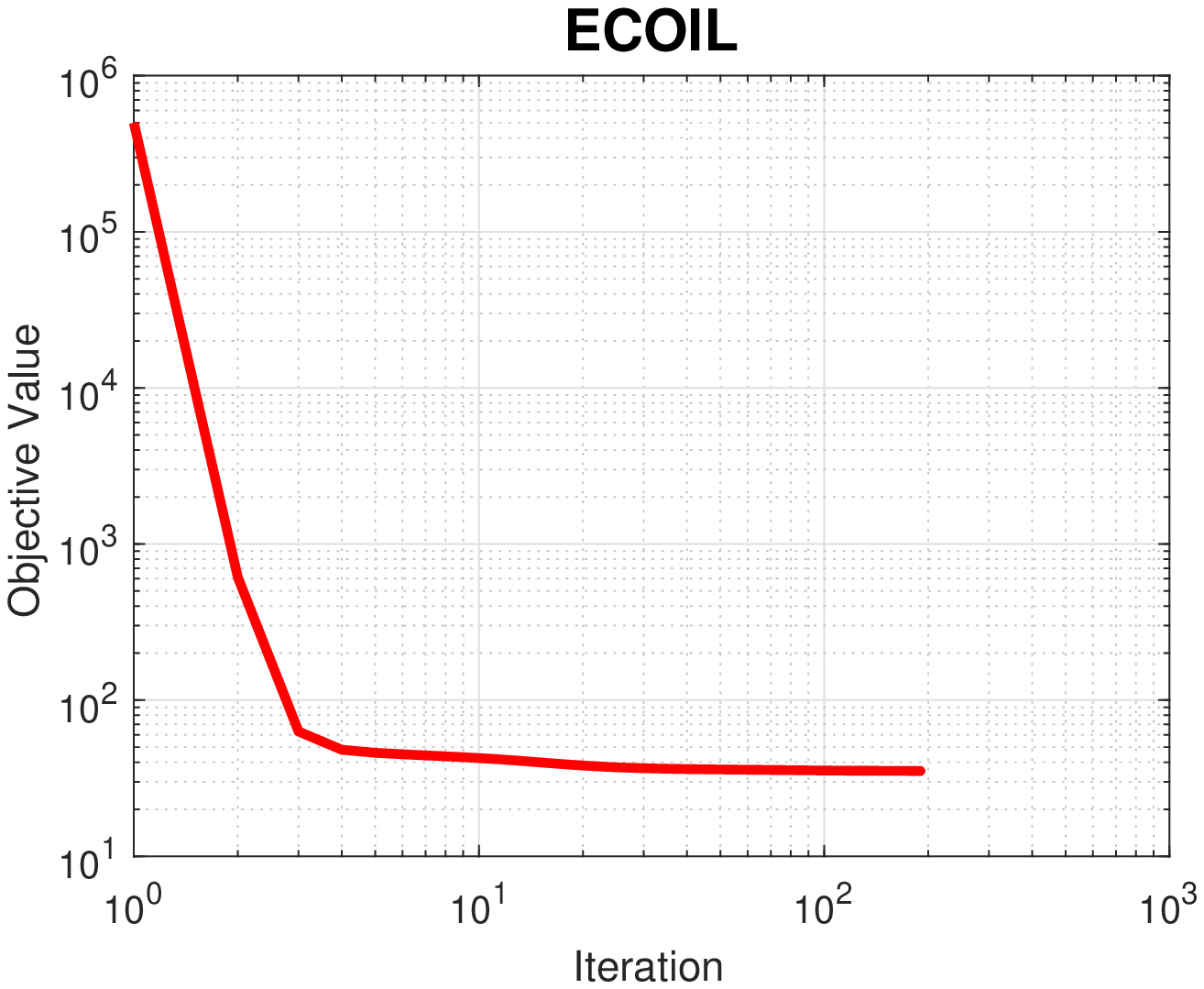,width=3.9cm}}
		%		\vspace{-0.1cm}
		%		\centerline{(a) COIL20}\medskip
	\end{minipage}
	\begin{minipage}[b]{0.195\linewidth}
		\centering
		\centerline{\epsfig{figure=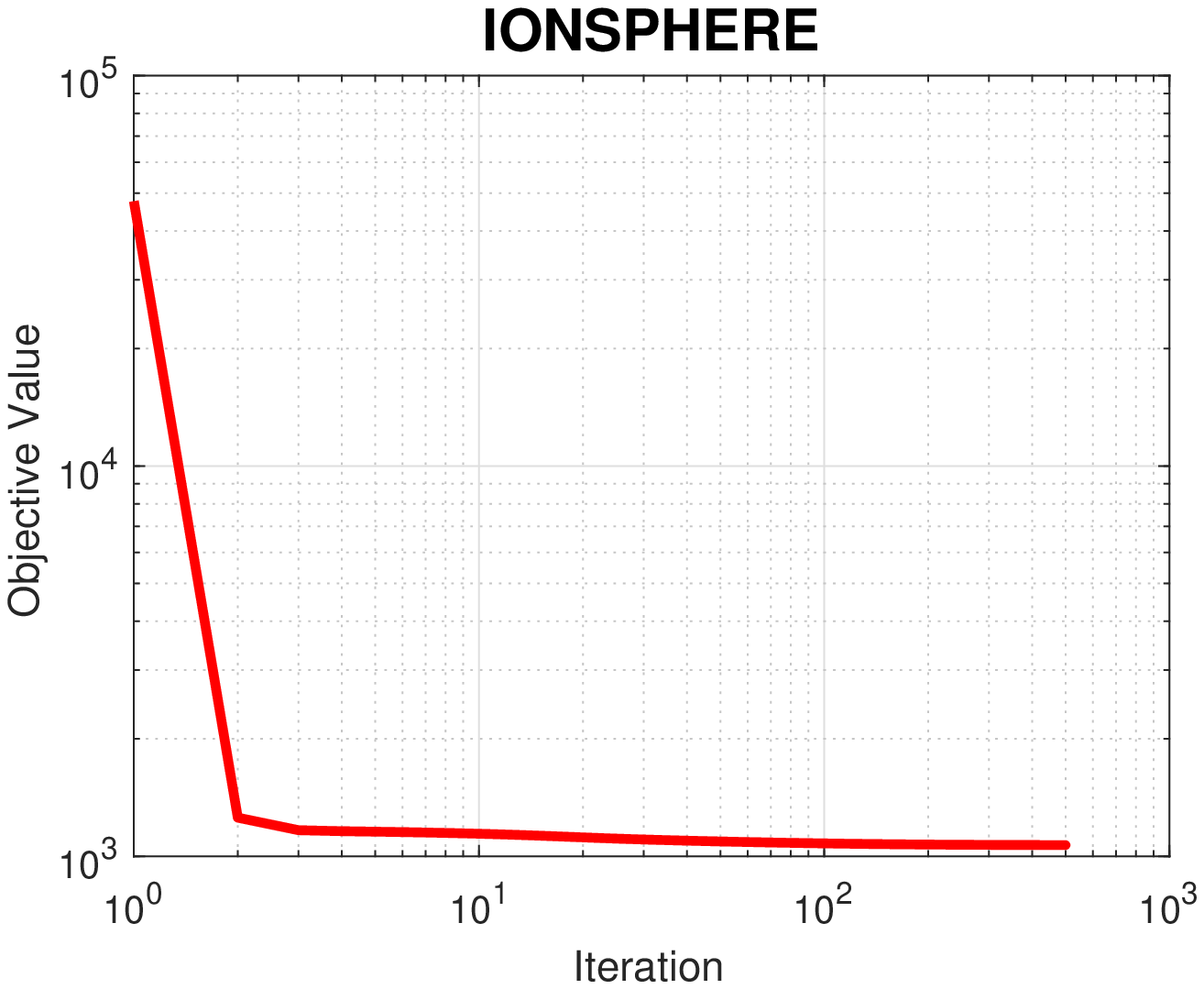,width=3.9cm}}
		%		\vspace{-0.1cm}
		%		\centerline{(a) COIL20}\medskip
	\end{minipage}
	\begin{minipage}[b]{0.195\linewidth}
		\centering
		\centerline{\epsfig{figure=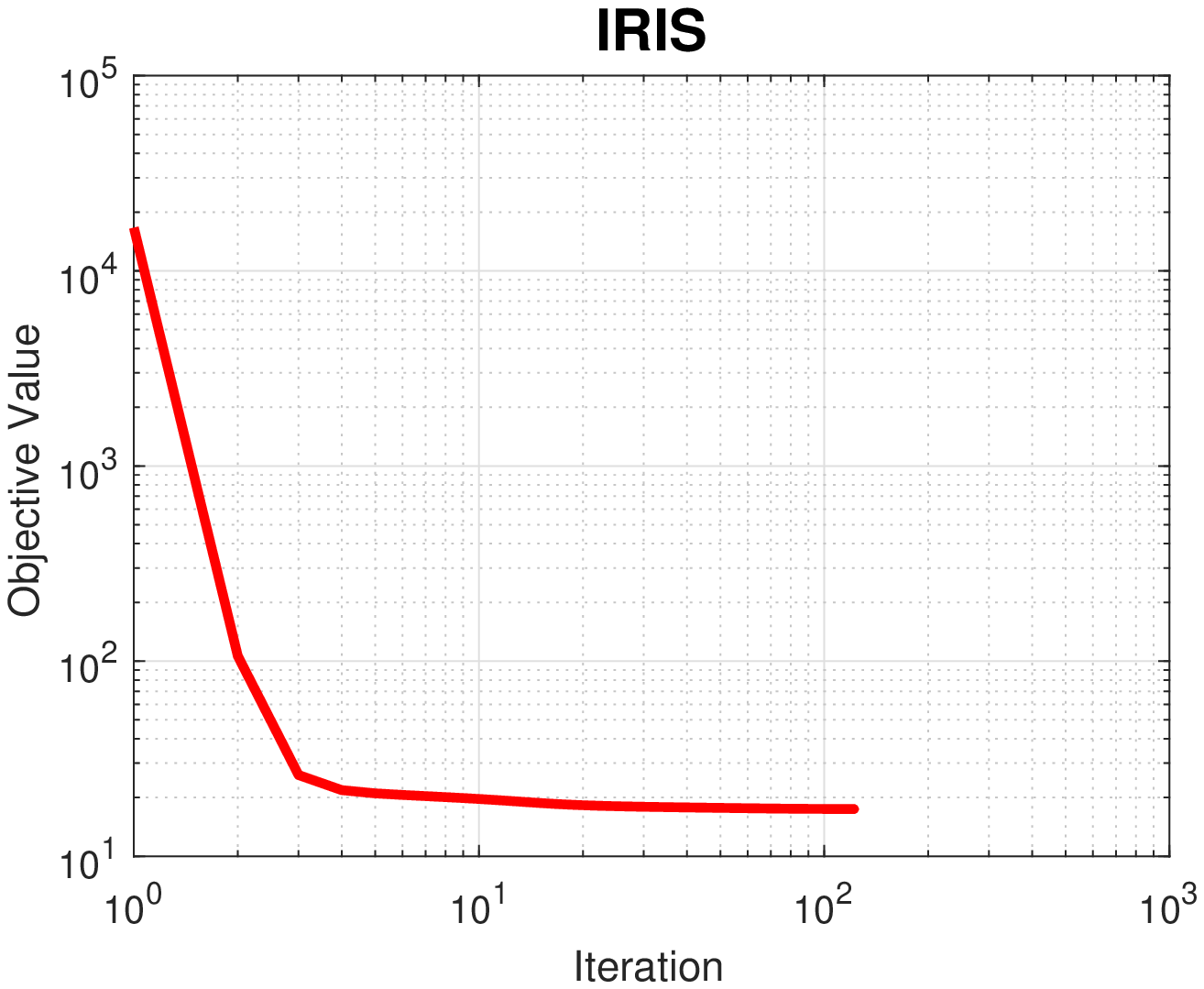,width=3.9cm}}
		%		\vspace{-0.1cm}
		%		\centerline{(a) COIL20}\medskip
	\end{minipage}
	\begin{minipage}[b]{0.195\linewidth}
		\centering
		\centerline{\epsfig{figure=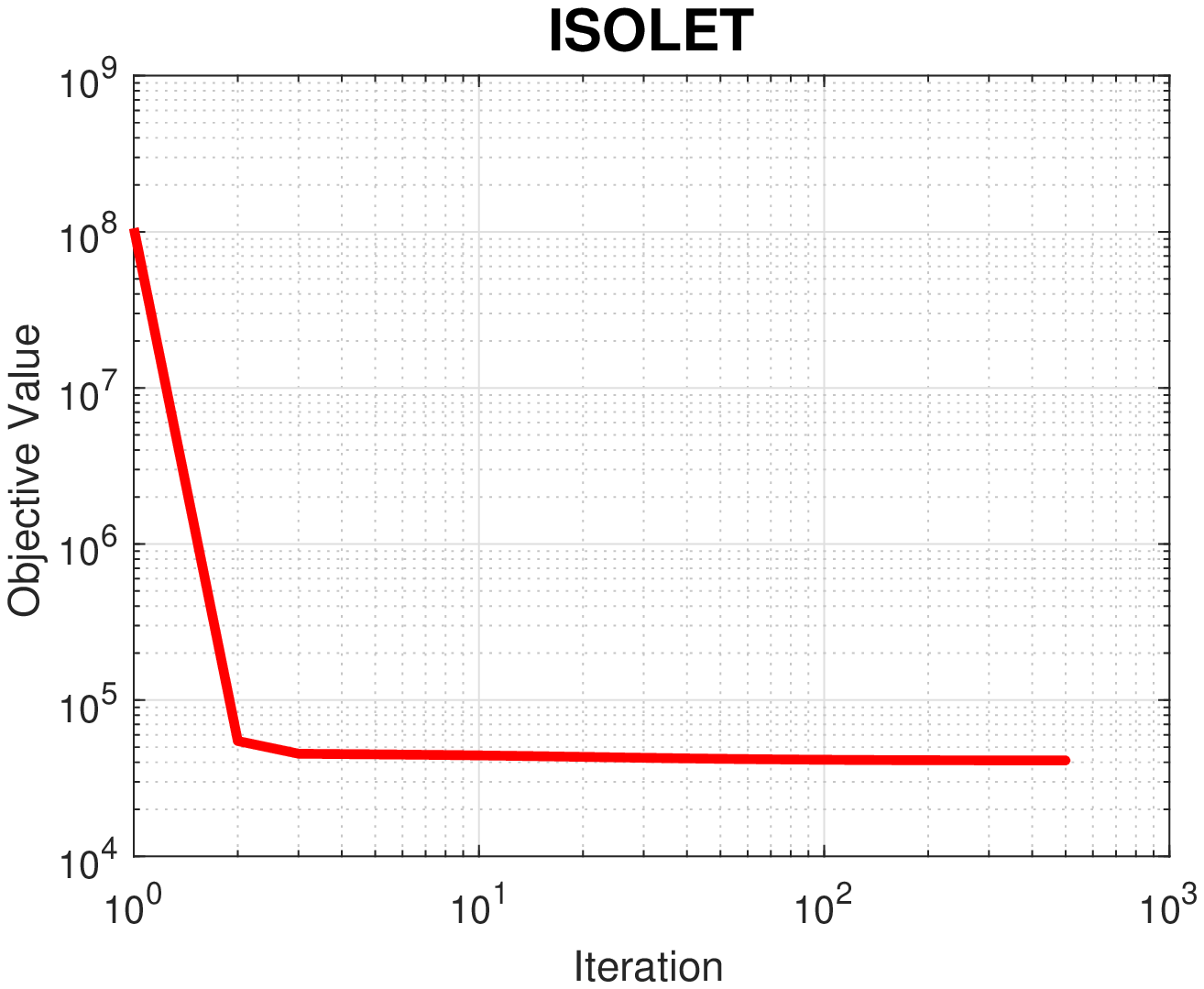,width=3.9cm}}
		%		\vspace{-0.1cm}
		%		\centerline{(a) COIL20}\medskip
	\end{minipage}
	\begin{minipage}[b]{0.195\linewidth}
		\centering
		\centerline{\epsfig{figure=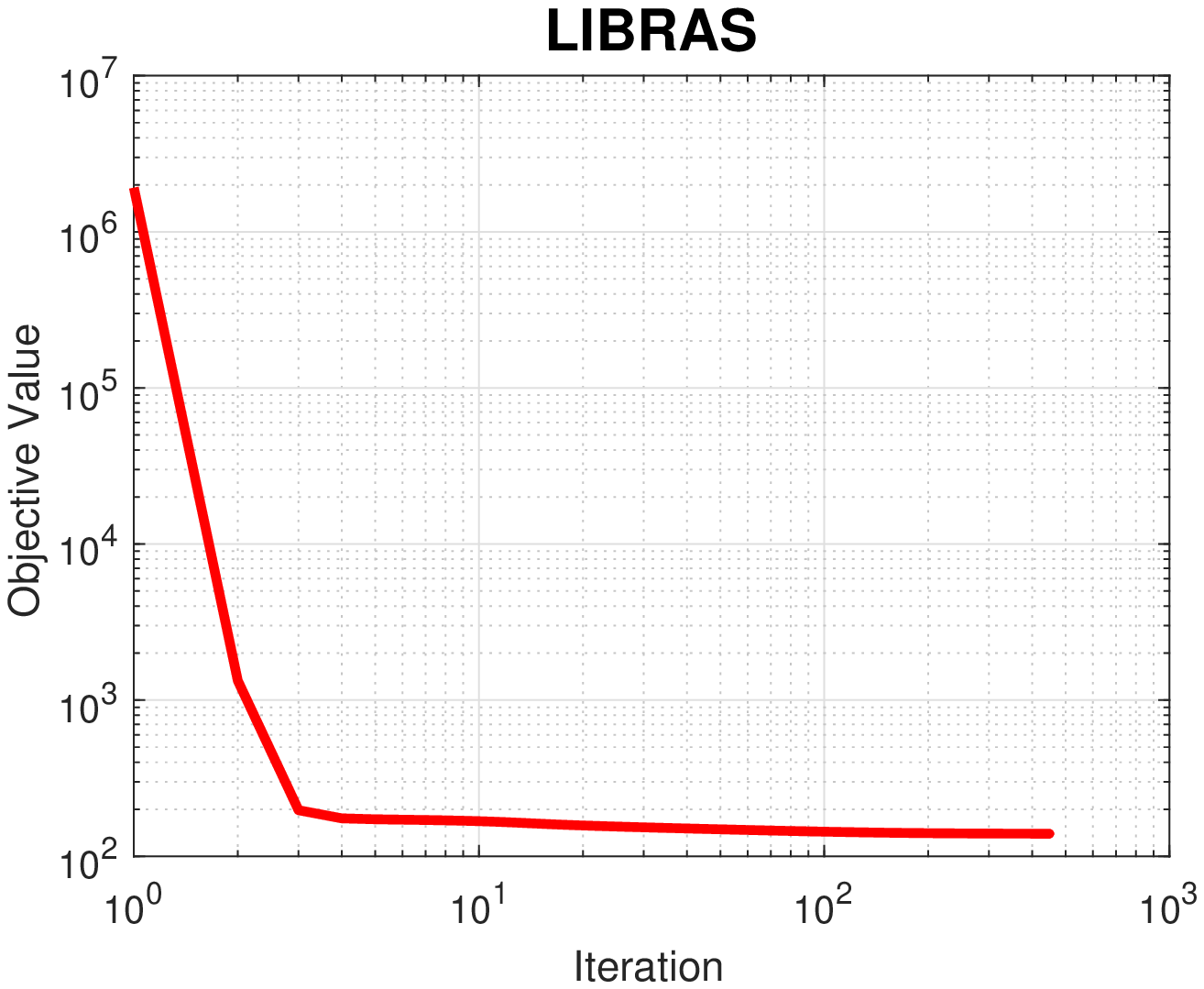,width=3.9cm}}
		%		\vspace{-0.1cm}
		%		\centerline{(a) COIL20}\medskip
	\end{minipage}
	\begin{minipage}[b]{0.195\linewidth}
		\centering
		\centerline{\epsfig{figure=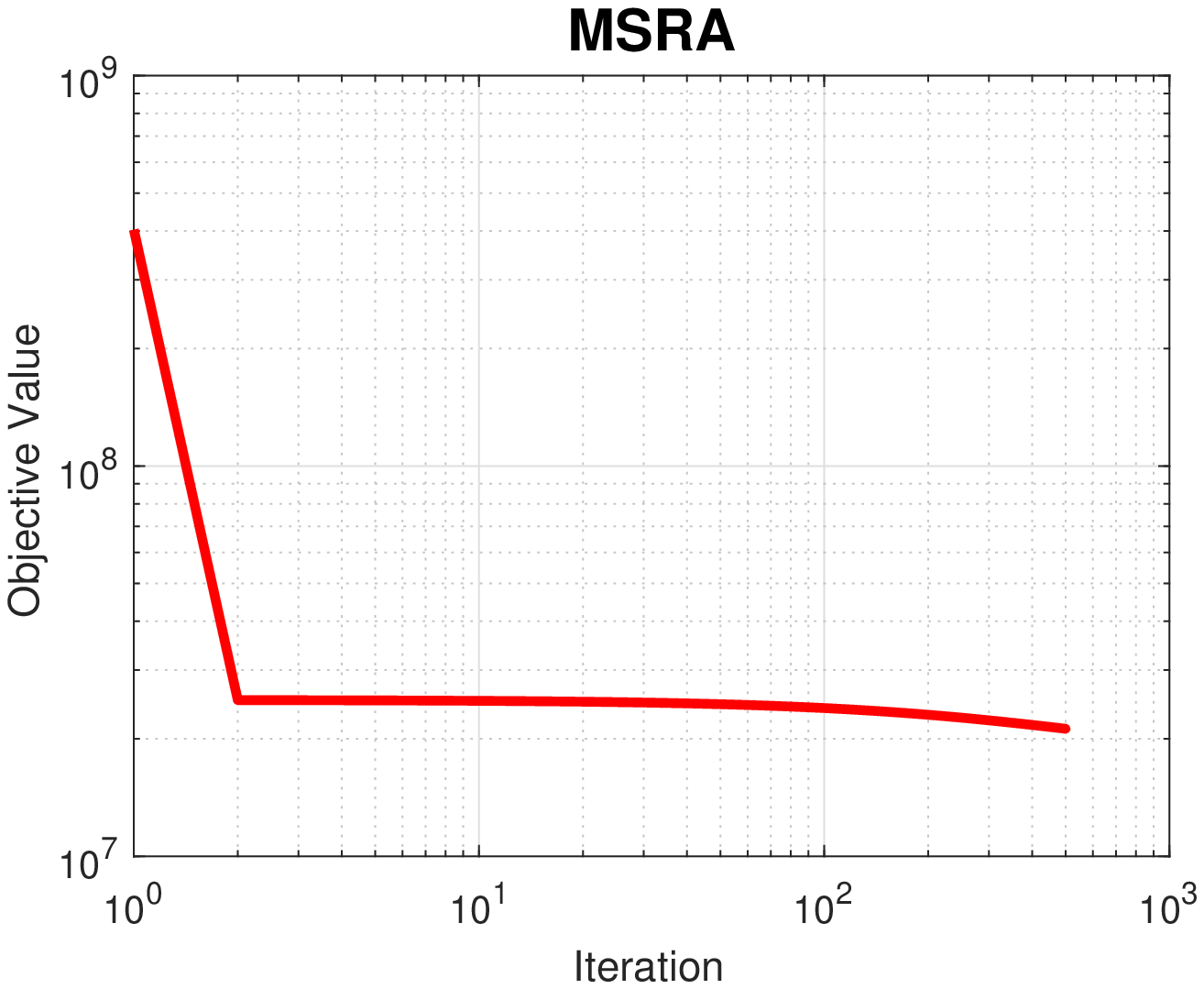,width=3.9cm}}
		%		\vspace{-0.1cm}
		%		\centerline{(a) COIL20}\medskip
	\end{minipage}
	\begin{minipage}[b]{0.195\linewidth}
		\centering
		\centerline{\epsfig{figure=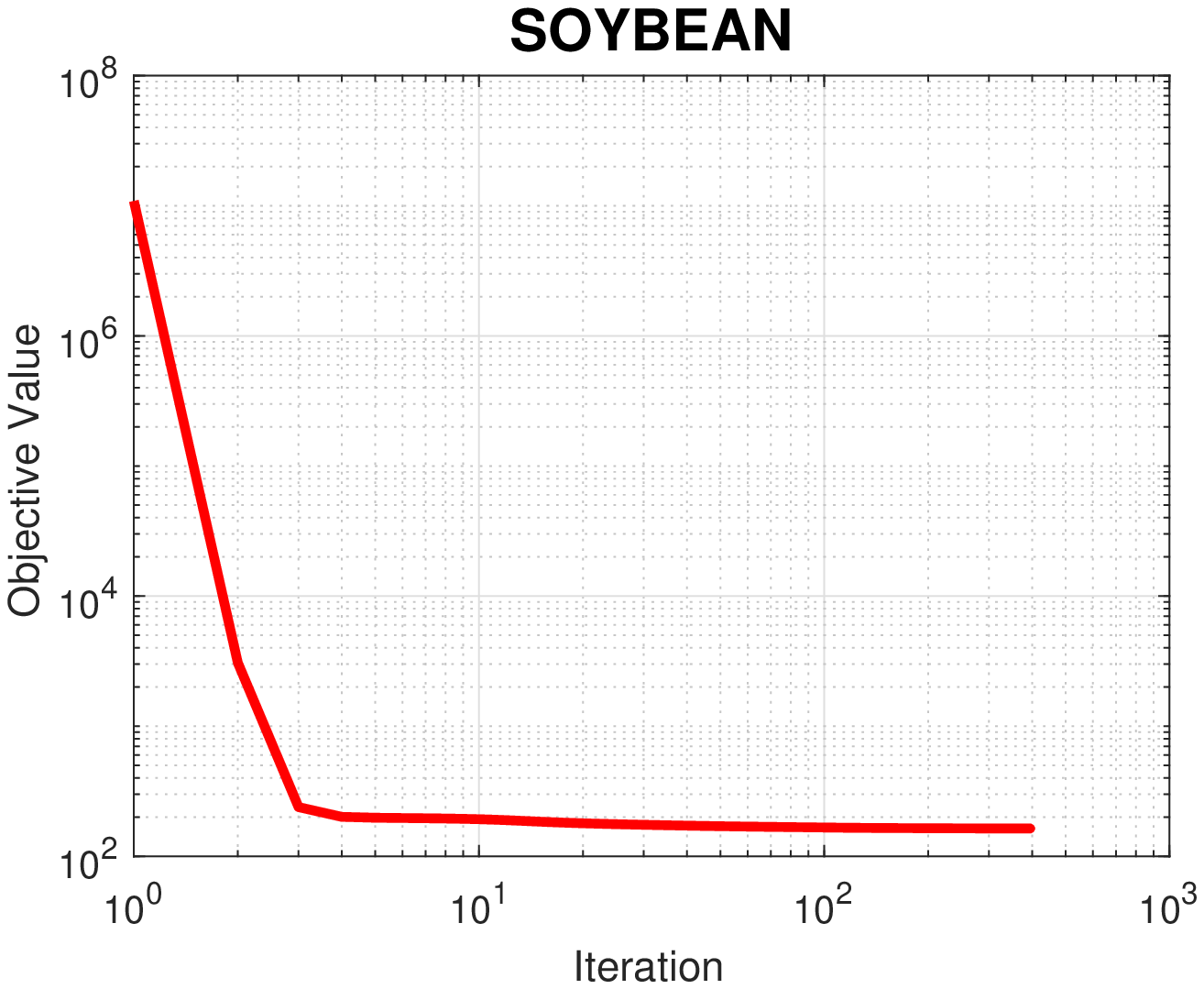,width=3.9cm}}
		%		\vspace{-0.1cm}
		%		\centerline{(a) COIL20}\medskip
	\end{minipage}
	\begin{minipage}[b]{0.195\linewidth}
		\centering
		\centerline{\epsfig{figure=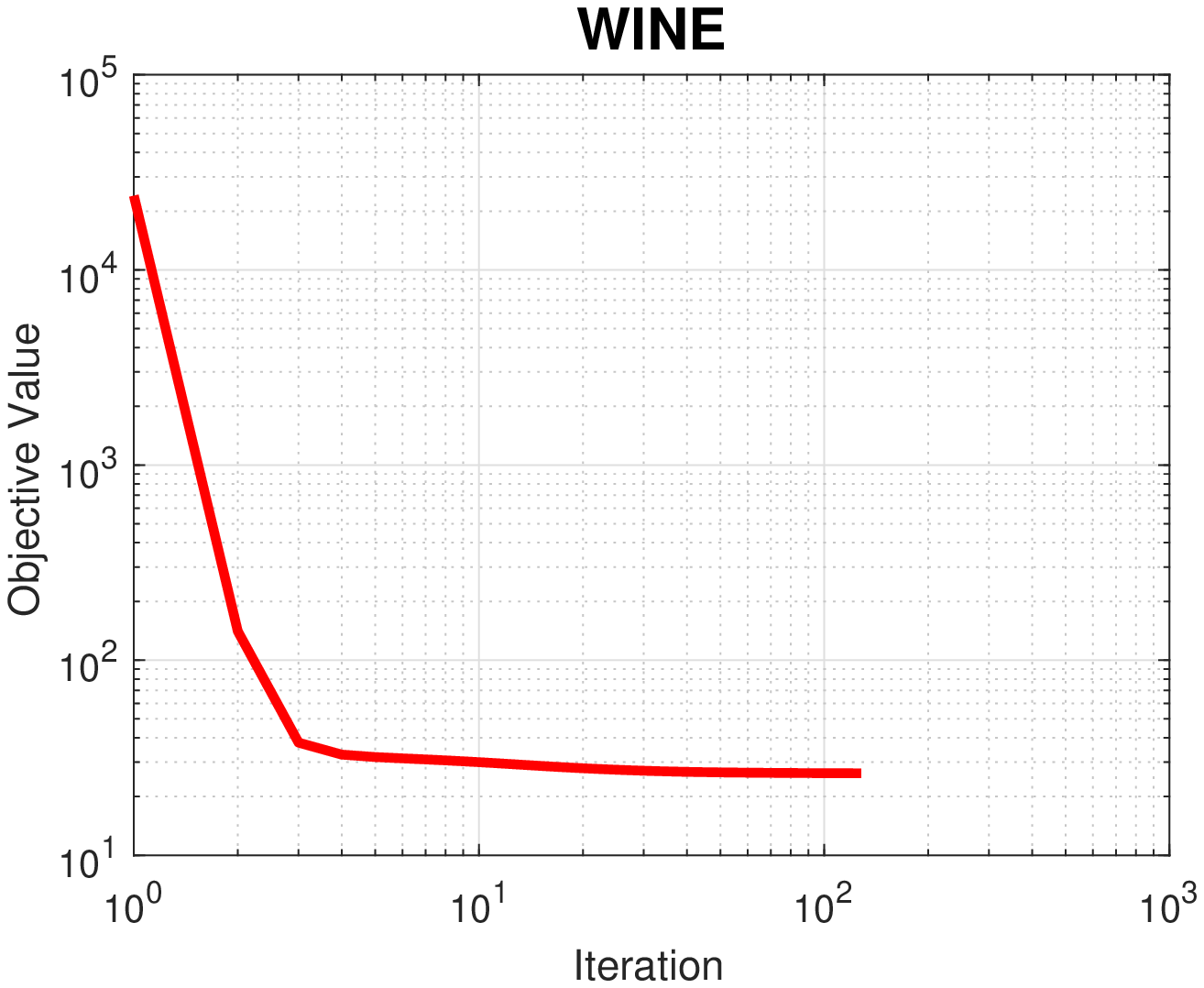,width=4cm}}
		%		\vspace{-0.1cm}
		%		\centerline{(a) COIL20}\medskip
	\end{minipage}
	\begin{minipage}[b]{0.195\linewidth}
		\centering
		\centerline{\epsfig{figure=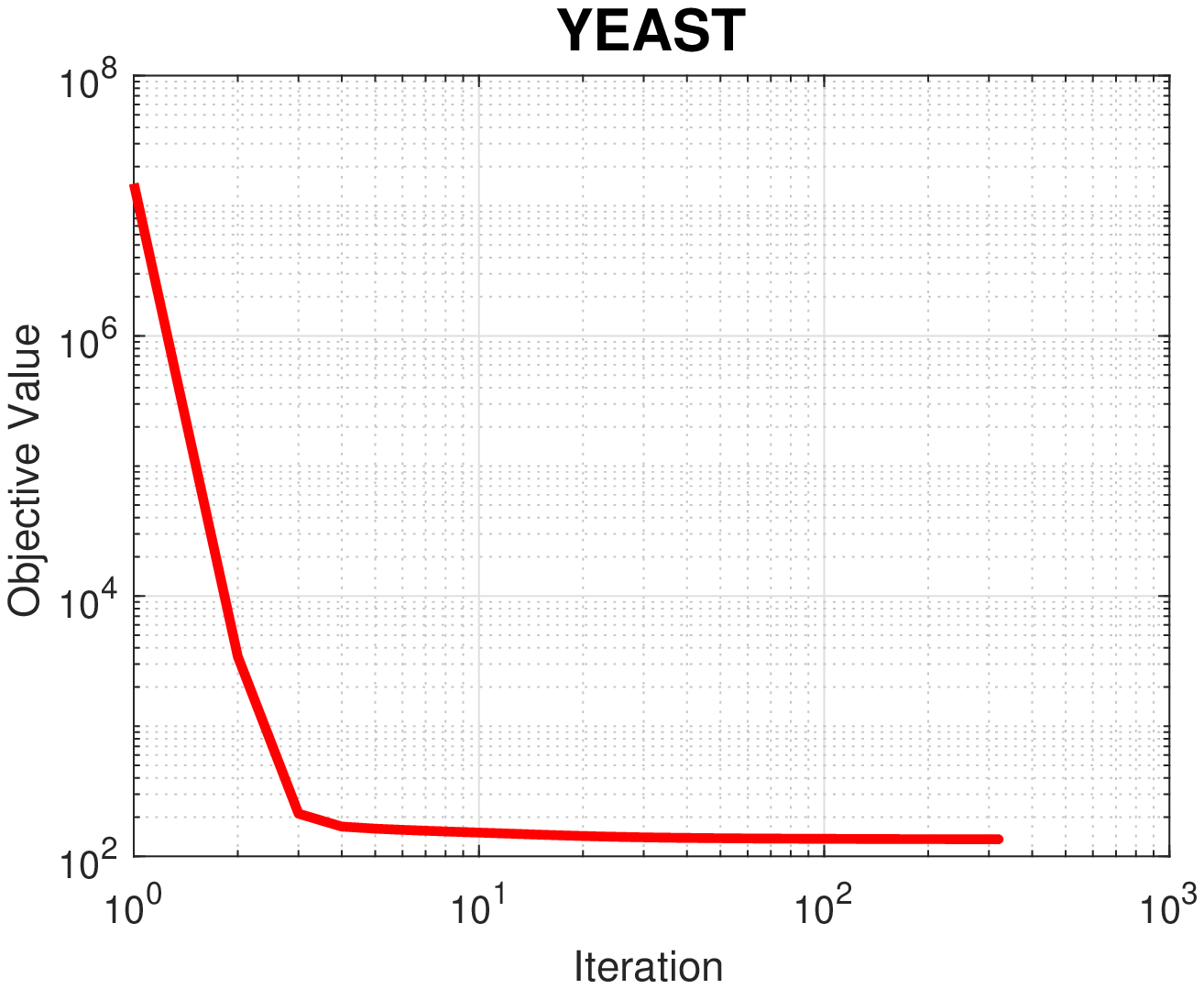,width=4cm}}
		%		\vspace{-0.1cm}
		%		\centerline{(a) COIL20}\medskip
	\end{minipage}
	\caption{Illustration of objective values against the number of iteration on 10 datasets. }
	\label{fig:convergence}
\end{figure*}

\begin{figure*}[!t]

		\centering
%	\centerline{\epsfig{figure=running_time_r1,width=18cm}}
%	\centerline{\epsfig{figure=running_time_r1-c.eps,width=18cm}}
	\includegraphics[width=0.95\linewidth]{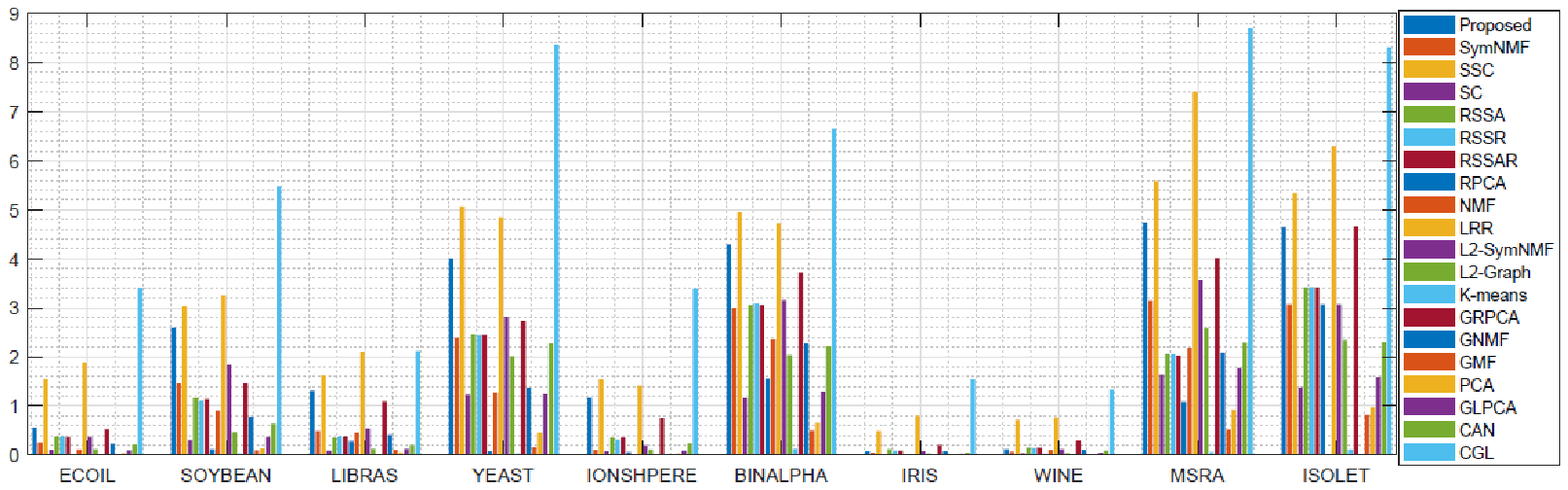}
		%		\vspace{-0.1cm}
		%		\centerline{(a) COIL20}\medskip

	\caption{\textcolor{black}{Illustration of the running time comparison for all the methods on different datasets.}}
	\label{fig:time}
\end{figure*}
%\subsubsection{Experiment Settings}
%The experimental setting are summarized as follows. 
%
%Clustering results analysis. 
Tables II-XI show the clustering performance of different methods, and Tables XII-XIII summarize the overall performance of the proposed model on all the datasets.
From those tables, we have the following conclusions.
\begin{enumerate}
	\item The proposed model always has higher ACC/NMI/PUR/ARI than SymNMF over all the datasets. 
	Especially, on IRIS, the ACC increases more than $35\%$ compared  with SymNMF. 
	Moreover, according to the Wilcoxon rank sum test, the improvements  under all the cases (40/40) are significant,  which validates our basic assumption that the predefined similarity graph is usually not the best choice. By learning a reasonable graph from raw features, the proposed model can  generate the graph with higher quality. 
	\item  NMF and GNMF require the input data to be nonnegative, so they are not applicable to  IONSHPERE and ISOLET due to that IONSHPERE and ISOLET consist of
	mixed signed data. Although our model also contains the nonnegative constraints on $\mathbf{S}$ and $\mathbf{V}$, it can  cope with the mixed sign data by separating the negative and positive components in $\mathbf{X}^\mathsf{T}\mathbf{X}$. Therefore, our method is more flexible than NMF-like methods. 
	\item SymNMF  performs better than SC in most cases (29/40). Taking IRIS as an example, ACC increases $45\%$ and PUR increases $105\%$. 
	Note that both SymNMF and SC utilize the same predefined   graph  in the experiments, the advantage of SymNMF over SC validates that directly generating data partition is beneficial to clustering. %, which is also adopted in our model. 
	\item CAN is a graph-based clustering method with an adaptive graph according to the raw features, and SSC is an advanced SC method based on the predefined graph. CAN  performs better than SSC on ECOIL, LIBRAS, YEAST and MSRA, while SSC performs better than CAN on IONSHPERE, ISOLET, SOYBEAN, BINALPHA, WINE and IRIS. This phenomenon demonstrates that  both the raw features and the predefined graph are important to clustering if they are well exploited.  
	
	%	\item subspace models vs RSSA, the advantages of Joint model
	%\item the importance of both local and global structures
	\item Real world datasets usually contain different types of noise and outliers, so the models that are robust to noise and outliers may  produce high quality clustering results. 
	For example, the robust models like RPCA and GRPCA  get quite well  performance on ECOIL, YEAST, IONSPHERE, IRIS, BINAPHPA and ISOLET. 
	%	Moreover, according to IPD properties of L2-norm \cite{peng2017constructing}, the L2-graph also establish high quality partition on those  datasets. From the view of IPD, our model is also robust to noise to some extend. 
	Moreover, according to the IPD property of L2 norm \cite{peng2017constructing}, the L2-Graph  is also robust to noise, which is also applicable to our model. 
	
	\item The methods with a graph regularizer  usually achieve better performance than the original models. 
	For example, GNMF and GMF perform better than NMF and  PCA, respectively.
	This phenomenon exposes the importance of exploiting the local  structures in clustering. 
	%	For example, the comparison between GNMF with NMF, and GMF with PCA can support this statement. 
	%	\item 
	Both PCA and NMF can be regarded as the variants of K-means \cite{ding2005equivalence,ding2004k} in a soft manner, and the graph regularizer  is highly related to spectral clustering \cite{kong2012iterative}. 
	These phenomena  suggest that graph-based clustering  can usually generate better clustering performance than K-means-like methods.
	
	%	\item best clustering performance
	%	\item The proposed perform significantly better than L2-graph on the majority cases ($34/40$) according to the Wilcoxon rank-sum test. 
	%	The proposed model obtain higher performance than L2-SymNMF on $(34/40)$, and $(30/40)$ of them are significantly better.
	%	For example, the proposed model generate approximate $50\%$ higher ACC than L2-graph and L2-SymNMF on ECOIL.  
	%	Both L2 graph L2-SymNMF learn  graph from the raw features with a Frobenius norm on the coefficient matrix like our model. However, our method is processed in a joint manner with cluster learning. 
	%	The superior performance of our model than L2 graph L2-SymNMF verifies the benefit of  learning graph and cluster indicator in a joint manner.
	\item The proposed model is significantly better than L2-Graph in most cases ($34/40$) in accordance with the Wilcoxon rank-sum test. 
	It also obtains better performance than L2-SymNMF in most cases ($34/40$). Especially,  $30$ out of $40$ cases of them are significantly better.
	Taking  ECOIL as an example, the proposed model generates approximate $50\%$ higher ACC than L2-Graph and L2-SymNMF.  
	Both L2-Graph and  L2-SymNMF learn  the graph from the raw features with a Frobenius norm on the coefficient matrix like our model. While our method is processed in a joint manner. 
	%	Those phenomena  verify the benefit of  the proposed joint manner.
	Those phenomena  verify that the constructed graph in our model is more suitable for  clustering.
	\item Table \ref{tab:res-summarization} summarizes  the ranking of the proposed model among all the 20 methods on all the datasets with different metrics. What is noteworthy in Table  \ref{tab:res-summarization} is that the proposed model ranks the first under 23 out of 40 cases ($57.5\%$). Moreover, the proposed model ranks top 3 under $90.0\%$ cases.  
	Table \ref{tab:res-sig} sums up whether the proposed method is significantly better,  or worse than the compared methods. It is apparent from Table \ref{tab:res-sig} that the proposed model performs significantly better  than the compared methods in approximate $90\%$  of  cases; and only in less than $3\%$  of cases that the proposed model  gets significantly worse results.
	The above analyses support the conclusion that the proposed model produces better clustering performance than the compared 19 models on those 10 datasets.
	%	\item The proposed model can be regarded as a special case of the kernel version with a linear inner product kernel, which only exploits the linear relation in the raw features. 
	%	When apply a RBF kernel, the nonlinear kernel version can generate higher clustering performance than the linear one on $65\%$ cases. Especially on IONSPHERE, the improvements on ACC, NIM, PUR, ARI are $4.6\%$, $22.6\%$, $4.2\%$ and $23.3\%$, respectively. 
\end{enumerate} 

\subsection{Visual Comparison of Learned Affinity Matrices}

\textcolor{black}{We visually show the affinity matrices constructed by the $p$NN graph, the L2-Graph, the proposed method, and the ground truth relationship on the IRIS dataset in Fig. \ref{fig:graph}. The IRIS dataset contains three categories, and the second and
	third categories are quite close to each other, which is challenging
	to distinguish.
Note that the L2-Graph \cite{peng2017constructing} and L2-SymNMF share the same affinity matrix in Fig. \ref{fig:graph}-(d), while SC \cite{ng2002spectral} and SymNMF \cite{kuang2012symmetric,kuang2015symnmf} share the same affinity matrix in Fig. \ref{fig:graph}-(a). 
Since the inner product of the lower-dimensional embeddings $\mathbf{V}^\mathsf{T}\mathbf{V}$ can indicate the similarity relations among samples directly, we also show  $\mathbf{V}^\mathsf{T}\mathbf{V}$ for different methods, including SC, SymNMF, L2-Graph, L2-SymNMF and our method.}

\textcolor{black}{Compared with the $p$NN graph and L2-Graph, the affinity matrix by our model exhibits denser correct connections and a more obvious block diagonal structure. 
Moreover, SC with the $p$NN graph as the affinity matrix  cannot distinguish the second category from the third category as shown in Fig. \ref{fig:graph}-(b). SymNMF performs slightly better than SC; however, there are still many incorrect connections as shown  in Fig.  \ref{fig:graph}-(c). In contrast to all the compared methods, $\mathbf{V}^\mathsf{T}\mathbf{V}$ of our method appears a much better block diagonal structure, which visually validates the advantage of our method. }

\subsection{Parameter Sensitivity Analysis}
There are two hyper-parameters in the proposed model, i.e., $\alpha$  and $\beta$ which adjust the contributions to graph construction from the raw features and the predefined graph, respectively. 
Fig. \ref{fig:Para} plots the values of  ACC w.r.t. different  $\alpha$  and $\beta$, where we can see that 
%As analyzed in Section VI-B, different \red{it should connect to the previous analysis}
\begin{enumerate}
	\item the highest ACC never occurs when $\alpha=0$ or $\beta=0$, which indicates that both $\alpha$ and $\beta$ are critical  to the proposed model. Moreover, the lowest value always  appears when both $\alpha=0$ and $\beta=0$. The reason is that,  when $\alpha=\beta=0$, no useful information can be transferred to the cluster membership matrix $\mathbf{V}$. 
	\item the optimal ACCs of all the datasets usually occur in a common range, i.e.,  $\alpha\in\{0.1,10\}$, and $\beta\in\{1,100\}$, which validates the robustness of our model to the hyper-parameters. 
	\item \textcolor{black}{nevertheless, how to adaptively  determine the optimal $\alpha$ and $\beta$ based on the characteristic of the input data is still a challenging problem. One possible solution is to resort to the methodology of Bayesian inference, which defines an explicit prior probability distribution over $\alpha$ and $\beta$, and then infers them by maximizing the type II likelihood \cite{tan2012automatic,jia2019sparse}}. 
\end{enumerate}

\subsection{Convergence Analysis}
The convergence of the proposed optimization algorithm has been theoretically  proven in Section III-D. Here,  we study its empirical convergence behavior. Specifically,  
Fig. \ref{fig:convergence} shows the objective function values according to the iteration number on all the datasets when $\alpha =1$ and $\beta=1$, 
%Fig. \ref{fig:convergence} plots objective function values against the increasing of the iteration numbers with $\alpha=1$ and $\beta=1$ on all the datasets, 
from which we can observe that the values of the objective function decrease monotonically on all the datasets with the increase of the number of iterations, which is consistent with the theoretical analysis.
%Moreover, on all the datasets the objective values decline rapidly in the first few iterations and get convergent in approximate 100 iterations,  which illustrates the high efficiency of our optimization method. 
Moreover, on all the datasets the objective values get convergent in approximately 100 iterations,  which illustrates the high efficiency of our optimization algorithm. 

%Usually less than 100 iterations  
\subsection{Comparison of Running Time}
The running time comparisons are shown in Fig.  \ref{fig:time} for all the methods. From  Fig.  \ref{fig:time}, we have the following observations.
\begin{enumerate}
	\item The proposed method is usually faster than SSC and LRR, and comparable to GRPCA. The reason is that SSC needs to compute the spectral decomposition of an $n\times n$ matrix many times, and both LLR and GRPCA  need to compute the SVD of  an $n\times n$ matrix repeatedly. Note that the computational complexities of both spectral decomposition and  SVD are as high as  $\mathsf{O}(n^3)$.
	\item Our model is only slightly slower than SymNMF. Taking the superior clustering performance of our model to SymNMF into consideration, sacrificing  a little training time is acceptable.
	\item The number of samples  determines how much time our model will take. How to reduce the computational complexity of our model is left for our future work.
%	\item comparable to SymNMF method.
\end{enumerate}

\section{Conclusion}
In this paper, we have presented a  graph-based clustering model that can  learn the  graph  and  partition the data   simultaneously.
 Since these two  tasks are optimized in a joint manner, the constructed graph is tailored to  the task of clustering. Therefore, the clustering performance can be further improved.
%  manner, which exploits the mutual enhancement relation between them. The graph matrix is adaptively constructed by collecting the information from the original graph, the raw features and the clustering membership matrix.  
%The proposed model is extended to a kernel version to explore the nonlinear relation in raw features. 
%In addition, the proposed model is solved via an alternative optimization method, which can converge to the KKT points under some mild conditions. 
\textcolor{black}{In addition, the proposed model is numerically  solved via an alternating iterative optimization algorithm, where the constraints can be naturally satisfied. }
The extensive experimental results demonstrate that the proposed model can achieve much better clustering performance than 19 state-of-the-art methods.
%Extensive experiments indicate that the proposed model has better clustering ability than 19 compared  state-of-the-art methods significantly  according to the Wilcoxon rank-sum test. 

The proposed model explores the information from raw features in a linear manner, i.e., ${\rm min}_{\mathbf{S}}\|\mathbf{X}-\mathbf{XS}\|_F^2$. 
Since \eq{update-V} only relates to the inner product of the input (i.e., $\mathbf{X}^\mathsf{T}\mathbf{X}$), the proposed model has potential for exploiting  the non-linear relation from the raw features with a kernel trick, which will be investigated in our future work.%  could use a kernel trick to exploit the non-linear relationship in the feature space. 

\appendices
\section{Proof of Theorem 1}
\subsection{Proof of \textbf{Theorem 1}-1}
%We first give the definition of the upper-bound auxiliary function, and the following lemma,
According to \cite{lee2001algorithms}, the following \textbf{Lemma 1} and \textbf{Definition 1} can be used to prove \textbf{Theorem 1}-1.

\noindent
\textcolor{black}{\textbf{Definition 1} \emph{$g(h,h')$ is an upper-bound auxiliary function for $f(h)$ if the following two conditions are satisfied}
\begin{equation}
g(h,h')\geq f(h), {\rm and}~ g(h=h',h')=f(h').
\end{equation}}\textcolor{black}{\textbf{Lemma 1} \emph{If $g$ is an upper-bound auxiliary function of $f$,  $f$ is decreasing\footnote{non-increasing, to be precise.} under the update }}
\begin{equation}
\textcolor{black}{h^*=\argmin_{h}g(h,h').}
\end{equation}
See the proof of \textbf{Lemma 1}  at \cite{lee2001algorithms}. 
Based on \textbf{Lemma 1}, if we can find appropriate upper-bound auxiliary functions for \eq{obj} w.r.t. $\mathbf{S}$ (with the fixed $\mathbf{V}$) and $\mathbf{V}$ (with the fixed $\mathbf{S}$), respectively, then show the updating rules in \eq{update-S} and \eq{update-V} decrease the corresponding upper-bound functions, \textcolor{black}{and together with the fact that \eq{obj} is lower-bounded,  \textbf{Theorem 1}-1 can be proved. }

%We first rewrite the objective function in \eq{obj} here:
%\begin{equation}
%\begin{split}
%\mathcal{O}&=\|\mathbf{S}-\mathbf{VV}^\mathsf{T}\|_F^2+\beta\|\mathbf{S}-\mathbf{W}\|^2_F+\alpha\|\mathbf{X}-\mathbf{XS}\|_F^2\\
%%&={\rm Tr}\left(\left(\mathbf{S}-\mathbf{VV}^\mathsf{T}\right)\left(\mathbf{S}-\mathbf{VV}^\mathsf{T}\right)^\mathsf{T}\right)\\
%%&~~~~+\alpha{\rm Tr}\left(\left(\mathbf{X}-\mathbf{XS}\right)\left(\mathbf{X}-\mathbf{XS}\right)^\mathsf{T}\right)+\beta{\rm Tr}\left(\left(\mathbf{S}-\mathbf{W}\right)\left(\mathbf{S}-\mathbf{W}\right)^\mathsf{T}\right)\\
%&={\rm Tr}\left(\mathbf{SS}^\mathsf{T}-2\mathbf{VV}^\mathsf{T}\mathbf{S}^\mathsf{T}+\mathbf{VV}^\mathsf{T}\mathbf{VV}^\mathsf{T}\right)\\
%&~~~~+\alpha{\rm Tr}\left(\mathbf{XX}^\mathsf{T}-2\mathbf{X}^\mathsf{T}\mathbf{XS}^\mathsf{T}+\mathbf{SS}^\mathsf{T}\mathbf{X}^\mathsf{T}\mathbf{X}\right)\\
%&~~~~+\beta{\rm Tr}\left(\mathbf{SS}^\mathsf{T}-2\mathbf{WS}^\mathsf{T}+\mathbf{WW}^\mathsf{T}\right).\\
%\end{split}
%\end{equation}
Excluding terms  uncorrelated to $\mathbf{S}$, the objective function w.r.t. $\mathbf{S}$ is written as 
\begin{equation}
\begin{aligned}
\mathcal{O}_\mathbf{S}=&{\rm Tr}
\Bigl( (1+\beta)\mathbf{SS}^\mathsf{T}+\alpha\mathbf{SS}^\mathsf{T}\mathbf{X}^\mathsf{T}\mathbf{X}-2\alpha\mathbf{X}^\mathsf{T}\mathbf{XS}^\mathsf{T}\\
&~~-2\left(\mathbf{VV}^\mathsf{T}+\beta\mathbf{W}\right)\mathbf{S}^\mathsf{T}\Bigr)+{\rm const.}\\
\propto&{\rm Tr}\left(\mathbf{S}^\mathsf{T}\mathbf{AS}-\mathbf{S}^\mathsf{T}\mathbf{BS}-\mathbf{S}^\mathsf{T}\mathbf{C}+\mathbf{S}^\mathsf{T}\mathbf{D}\right),
\end{aligned}
\label{S-obj}
\end{equation}
%\begin{small}
%	\begin{equation}
%	\begin{split}
%	\mathcal{O}_\mathbf{S}
%%	&={\rm Tr}\left(\mathbf{SS}^\mathsf{T}-2\mathbf{VV}^\mathsf{T}\mathbf{S}^\mathsf{T}+\mathbf{VV}^\mathsf{T}\mathbf{VV}^\mathsf{T}\right)\\
%%	&~~~~+\alpha{\rm Tr}\left(\mathbf{XX}^\mathsf{T}-2\mathbf{X}^\mathsf{T}\mathbf{XS}^\mathsf{T}+\mathbf{SS}^\mathsf{T}\mathbf{X}^\mathsf{T}\mathbf{X}\right)\\
%%	&~~~~+\beta{\rm Tr}\left(\mathbf{SS}^\mathsf{T}-2\mathbf{WS}^\mathsf{T}+\mathbf{WW}^\mathsf{T}\right)\\
%	&={\rm Tr}\left((1+\beta)\mathbf{SS}^\mathsf{T}+\alpha\mathbf{SS}^\mathsf{T}\mathbf{X}^\mathsf{T}\mathbf{X}-2\left(\mathbf{VV}^\mathsf{T}+\beta\mathbf{W}\right)\mathbf{S}^\mathsf{T}-2\alpha\mathbf{X}^\mathsf{T}\mathbf{XS}^\mathsf{T}\right)\\
%	&~~~+{\rm const.}\\
%	&\propto{\rm Tr}\left(\mathbf{S}^\mathsf{T}\mathbf{AS}-\mathbf{S}^\mathsf{T}\mathbf{BS}-\mathbf{S}^\mathsf{T}\mathbf{C}+\mathbf{S}^\mathsf{T}\mathbf{D}\right)
%	\end{split}
%	\label{S-obj}
%	\end{equation}
%\end{small}where
where $\mathbf{A}=(1+\beta)\mathbf{I}+\alpha\left(\mathbf{X}^\mathsf{T}\mathbf{X}\right)^+$, $\mathbf{B}=\alpha\left(\mathbf{X}^\mathsf{T}\mathbf{X}\right)^-$
, $\mathbf{C}=2\mathbf{VV}^\mathsf{T}+2\beta\mathbf{W}+2\alpha\left(\mathbf{X}^\mathsf{T}\mathbf{X}\right)^+$, and $\mathbf{D}=2\mathbf{B}=2\alpha\left(\mathbf{X}^\mathsf{T}\mathbf{X}\right)^-$.

For the $\mathbf{V}$-block, the corresponding objective function is
\begin{equation}
\begin{split}
\mathcal{O}_\mathbf{V}
&={\rm Tr}\left(-2\mathbf{VV}^\mathsf{T}\mathbf{S}^\mathsf{T}+\mathbf{VV}^\mathsf{T}\mathbf{VV}^\mathsf{T}\right)+{\rm const}\\
&\propto{\rm Tr}\left(-2\mathbf{VV}^\mathsf{T}\mathbf{S}^\mathsf{T}+\mathbf{VV}^\mathsf{T}\mathbf{VV}^\mathsf{T}\right).
\end{split}
\label{V-obj}
\end{equation}
The adopted  upper-bound auxiliary functions for Eqs. \eqref{S-obj} and \eqref{V-obj}  are given in the following two lemmas. 

%In this section, we first \eq{update-S} decreases \eq{obj} in each iteration. The corresponding upper-bound auxiliary function is given in Lemma 2. 
\noindent
\textbf{Lemma 2} \emph{The upper bound auxiliary function for \eq{S-obj}  is}\textcolor{black}{
\begin{equation}
\begin{split}
&f_s(\mathbf{S},\mathbf{S}')=-\sum_{ijk}\mathbf{B}_{ik}\mathbf{S}_{kj}'\mathbf{S}_{ij}'\left(1+{\rm log}\frac{\mathbf{S}_{kj}\mathbf{S}_{ij}}{\mathbf{S}_{kj}'\mathbf{S}_{ij}'}\right)\\
&~~+\sum_{ij}^{}\frac{\left(\mathbf{A}\mathbf{S}'\right)_{ij}\mathbf{S}_{ij}^2}{\mathbf{S}_{ij}'}+\sum_{ij}^{}\mathbf{D}_{ij}\frac{\mathbf{S}_{ij}^2+\mathbf{S}_{ij}^{'2}}{2\mathbf{S}_{ij}'}\\
&~~-\sum_{ij}^{}\mathbf{C}_{ij}\mathbf{S}_{ij}'\left(1+{\rm log}\frac{\mathbf{S}_{ij}}{\mathbf{S}_{ij}'}\right).
\end{split}
\label{aux-S}
\end{equation}}\textbf{Lemma 3} \emph{The upper-bound auxiliary function for \eq{V-obj} is}
\begin{equation}
\begin{split}
&f_v\left(\mathbf{V},\mathbf{V}'\right)=\sum_{ij}^n\sum_{k=1}^{c}\left(\mathbf{V}'\mathbf{V}'^{\mathsf{T}}\right)_{ij}\mathbf{V}_{ik}'\times\frac{\mathbf{V}_{jk}^4}{\mathbf{V}_{jk}^{'3}}\\
&~~~~-2\sum_{ij}^n\sum_{k=1}^{c}\mathbf{S}_{ji}\mathbf{V}_{jk}'\mathbf{V}_{ik}'\left(1+{\rm log}\frac{\mathbf{V}_{jk}\mathbf{V}_{ik}}{\mathbf{V}_{jk}'\mathbf{V}_{ik}'}\right).
\end{split}
\label{aux-V}
\end{equation}

\noindent
\textbf{Proof of Lemma 2}: \textbf{Lemma 2} can be proved based on the following 4 
inequalities. 

\noindent
\textbf{Proposition 1} \emph{For any positive matrices $\mathbf{A}>0$, $\mathbf{B}>0$, $\mathbf{C}>0$, $\mathbf{D}>0$, $\mathbf{S}>0$ and  $\mathbf{S}'>0$, with $\mathbf{A}$ symmetric, the following equations hold:}
\begin{subequations}
%	\begin{cases}
		\begin{align}
	&{\rm tr}\left(\mathbf{S}^\mathsf{T}\mathbf{AS}\right)\leq\sum_{ij}^{}\frac{\left(\mathbf{A}\mathbf{S}'\right)_{ij}\mathbf{S}_{ij}^2}{\mathbf{S}_{ij}'}, \\
	&{\rm tr}\left(\mathbf{S}^\mathsf{T}\mathbf{BS}\right)\geq\sum_{ijk}\mathbf{B}_{jk}\mathbf{S}_{ki}'\mathbf{S}_{ji}'\left(1+{\rm log}\frac{\mathbf{S}_{ki}\mathbf{S}_{ji}}{\mathbf{S}_{ki}'\mathbf{S}_{ji}'}\right), \\
	&{\rm tr}\left(\mathbf{S}^\mathsf{T}\mathbf{C}\right)\geq\sum_{ij}^{}\mathbf{C}_{ij}\mathbf{S}_{ij}'\left(1+{\rm log}\frac{\mathbf{S}_{ij}}{\mathbf{S}_{ij}'}\right), \\
	&{\rm tr}\left(\mathbf{S}^\mathsf{T}\mathbf{D}\right)\leq\sum_{ij}^{}\mathbf{D}_{ij}\frac{\mathbf{S}_{ij}^2+\mathbf{S}_{ij}^{'2}}{2\mathbf{S}_{ij}'}.
	\end{align}
	\label{prop1}
\end{subequations}\emph{Moreover, all the equalities hold when $\mathbf{S}=\mathbf{S}'$.}

See the proofs of those inequalities in Appendix B. 
%\begin{equation}
%{\rm tr}\left(\mathbf{S}^\mathsf{T}\mathbf{C}\right)=\sum_{i,j}^{}\mathbf{S}_{i,j}\mathbf{C}_{i,j}\leq\sum_{i,j}^{}\mathbf{C}_{i,j}\frac{\mathbf{S}_{i,j}^2+\mathbf{S}_{i,j}^{'2}}{2\mathbf{S}_{i,j}^{'2}}
%\end{equation}
%
%\begin{equation}
%{\rm tr}\left(\mathbf{S}^\mathsf{T}\mathbf{D}\right)=\sum_{i,j}^{}\mathbf{D}_{i,j}\mathbf{S}_{i,j}\geq\sum_{i,j}^{}\mathbf{D}_{i,j}\mathbf{S}_{i,j}'\left(1+{\rm log}\frac{\mathbf{S}_{i,j}}{\mathbf{S}_{i,j}'}\right)
%\end{equation}
%
%
%
According to \textbf{Proposition 1}, \textbf{Lemma 2}  can be easily proved. To find the minimum of \eq{aux-S}, we take
\begin{equation}
\begin{split}
&\frac{\partial f_s(\mathbf{S}_{ij},\mathbf{S}_{ij}')}{\partial\mathbf{S}_{ij}}=\frac{2\left(\mathbf{A}\mathbf{S}'\right)_{ij}\mathbf{S}_{ij}}{\mathbf{S}_{ij}'}-\frac{\left(\mathbf{BS}'\right)_{ij}\mathbf{S}_{ij}'}{\mathbf{S}_{ij}}
\\
&~~~~~~~~-\frac{\left(\mathbf{B}^\mathsf{T}\mathbf{S}'\right)_{ij}\mathbf{S}_{ij}'}{\mathbf{S}_{ij}}+\mathbf{D}_{ij}\frac{\mathbf{S}_{ij}}{\mathbf{S}_{ij}'}-\mathbf{C}_{ij}\frac{\mathbf{S}_{ij}'}{\mathbf{S}_{ij}}.
\end{split}
\end{equation}
The detailed  calculation of those derivatives can be found in Appendix C. 
Moreover, the Hessian matrix containing the second order derivatives 
%\begin{equation}
%\begin{split}
%&\frac{\partial^2f_s(\mathbf{S},\mathbf{S}')}{\mathbf{S}_{i,j}\mathbf{S}_{l,k}}=\delta_{il}\delta_{jk}\frac{2\left(\mathbf{A}\mathbf{S}'\right)_{i,j}+\mathbf{D}_{i,j}}{\mathbf{S}_{i,j}'}\\
%&~~~~~~~~~~+\delta_{il}\delta_{jk}\frac{2\left(\mathbf{BS}'\right)_{i,j}\mathbf{S}_{i,j}'+\mathbf{C}_{i,j}\mathbf{S}'_{i,j}}{\mathbf{S}_{i,j}^2},
%\end{split}
%\end{equation}
\begin{equation}
\begin{split}
&\frac{\partial^2f_s(\mathbf{S},\mathbf{S}')}{\mathbf{S}_{ij}\mathbf{S}_{lk}}=\delta_{il}\delta_{jk}\frac{2\left(\mathbf{A}\mathbf{S}'\right)_{ij}+\mathbf{D}_{ij}}{\mathbf{S}_{ij}'}\\
&~~~~~~~~~~+\delta_{il}\delta_{jk}\frac{2\left(\mathbf{BS}'\right)_{ij}\mathbf{S}_{ij}'+\mathbf{C}_{ij}\mathbf{S}'_{ij}}{\mathbf{S}_{ij}^2}
\end{split}
\end{equation}
is a diagonal matrix with each element no less than $0$, 
where $\delta_{ij}$ is a delta function, i.e.,
\begin{equation}
\delta_{ij}=\begin{cases}
1,~{\rm if}~i==j,\\
0,~{\rm if}~i\neq j.
\end{cases}
\end{equation}
Therefore, \eq{aux-S} is a convex function, where we can get its global minimization by setting $\frac{\partial f_s(\mathbf{S},\mathbf{S}')}{\partial\mathbf{S}_{ij}}=0$, i.e.,
\textcolor{black}{
\begin{equation}
\begin{split}
&\mathbf{S}_{ij}=\mathbf{S}_{ij}'\sqrt{\frac{\mathbf{C}_{ij}+2\left(\mathbf{BS'}\right)_{ij}}{2\left(\mathbf{AS}'\right)_{ij}+\mathbf{D}_{ij}}}=\\
&\mathbf{S}_{ij}'\sqrt{\frac{\left(\mathbf{VV}^\mathsf{T}+\beta\mathbf{W}+\alpha\left(\mathbf{X}^\mathsf{T}\mathbf{X}\right)^++\alpha\left(\mathbf{X}^\mathsf{T}\mathbf{X}\right)^-\mathbf{S}'\right)_{ij}}{\left(\mathbf{S}'+\beta\mathbf{S}'+\alpha\left(\mathbf{X}^\mathsf{T}\mathbf{X}\right)^+\mathbf{S}'+\alpha\left(\mathbf{X}^\mathsf{T}\mathbf{X}\right)^-\right)_{ij}}},
\end{split}
\end{equation}}which is exactly the same as \eq{update-S}. Accordingly, we can conclude that the S-step decreases the objective function of \eq{obj} according to \textbf{Lemma 1}.

\noindent
\textbf{Proof of Lemma 3}: \textbf{Lemma 3} can be proved based on the following 2 
inequalities. 

\noindent
\textbf{Proposition 2} \emph{For any positive matrices $\mathbf{V}>0$, $\mathbf{S}>0$ and  $\mathbf{V}'>0$, the following equations hold:}
\begin{subequations}
	%	\begin{cases}
	\begin{align}
	&{\rm Tr}\left(\mathbf{VV}^\mathsf{T}\mathbf{VV}^\mathsf{T}\right)\leq\sum_{ij}^n\sum_{k=1}^{c}\left(\mathbf{V}'\mathbf{V'}^{\mathsf{T}}\right)_{ij}\mathbf{V}_{ik}'\times\frac{\mathbf{V}_{jk}^4}{\mathbf{V'}_{jk}^{3}}, \\
	&{\rm Tr}\left(-\mathbf{S}\mathbf{VV}^\mathsf{T}\right)\leq-\sum_{ij}^n\sum_{k=1}^{c}\mathbf{S}_{ij}\mathbf{V}_{ik}'\mathbf{V}_{jk}'\left(1+{\rm log}\frac{\mathbf{V}_{ik}\mathbf{V}_{jk}}{\mathbf{V}_{ik}'\mathbf{V}_{jk}'}\right).
	\end{align}
	%\end{cases}
	%
	\label{prop2}
\end{subequations}
\emph{Moreover, all the equalities hold when $\mathbf{V}=\mathbf{V}'$.}

See the proof of \textbf{Proposition 2} in Appendix B.  Let's take
%
%%For $\mathbf{S}$ with fixed $\mathbf{V}$, we have
%
%
%
%%\begin{small}
%
%%\end{small}
%In this section, we provide several theoretical guarantees for the proposed optimization in Algorithm \ref{alg: proposed}, which is summarized in Theorem 1.
%
%
%
%
\begin{equation}
\begin{split}
\frac{\partial f_v\left(\mathbf{V},\mathbf{V}'\right)}{\partial\mathbf{V}_{jk}}=&-2\frac{\left(\mathbf{SV}'\right)_{jk}\mathbf{V}_{jk}'}{\mathbf{V}_{jk}}-2\frac{\left(\mathbf{S}^\mathsf{T}\mathbf{V}'\right)_{jk}\mathbf{V}_{jk}'}{\mathbf{V}_{jk}}\\
&+4\left(\mathbf{V'V'}^\mathsf{T}\mathbf{V}'\right)_{jk}\frac{\mathbf{V}_{jk}^3}{\mathbf{V}_{jk}^{'3}}.
\end{split}
\end{equation}
$f_v\left(\mathbf{V},\mathbf{V}'\right)$'s Hessian matrix is also a positive diagonal matrix with
\begin{equation}
\begin{split}
&\frac{\partial^2 f_v\left(\mathbf{V},\mathbf{V}'\right)}{\partial\mathbf{V}_{jk}\partial\mathbf{V}_{il}}=12\delta_{ji}\delta_{kl}\left(\mathbf{V'V'}^\mathsf{T}\mathbf{V}'\right)_{jk}\frac{\mathbf{V}_{jk}^2}{\mathbf{V}_{jk}^{'3}}\\
&~~~+2\delta_{ji}\delta_{kl}\frac{\left(\mathbf{SV}'\right)_{jk}\mathbf{V}_{jk}'}{\mathbf{V}_{jk}^2}+2\delta_{ji}\delta_{kl}\frac{\left(\mathbf{S}^\mathsf{T}\mathbf{V}'\right)_{jk}\mathbf{V}_{jk}'}{\mathbf{V}_{jk}^2}.
\end{split}
\end{equation}
Therefore, $f_v(\mathbf{V},\mathbf{V}')$ is convex w.r.t. $\mathbf{V}_{jk}$. Let  $\frac{\partial g\left(\mathbf{V},\mathbf{V}'\right)}{\partial\mathbf{V}_{jk}}=0$, we get the minimum of $f_v(\mathbf{V},\mathbf{V}')$ at 
\begin{equation}
\mathbf{V}_{jk}=\mathbf{V}'_{jk}\times ~^4\sqrt{\frac{\left(\mathbf{SV}'\right)_{jk}+\left(\mathbf{S}^\mathsf{T}\mathbf{V}'\right)_{jk}}{2\left(\mathbf{V}'\mathbf{V}^{'\mathsf{T}}\mathbf{V}'\right)_{jk}}}
\end{equation} 
which is exactly the same as \eq{update-V}. Accordingly, we can conclude that the V-step decreases the objective function of \eq{obj}. % according to Lemma 1.
%Combining with Lemma 2, 
%we can conclude that each optimization step can decrease the objective function of \eq{obj}. 
\textcolor{black}{In addition, it is apparent that \eq{obj} is lower-bounded by $0$.  
The proof of \textbf{Theorem 1}-1 is complete}.

%It is easy to verify that $g\left(\mathbf{V}^t,\mathbf{V}^t\right)=\mathcal{O}_{\mathbf{V}}(\mathbf{V}^t)$, and $g\left(\mathbf{V},\mathbf{V}^t\right)\geq \mathcal{O}_{\mathbf{V}}$ according to the following 2 propositions. 
%
%
%%
%%
%%
%Therefore \textbf{Lemma 4} holds. 
%Moreover, $g''\left(\mathbf{V},\mathbf{V}^t\right)>0$, thus \eq{aux-V} is convex, where its global minimum is obtained at $g'\left(\mathbf{V},\mathbf{V}^t\right)=0$, i.e.,
%\begin{equation}
%\mathbf{V}_{i,j}^{t+1}=\mathbf{V}_{i,j}^{t}\sqrt[4]{\frac{\left(\mathbf{SV}^t+\mathbf{S}^\mathsf{T}\mathbf{V}^t+\gamma\mathbf{V}^t\right)_{i,j}}{(2+2\gamma)\left(\mathbf{V}^t\mathbf{V}^{t^\mathsf{T}}\mathbf{V}^t\right)_{i,j}}},
%\end{equation}
%which is exactly the same as \eq{update-V}. As a conclusion, \textbf{Theorem 3} is proved.
%
%
%Moreover, the objective function in \eq{obj-orth} is  lower bounded. Therefore, the proof of Theorem -2 is completed.  

\subsection{Proof of \textbf{Theorem 1}-2}

\textcolor{black}{
At the first iteration, for $i\neq j$, we have
\begin{small}
\begin{equation}
\begin{split}
&\mathbf{S}_{ij}^{1}=\\
&\mathbf{S}_{ij}^0 \left(\frac{\left(\mathbf{V}^0{\mathbf{V}^0}^\mathsf{T}+\alpha\left(\mathbf{X}^\mathsf{T}\mathbf{X}\right)^{+}+\alpha\left(\mathbf{X}^\mathsf{T}\mathbf{X}\right)^{-}\mathbf{S}^0+\beta\mathbf{W}\right)_{ij}}{\left(\mathbf{S}^0+\alpha\left(\mathbf{X}^\mathsf{T}\mathbf{X}\right)^{+}\mathbf{S}^0+\alpha\left(\mathbf{X}^\mathsf{T}\mathbf{X}\right)^{-}+\beta\mathbf{S}^0\right)_{ij}}\right)^{\frac{1}{2}}.
\end{split}
\label{update-SS}
\end{equation}
\end{small}Since 
\begin{small}
\begin{equation}
\begin{split}
&\left(\mathbf{V}^0{\mathbf{V}^0}^\mathsf{T}+\alpha\left(\mathbf{X}^\mathsf{T}\mathbf{X}\right)^{+}+\alpha\left(\mathbf{X}^\mathsf{T}\mathbf{X}\right)^{-}\mathbf{S}^0+\beta\mathbf{W}\right)_{ij}\\
&\geq \left(\mathbf{V}^0{\mathbf{V}^0}^\mathsf{T}\right)_{ij}>0,
\end{split}
\end{equation}
\end{small}and 
\begin{equation}
\left(\mathbf{S}^0+\alpha\left(\mathbf{X}^\mathsf{T}\mathbf{X}\right)^{+}\mathbf{S}^0+\alpha\left(\mathbf{X}^\mathsf{T}\mathbf{X}\right)^{-}+\beta\mathbf{S}^0\right)_{ij}\geq \mathbf{S}_{ij}^0>0,
\end{equation}
we have $\mathbf{S}_{ij}^{1}>0, \forall i,j, {\rm and},i\neq j$.} 

\textcolor{black}{
For $\mathbf{S}_{ii}^{1}, \forall i$, we have 
\begin{small}
\begin{equation}
\begin{split}
&\mathbf{S}_{ii}^{1}=\mathbf{S}_{ii}^0 \left(\frac{\left(\mathbf{V}^0{\mathbf{V}^0}^\mathsf{T}+\alpha\left(\mathbf{X}^\mathsf{T}\mathbf{X}\right)^{+}+\alpha\left(\mathbf{X}^\mathsf{T}\mathbf{X}\right)^{-}\mathbf{S}^0+\beta\mathbf{W}\right)_{ii}}{\left(\mathbf{S}^0+\alpha\left(\mathbf{X}^\mathsf{T}\mathbf{X}\right)^{+}\mathbf{S}^0+\alpha\left(\mathbf{X}^\mathsf{T}\mathbf{X}\right)^{-}+\beta\mathbf{S}^0\right)_{ii}}\right)^{\frac{1}{2}}\\
&=0\times\left(\frac{\left(\mathbf{V}^0{\mathbf{V}^0}^\mathsf{T}+\alpha\left(\mathbf{X}^\mathsf{T}\mathbf{X}\right)^{+}+\alpha\left(\mathbf{X}^\mathsf{T}\mathbf{X}\right)^{-}\mathbf{S}^0+\beta\mathbf{W}\right)_{ii}}{\left(\mathbf{S}^0+\alpha\left(\mathbf{X}^\mathsf{T}\mathbf{X}\right)^{+}\mathbf{S}^0+\alpha\left(\mathbf{X}^\mathsf{T}\mathbf{X}\right)^{-}+\beta\mathbf{S}^0\right)_{ii}}\right)^{\frac{1}{2}}\\
&=0.
\end{split}
\label{Szero}
\end{equation}
\end{small}To avoid the possible numerical inaccuracy cased by the extreme small values in the denominator of Eq. \eqref{Szero}, we could add a small positive value in the denominator. Note that, this additional process does not change the value of $\mathbf{S}_{ii}^{1}, \forall i$. }

\textcolor{black}{
For $\mathbf{V}_{ij}^{1}, \forall i,j$, we have
\begin{equation}
\mathbf{V}_{ij}^{1}=\mathbf{V}_{ij}^0\left(\frac{\left(\mathbf{S}^1\mathbf{V}^0+\mathbf{S}^{1^\mathsf{T}}\mathbf{V}^0\right)_{ij}}{\left(2\mathbf{V}^0\mathbf{V}^{0^\mathsf{T}}\mathbf{V}^0\right)_{ij}}\right)^{\frac{1}{4}}.
\label{update-VV}
\end{equation}
Since $\left(\mathbf{S}^1\mathbf{V}^0+\mathbf{S}^{1^\mathsf{T}}\mathbf{V}^0\right)_{ij}=\sum_{k}\left(\mathbf{S}^1_{ik}\mathbf{V}^0_{kj}+\mathbf{S}^{1^\mathsf{T}}_{ik}\mathbf{V}^0_{kj}\right)\geq \sum_{k\neq i}\mathbf{S}^1_{ik}\mathbf{V}^0_{kj}>0$, and 
$\left(2\mathbf{V}^0\mathbf{V}^{0^\mathsf{T}}\mathbf{V}^0\right)_{ij}>0$, we have $\mathbf{V}_{ij}^{1}>0, \forall i,j$. }

\textcolor{black}{
Based on the above analysis, we get that 
\begin{equation}
\mathbf{V}_{ij}^1>0,\forall i,j, {\rm and }~\begin{cases}
\mathbf{S}_{ij}^1>0,\forall i,j,{\rm and }~i\neq j\\
\mathbf{S}_{ii}^1=0,\forall i.
\end{cases}
\end{equation}
According to mathematical induction, the \textbf{Theorem 1}-2 can be proved. Another additional advantage of our algorithm is that at each iteration ${\rm diag}(\mathbf{S}^t)=0$ holds, such that our algorithm  can remove the trivial solution naturally. }

%\section{Proof of Proposition 1 and Corresponding Derivatives}
\section{Proof of Propositions 1 and 2}
\textcolor{black}{\subsection{Proof of \textbf{Proposition 1} }}
\noindent
\textbf{Proof of Eq. (\ref{prop1}a)}:
Let $\mathbf{S}_{ij}=u_{ij}\mathbf{S}_{ij}'$ and $u_{ij}>0$, $\forall ij$,  we have
\begin{equation}
\begin{split}
&\sum_{ij}^{}\frac{\left(\mathbf{A}\mathbf{S}'\right)_{ij}\mathbf{S}_{ij}^2}{\mathbf{S}_{ij}'}-{\rm tr}\left(\mathbf{S}^\mathsf{T}\mathbf{AS}\right)\\
&=\sum_{ijk}\mathbf{A}_{ik}\mathbf{S}'_{kj}\mathbf{S}'_{ij}u_{ij}^2-\sum_{ijk}\mathbf{A}_{ik}\mathbf{S}'_{kj}\mathbf{S}'_{ij}u_{ij}u_{kj}=\Delta
\end{split}
\label{prop-1a}
\end{equation}
Since $\mathbf{A}$ is a symmetric matrix, we can exchange the indicator ($ik$) in \eq{prop-1a} and get
\begin{equation}
\Delta=\sum_{ijk}\mathbf{A}_{ik}\mathbf{S}'_{kj}\mathbf{S}'_{ij}u_{kj}^2-\sum_{ijk}\mathbf{A}_{ik}\mathbf{S}'_{kj}\mathbf{S}'_{ij}u_{ij}u_{kj}.\
\label{prop-1b}
\end{equation} 
Combining \eq{prop-1a} and \eq{prop-1b} together, we have 
\begin{equation}
	\Delta=\frac{1}{2}\sum_{ijk}\mathbf{A}_{ik}\mathbf{S}'_{kj}\mathbf{S}'_{ij}\left(u_{kj}^2+u_{ij}^2-2u_{ij}u_{kj}\right)\geq 0
\end{equation}
Thus, Eq. (\ref{prop1}a) holds. 
%Moreover, the derivative of $\sum_{ij}\frac{\left(\mathbf{A}\mathbf{S}'\right)_{ij}\mathbf{S}_{ij}^2}{\mathbf{S}_{ij}'}$ w.r.t. $\mathbf{S}_{ij}$ is 
%\begin{equation}
%	\frac{\partial \sum_{ij}\frac{\left(\mathbf{A}\mathbf{S}'\right)_{ij}\mathbf{S}_{ij}^2}{\mathbf{S}_{ij}'}}{\partial \mathbf{S}_{ij}}=2\frac{\left(\mathbf{A}\mathbf{S}'\right)_{ij}\mathbf{S}_{ij}}{\mathbf{S}_{ij}'}.
%\end{equation}
%

\noindent
\textbf{The proof of Eq. (\ref{prop1}b)}:
\begin{equation}
\begin{split}
&{\rm tr}\left(\mathbf{S}^\mathsf{T}\mathbf{BS}\right)-\sum_{ijk}\mathbf{B}_{jk}\mathbf{S}_{ki}'\mathbf{S}_{ji}'\left(1+{\rm log}\frac{\mathbf{S}_{ki}\mathbf{S}_{ji}}{\mathbf{S}_{ki}'\mathbf{S}_{ji}'}\right)=\\
&\sum_{ijk}\mathbf{B}_{jk}\mathbf{S}_{ki}\mathbf{S}_{ji}-\sum_{ijk}\mathbf{B}_{jk}\mathbf{S}_{ki}'\mathbf{S}_{ji}'\left(1+{\rm log}\frac{\mathbf{S}_{ki}\mathbf{S}_{ji}}{\mathbf{S}_{ki}'\mathbf{S}_{ji}'}\right).
\end{split}
\label{prop-2a}
\end{equation}
According to the inequality $x>1+{\rm log}(x),\forall x>0$, and let $x=\frac{\mathbf{S}_{ki}\mathbf{S}_{ji}}{\mathbf{S}_{ki}'\mathbf{S}_{ji}'}$, we can prove \eq{prop-2a}$\geq0$ holds and likewise Eq. (\ref{prop1}b).

\noindent
\textbf{Proof of Eq. (\ref{prop1}c)}:
%\begin{equation}
%{\rm tr}\left(\mathbf{S}^\mathsf{T}\mathbf{C}\right)=\sum_{i,j}^{}\mathbf{C}_{i,j}\mathbf{S}_{i,j}\geq\sum_{i,j}^{}\mathbf{C}_{i,j}\mathbf{S}_{i,j}'\left(1+{\rm log}\frac{\mathbf{S}_{i,j}}{\mathbf{S}_{i,j}'}\right)
%\end{equation}
\begin{equation}
\begin{split}
&{\rm tr}\left(\mathbf{S}^\mathsf{T}\mathbf{C}\right)-\sum_{ij}^{}\mathbf{C}_{ij}\mathbf{S}_{ij}'\left(1+{\rm log}\frac{\mathbf{S}_{ij}}{\mathbf{S}_{ij}'}\right)\\
&=\sum_{ij}^{}\mathbf{C}_{ij}\mathbf{S}_{ij}-\sum_{ij}^{}\mathbf{C}_{ij}\mathbf{S}_{ij}'\left(1+{\rm log}\frac{\mathbf{S}_{ij}}{\mathbf{S}_{ij}'}\right).
\end{split}
\label{prop-3a}
\end{equation}
According to inequality $x>1+{\rm log}(x),\forall x>0$, and let $x=\frac{\mathbf{S}_{ij}}{\mathbf{S}_{ij}'}$, we can prove \eq{prop-3a}$\geq0$ holds and likewise Eq. (\ref{prop1}c).
%The corresponding derivative is calculated as 
%\begin{equation}
%\frac{\partial\sum_{ij}^{}\mathbf{C}_{ij}\mathbf{S}_{ij}'\left(1+{\rm log}\frac{\mathbf{S}_{ij}}{\mathbf{S}_{ij}'}\right)}{\partial\mathbf{S}_{ij}}=\mathbf{C}_{ij}\frac{\mathbf{S}_{ij}'}{\mathbf{S}_{ij}}.
%\end{equation}
%
%

\noindent
\textbf{Proof of Eq. (\ref{prop1}d)}:
%\begin{equation}
%{\rm tr}\left(\mathbf{S}^\mathsf{T}\mathbf{D}\right)=\sum_{i,j}^{}\mathbf{S}_{i,j}\mathbf{D}_{i,j}\leq\sum_{i,j}^{}\mathbf{D}_{i,j}\frac{\mathbf{S}_{i,j}^2+\mathbf{S}_{i,j}^{'2}}{2\mathbf{S}_{i,j}'}
%\end{equation}
%
\begin{equation}
\begin{split}
&{\rm tr}\left(\mathbf{S}^\mathsf{T}\mathbf{D}\right)-\sum_{i,j}^{}\mathbf{D}_{ij}\frac{\mathbf{S}_{ij}^2+\mathbf{S}_{ij}^{'2}}{2\mathbf{S}_{ij}'}\\
&=\sum_{ij}^{}\mathbf{S}_{ij}\mathbf{D}_{ij}-\sum_{ij}^{}\mathbf{D}_{ij}\frac{\mathbf{S}_{ij}^2+\mathbf{S}_{ij}^{'2}}{2\mathbf{S}_{ij}'}.\\
\end{split}
\label{prop-4a}
\end{equation}
According to the Janson inequality $a^2+b^2-2ab\geq0,\forall a,b$, and let $a=\mathbf{S}_{ij}$, $b=\mathbf{S}_{ij}'$, we can prove \eq{prop-4a}$\leq0$ holds and likewise Eq. (\ref{prop1}d).

\textcolor{black}{Besides, for all the functions, the equality hold when $\mathbf{S}=\mathbf{S}'$, and thus the proof of \textbf{Proposition 1} is complete.}

%The corresponding derivative is calculated as 
%\begin{equation}
%\frac{\partial\sum_{ij}^{}\mathbf{D}_{ij}\frac{\mathbf{S}_{ij}^2+\mathbf{S}_{ij}^{'2}}{2\mathbf{S}_{ij}'}}{\partial\mathbf{S}_{ij}}=\mathbf{D}_{ij}\frac{\mathbf{S}_{ij}}{\mathbf{S}_{ij}'}.
%\end{equation}

\textcolor{black}{\subsection{Proof of \textbf{Proposition 2}}}
\noindent
\textbf{Proof of Eq. (\ref{prop2}a)}:
Let $\mathbf{V}_{ij}=u_{ij}\mathbf{V}_{ij}'$ and $u_{ij}>0, \forall i,j$, and $u_{ij}>0$, we have
\begin{equation}
\begin{split}
&\sum_{ij}^n\sum_{k=1}^{c}\left(\mathbf{V}'\mathbf{V}'^{\mathsf{T}}\right)_{ij}\mathbf{V}_{ik}'\times\frac{\mathbf{V}_{jk}^4}{\mathbf{V}_{jk}'^{3}}-{\rm Tr}\left(\mathbf{VV}^\mathsf{T}\mathbf{VV}^\mathsf{T}\right)\\
&=\sum_{ij}^n\sum_{kl}^c\mathbf{V'}_{il}\mathbf{V'}_{jl}\mathbf{V'}_{ik}\mathbf{V'}_{jk}u_{jk}^4\\
&~~~~~~-\sum_{ij}^n\sum_{kl}^c\mathbf{V'}_{il}\mathbf{V'}_{jl}\mathbf{V'}_{ik}\mathbf{V'}_{jk}u_{il}u_{jl}u_{ik}u_{jk}=\Delta.
\end{split}
\label{prop2-1a}
\end{equation}
Denote  $\gamma=\sum_{ij}^n\sum_{kl}^c\mathbf{V'}_{il}\mathbf{V'}_{jl}\mathbf{V'}_{ik}\mathbf{V'}_{jk}u_{il}u_{jl}u_{ik}u_{jk}$, exchanging the indicators $i$ and $j$ in \eq{prop2-1a}, we have
\begin{equation}
\Delta=\sum_{ij}^n\sum_{kl}^c\mathbf{V'}_{il}\mathbf{V'}_{jl}\mathbf{V'}_{ik}\mathbf{V'}_{jk}u_{ik}^4-\gamma.
\label{prop2-1b}
\end{equation}
Exchanging the indicators $k$ and $l$ in \eq{prop2-1a}, we have
\begin{equation}
\Delta=\sum_{ij}^n\sum_{kl}^c\mathbf{V'}_{il}\mathbf{V'}_{jl}\mathbf{V'}_{ik}\mathbf{V'}_{jk}u_{jl}^4-\gamma.
\label{prop2-1c}
\end{equation}
Exchanging the indicators $ik$ and $jl$ in \eq{prop2-1a}, we have
\begin{equation}
\Delta=\sum_{ij}^n\sum_{kl}^c\mathbf{V'}_{il}\mathbf{V'}_{jl}\mathbf{V'}_{ik}\mathbf{V'}_{jk}u_{il}^4-\gamma.
\label{prop2-1d}
\end{equation}
Combining Eqs \eqref{prop2-1a}-\eqref{prop2-1d} together, we have  
\begin{equation}
\begin{aligned}
&\Delta=\sum_{ij}^n\sum_{kl}^c\mathbf{V'}_{il}\mathbf{V'}_{jl}\mathbf{V'}_{ik}\mathbf{V'}_{jk}\Bigl(\frac{u_{il}^4+u_{jl}^4+u_{ik}^4+u_{jk}^4}{4}\\
&~-u_{il}u_{jl}u_{ik}u_{jk}\Bigr)\geq\sum_{ij}^n\sum_{kl}^c\mathbf{V'}_{il}\mathbf{V'}_{jl}\mathbf{V'}_{ik}\mathbf{V'}_{jk}\times\\
&~\left(\frac{u_{il}^2u_{jl}^2+u_{ik}^2u_{jk}^2}{2}-u_{il}u_{jl}u_{ik}u_{jk}\right)\geq0.
\end{aligned}
\end{equation}
\textcolor{black}{
Together with the fact that the equality holds when $\mathbf{V}=\mathbf{V}'$, the proof of Eq. (\ref{prop2}a) is completed. }

%For the corresponding  derivative, we have
%\begin{equation}
%\frac{\partial\sum_{i,j}^n\sum_{k=1}^{c}\left(\mathbf{V}'\mathbf{V}'^{\mathsf{T}}\right)_{ij}\mathbf{V}_{ik}'\times\frac{\mathbf{V}_{jk}^4}{\mathbf{V}_{jk}'^{3}}}{\partial \mathbf{V}_{jk}}=4\left(\mathbf{V}'\mathbf{V}'^{\mathsf{T}}\right)_{ij}\mathbf{V}_{ik}'\mathbf{V}_{jk}^3.
%\end{equation}
%\noindent
%\textbf{Proposition 1}  \emph{For any matrices $\mathbf{V}>0$ and $\mathbf{V}^t>0$,}
%\begin{equation}
%{\rm Tr}\left(\mathbf{VV}^\mathsf{T}\mathbf{VV}^\mathsf{T}\right)\leq\sum_{i,j}^n\sum_{k=1}^{c}\left(\mathbf{V}^t\mathbf{V}^{t\mathsf{T}}\right)_{i,j}\mathbf{V}_{ik}^t\times\frac{\mathbf{V}_{jk}^4}{\mathbf{V}_{jk}^{t^3}},
%\end{equation}
%\emph{holds.}
%

\noindent
\textbf{Proof of Eq. (\ref{prop2}b):}     
\noindent
The proof of Eq. (\ref{prop2}b) is equivalent to the proof of Eq. (\ref{prop1}b). 

%\noindent
%\textbf{Proposition 2} \cite{wu2018pairwise}
%\emph{For any matrices $\mathbf{V}>0$ and $\mathbf{S}>0$, }
%\begin{equation}
%{\rm Tr}\left(-\mathbf{S}\mathbf{VV}^\mathsf{T}\right)\leq-\sum_{i,j}^n\sum_{k=1}^{c}\mathbf{S}_{i,j}\mathbf{V}_{i,k}^t\mathbf{V}_{j,k}^t\left(1+{\rm log}\frac{\mathbf{V}_{ik}\mathbf{V}_{jk}}{\mathbf{V}_{ik}^t\mathbf{V}_{jk}^t}\right),
%\end{equation}
%
%\emph{holds.}

%% use section* for acknowledgment
%\section*{Acknowledgment}


\section{Derivate of \eq{prop1}}
\textcolor{black}{
Here, we only present how to calculate the derivate of \eq{prop1}-b, as those of \eq{prop1}-a, \eq{prop1}-c and \eq{prop1}-d are easy to solve. }

\textcolor{black}{
Let  $f=\sum_{ijk}\mathbf{B}_{jk}\mathbf{S}_{ki}'\mathbf{S}_{ji}'\left(1+{\rm log}\frac{\mathbf{S}_{ki}\mathbf{S}_{ji}}{\mathbf{S}_{ki}'\mathbf{S}_{ji}'}\right) \propto\sum_{ijk}\mathbf{B}_{jk}\mathbf{S}_{ki}'\mathbf{S}_{ji}'\left({\rm log}\mathbf{S}_{ki}+{\rm log}\mathbf{S}_{ji}\right)$, to calculate the derivative of $f$ with respect to $\mathbf{S}_{ij}$, we separate $f=f_1+f_2$, where  $f_1=\sum_{ijk}\mathbf{B}_{jk}\mathbf{S}_{ki}'\mathbf{S}_{ji}'\left({\rm log}\mathbf{S}_{ki}\right)$ and $f_2=\sum_{ijk}\mathbf{B}_{jk}\mathbf{S}_{ki}'\mathbf{S}_{ji}'\left({\rm log}\mathbf{S}_{ji}\right)$. 
Particularly, exchanging $i$ with $j$ in $f_1$, we have  $f_1=\sum_{ijk}\mathbf{B}_{ik}\mathbf{S}_{kj}'\mathbf{S}_{ij}'\left({\rm log}\mathbf{S}_{kj}\right)$.  Then exchanging $i$ and $k$, we have $f_1=\sum_{ijk}\mathbf{B}_{ki}\mathbf{S}_{ij}'\mathbf{S}_{kj}'\left({\rm log}\mathbf{S}_{ij}\right)$. Accordingly, 
\begin{equation}
\frac{\partial f_1}{\partial\mathbf{S}_{ij}}=\frac{\sum_k\mathbf{B}_{ki}\mathbf{S}_{kj}'\mathbf{S}_{ij}'}{\mathbf{S}_{ij}}=\frac{\left(\mathbf{B}^\mathsf{T}\mathbf{S}'\right)_{ij}\mathbf{S}_{ij}'}{\mathbf{S}_{ij}}.
\end{equation}
Exchanging $i$ with $j$ in $f_2$, we have 
$f_2=\sum_{ijk}\mathbf{B}_{ik}\mathbf{S}_{kj}'\mathbf{S}_{ij}'\left({\rm log}\mathbf{S}_{ij}\right)$. Accordingly, 
\begin{equation}
\frac{\partial f_2}{\partial\mathbf{S}_{ij}}=\frac{\sum_k\mathbf{B}_{ik}\mathbf{S}_{kj}'\mathbf{S}_{ij}'}{\mathbf{S}_{ij}}=\frac{\left(\mathbf{BS}'\right)_{ij}\mathbf{S}_{ij}'}{\mathbf{S}_{ij}}.
\end{equation}
Finally,  the corresponding derivative is
\begin{equation}
\frac{\partial f}{\partial\mathbf{S}_{ij}}=\frac{\left(\mathbf{BS}'\right)_{ij}\mathbf{S}_{ij}'}{\mathbf{S}_{ij}}+\frac{\left(\mathbf{B}^\mathsf{T}\mathbf{S}'\right)_{ij}\mathbf{S}_{ij}'}{\mathbf{S}_{ij}}.
\end{equation}
When $\mathbf{B}$ is a symmetric matrix, we have
\begin{equation}
\frac{\partial f}{\partial\mathbf{S}_{ij}}=\frac{2\left(\mathbf{BS}'\right)_{ij}\mathbf{S}_{ij}'}{\mathbf{S}_{ij}}.
\end{equation}}
\ifCLASSOPTIONcaptionsoff
  \newpage
\fi

\bibliographystyle{IEEEtran}
\bibliography{IEEEabrv,bib}

\end{document}